\documentclass[journal]{IEEEtran}
\usepackage{amsmath,amsfonts}
\usepackage{algorithmic}
\usepackage{algorithm}
\usepackage{array}
\usepackage[caption=false,font=normalsize,labelfont=sf,textfont=sf]{subfig}
\usepackage{textcomp}
\usepackage{stfloats}
\usepackage{url}
\usepackage{verbatim}
\usepackage{graphicx}
\usepackage{cite}
\usepackage{amssymb}
\usepackage{mathtools}
\newcounter{MYtempeqncnt}
\usepackage{booktabs}
\usepackage{supertabular}
\usepackage{siunitx} 
\usepackage{multirow}
\usepackage{makecell}
\usepackage{xcolor} 
\usepackage{balance}

\newtheorem{theorem}{Theorem}
\newtheorem{proof}{Proof}

\newcommand{\hh}[1]{{\color{blue}#1}}

\newcommand{\tabincell}[2]{\begin{tabular}{@{}#1@{}}#2\end{tabular}}

\usepackage[utf8]{inputenc}
\usepackage[french]{nomencl}
\usepackage{xcolor}
\newtheorem{remark}{Remark}

\makenomenclature

\usepackage{extarrows}
\begin{document}

\title{Piecewise Linear Strain Cosserat Model for Soft Slender Manipulator}

\author{Haihong Li, Lingxiao Xun, Gang Zheng,~\IEEEmembership{Senior member,~IEEE}
        % <-this % stops a space
        
\thanks{Haihong Li, Lingxiao Xun, Gang Zheng are in Defrost team, Inria, university of Lille, Centrale Lille, CRIStAL - Centre de Recherche en Informatique Signal et Automatique de Lille - UMR 9189, France (e-mail: haihong.li@inria.fr; lingxiao.xun@inria.fr; gang.zheng@inria.fr).}}% <-this % stops a space

\markboth{IEEE TRANSACTIONS ON ROBOTICS}%
{Shell \MakeLowercase{\textit{et al.}}: A Sample Article Using IEEEtran.cls for IEEE Journals}


\maketitle

\begin{abstract}
Recently soft robotics has rapidly become a novel and promising area of research with many designs and applications due to their flexible and compliant structure. However, it is more difficult to derive the nonlinear dynamic model of such soft robots. The differential kinematics and dynamics of the soft manipulator can be formulated as a set of highly nonlinear partial differential equations (PDEs) via the classic Cosserat rod theory. In this work, we propose a discrete modeling technique named piecewise linear strain (PLS) to solve the PDEs of Cosserat-based models, based on which the associated analytic models are deduced. 
%{\color{red}The approach discretizes the slender rod into several sections composed of a number of infinitesimal segments whose states are represented by constant strain twists, while the linearly varying ones are used to describe the states of the sections.} This results in the inherent system independent on arc length of the soft manipulator in a small interval, producing analytic ordinary differential equations which can be easily used for controller design. 
To validate the accuracy of the proposed Cosserat model, the static model of the conical cantilever rod under gravity as a simple example is simulated by using different discretization methods. Results indicate that PLS Cosserat model is comparable to the mechanical deformation behavior of real-world soft manipulator. Finally, a parameters identification scheme for this model is established, and the simulation as well as experimental validation demonstrate that using this method can identify the model physical parameters with high accuracy.
\end{abstract}


\section{Introduction}
\subsection{Review of relevant literature}
\IEEEPARstart{T}{he} soft robots, originally inspired by the structure and behavior of animal species such as octopus tentacles, elephant trunks, snakes and caterpillars \cite{rus2015design}, \cite{rozen2021design}, are an emerging type of robots usually made of continuous deformable elastic elements and idealized as one-dimensional slender objects. Slender elastic rods demonstrating large deflections are increasingly prevalent not only in continuum robots, but also for the applications of the soft robots \cite{till2017elastic}, \cite{chen2020design}. Despite their potential strengths, the deformability of soft robots produces an infinite degree-of-freedom and highly coupled nonlinear system that is much tougher to model. The difficulty in modeling soft robots obviously results from falling short of tools in the field of continuum mechanics.

The beam/rod theory as a subclass of continuum mechanics provides a theoretical guidance to model numerous problems in engineering due to its generality and simplicity. The classical Euler-Bernoulli beam theory has been a widely used approach for the model analysis in the past years, but it depends on the assumption of small deflections. \cite{olson2020euler} developed a quasi-static bending model from a geometrically exact Euler–Bernoulli formulation that generalizes a sequence of soft arm designs to predict the result of design changes. Nevertheless, there exists the drawback that the model ignores bending stiffness and shearing deformation. Timoshenko beam model considering rotational and shear effects has been investigated and applied in soft robots modeling  \cite{lindenroth2016stiffness}, \cite{godaba2019payload}. In applications with large deflections, classical rod theories in nonlinear elasticity are required. The Kirchhoff rod as a geometrically nonlinear extension of the Euler-Bernoulli beam, can only describe bending and torsion deformations and be conveniently adopted to derive general models of soft continuum robots. In other words, it is a promising approach originated from the unshearable and unstretchable Cosserat rod \cite{10.1115/1.4003625}, \cite{novelia2018discrete}. By contrast, the Cosserat rod theory is the geometrically nonlinear generalization of the Timoshenko-Reissner beam, which can model not only bending and torsion, but as well shearing and extension \cite{2016Poincar}, \cite{black2017parallel}, \cite{rucker2011statics}. The Cosserat rod theory has been widely applied to static and dynamic modeling of soft continuum robots. The dissertation about the problems of statics, dynamics, and stability for continuum robots with slender elastic rod on the basis of Cosserat was exhaustively presented in \cite{till2019statics}. Based on the work of \cite{till2019statics}, the authors put forward a new implicit dynamics framework for solving Cosserat partial differential equation (PDE) models applied to soft continuum robots, aiming at addressing the issue of computational difficulties \cite{till2019real}. %In addition, the dynamic model for concentric tube robots was derived to describe the inertial dynamics of a concentric pre-shaped tube system by adapting the dynamic Cosserat rod PDEs \cite{till2020dynamic}
\cite{renda2020geometric} proposed a novel variable-strain parameterization by discretizing the continuous Cosserat rod model onto a finite set of strain basis functions for soft manipulators. Besides, a dynamic Cosserat modeling approach via the strain nonlinear parameterization has been proposed, which is numerically simple with a good accuracy and less degrees of freedom (DoFs). However, this technique leads to computational complexities owing to higher order of basis functions required when simulating significantly complicated deformation such as global buckling behaviors \cite{boyer2020dynamics}. The pseudo-rigid body (PRB) 3R model was first established to account for the large deformation of the flexible beam subject to the tip load, and utilized to analyze the compliant mechanisms with high computational efficiency \cite{10.1115/1.3046148}. On the basis of the PRB 3R model, \cite{huang20193d} presented a three-dimensional (3D) static modeling method with high model precision for cable-driven continuum robot for the purpose of the extension from the PRB 3R model to 3D applications.% while taking into account of the effects of elastic force and external loads.

The rod modeling frameworks to describe soft continuum robots have been laid down in light of the above reviewed beam theories. Continuous rod models have infinite-dimension states, and as such, they are significantly difficult to model and control than the discrete systems with finite DoFs. Modeling and control of soft robots for some practical applications calls for accurate and efficient models that will enable simulation of mechanical behavior, improve structural design and development of control systems. The existing alternative to yield to a finite-dimensional model of the robot can be useful for control design purposes. The mathematical foundations from the rod theories facilitate an assortment of discrete computational models. In practice, these discrete approaches allow to significantly reduce the number of DoFs of the calculated mechanical model, and simultaneously ensure the model precision. There are several typically discrete modeling techniques for soft continuum robots, which mainly contains the piecewise constant curvature (PCC) models \cite{webster2010design}, finite element models/methods (FEM) \cite{sonneville2014geometrically}, \cite{largilliere2015real}, \cite{zheng2019controllability} and the piecewise constant strain (PCS) Cosserat models \cite{renda2018discrete}.

PCC modeling approach describes soft robots with a finite number of arcs parameterized by three quantities (i.e., (curvature, arc length, and bending plane) \cite{webster2010design}, leading to reduced-order and relatively simple mathematical models. This method has proven to be an excellent technique with a wide range of applications \cite{godage2016dynamics}, \cite{jones2006kinematics},  \cite{falkenhahn2016dynamic}. However, when considering external loads, the PCC may not be accurate owing to ignoring the presence of torsion, shear and extension. Besides, the PCC parameterization of the soft robot is independent on intrinsic variables, which results in kinematic discontinuities and singularities although they can be avoided with alternative parameterization \cite{della2020improved}, \cite{allen2020closed}. These flaws can potentially produce critical behaviors in the real-world applications.

FEM-based method which is formulated as ways of approximate solutions of PDEs was used to model physical behavior of soft robots \cite{de2017development}, \cite{grazioso2019geometrically}. Nevertheless, to obtain exact modeling precision, this method demands that the number of nodes should tend to infinity, which negatively increases the dimension of the system and leads to a higher computation complexity. Moreover, FEM models for large-deformation 3D nonlinear elasticity always involve unnecessary computational expense when modeling long, slender arms like rod-based soft robots because general deformations of the cross sections of the model are included. 

%To overcome these shortcomings, model-order reduction techniques of the FEM approach have been introduced in \cite{qu2004model}, \cite{goury2018fast}, \cite{sadati2021reduced}, aiming at decreasing the number of DoFs and consequently improving the computational efficiency, but this improvement in turn may decrease the modeling precision.

PCS Cosserat modeling technique, a discrete version of the continuous Cosserat approach \cite{cao2008nonlinear}, \cite{renda2014dynamic}, \cite{haibin2018modeling},  \cite{zhang2019modeling}, was presented in \cite{renda2018discrete}. It employed a finite set of piecewise constant strains with discontinuities happening at fixed points along the rod to model the deformation of the soft robots. This modeling method provides advantages to use with not only a few state variables, but also a relatively high modeling precision. As a potential and powerful alternative to the 3D FEM, the PCS Cosserat dynamic models have been increasingly applied to robots \cite{hussain2021compliant},  \cite{armanini2021discrete}, \cite{thuruthel2018model}. However, from practical point of view, with regards to the soft robots modeled through the PCS approximation, the number of sections to be divided depends on the designer of the model as a result of application-specific considerations. Remarkably, the deformation field of any certain section under external forces is not necessarily constant along the soft arm in a real scenario. The PCS is able to exactly approximate to the continuum formulation when the arm is divided into more sections, but the dimension of state variables and computational cost will highly increase, which is difficult to be used for control purposes. Consequently, a trade-off between model order reduction and accuracy should be established when soft robots modeled by the PCS approach are used for simulation and controller design. 

In the past three decades, numerous model parameters identification techniques have been proposed by researchers. Several major contributions for dynamic model identification and its applications in the robotics control have been reported in \cite{wu2010overview}, \cite{jovic2016humanoid}. These studies assumed that the kinematic models were accurate, however, some typical investigations presented uncertainties of kinematic models \cite{zhang2017adaptive}, \cite{chen2017tracking}. The parameters identification of static model is more obtainable than the dynamic model in practical applications since it just needs information of joint position rather than joint velocity and acceleration. The contributors of \cite{palpacelli2014experimental} derived the gravity torque by employing a Lagrangian method for a specific robot and acquired fine experimental results.  \cite{dumas2011joint} developed a new methodology for the joint stiffness identification of six-revolute industrial serial robots, and proposed a fast and robust procedure that can be used for the stiffness identification of the robot. In \cite{han2019static}, the authors introduced a procedure of the static model identification towards general serial articulated manipulator and presented an application of the identified static model to validate the proposed methods. More recently, to the best of our knowledge, there remains short of the systematic approach for physical parameters identification of the static model for soft robots.

The Cosserat rod model is the closest to the mechanics of deformation of the soft robots because it can produce an exact nonlinearity in the deformations due to bending, torsion, extension, and shearing. In this paper, we first put forward a piecewise linear parameterization of the rod shape by its strain fields, namely piecewise linear strain (PLS) Cosserat model. %On the one hand, the PLS that can guarantee a better deformation continuity locally approximates to the configuration of the arm, which leads to a better precision although producing the higher dimension than the nonlinear parameterization of strain fields when describing high-order buckling behavior; on the other hand, this approach can precisely simulate the configuration of the soft manipulator by being divided into fewer sections than the existing discrete Cosserat model such as the PCS, resulting in an accurate modeling of the real system with much less DoFs. 
This discrete Cosserat technique, striking a balance between accuracy, robustness and computational complexity, aims to address the lack of rigorous modeling techniques in soft robotics and develop a general framework between soft and rigid robots. Apart from the proposed modeling method, a systematic parameters identification framework based on the PLS Cosserat model for soft manipulators with arbitrary actuation manner is constructed, which is necessary and promising towards the development of model-based controller design.% Both simulation and experimental results on the trunk-like manipulator demonstrate the effectiveness of the proposed parameters identification scheme.

\subsection{Contributions}
The soft manipulator is modeled as a continuous assembling of cross sections moving on the midline in the 3D space via infinite rigid transformations dependent on the internal deformations. The geometrical assumption of considering no cross-section deformation allows to describe the soft arm by using ideas from the Lie group structure of rigid body motions and the other concepts from the techniques of differential geometry. In conclusion, the distinct contributions of this paper are summarized as follows:
\begin{itemize}
	\item[*] Propose a discrete modeling approach named PLS for large elastic deformations via the Cosserat rod theory with application to soft manipulators. The analytically integrable models involving geometric, differential kinematic and dynamic models are obtained. Simultaneously, derivation of a novel strain mode selection scheme via PLS Cosserat is provided to model different simplified beam models. 
\end{itemize}
\begin{itemize}
	\item[*] Present a physical parameter identification strategy for the discrete Cosserat model with arbitrary actuation manner by solving a nonlinear programming (NLP) problem. Experimental setup using a cable-driven soft manipulator made of silicone indicates that the proposed method can effectively get more accurate model physical parameters. The PLS Cosserat model with the identified material parameters is capable of predicting the position of end-effector of the arm  with a small relative error.
\end{itemize}
\begin{itemize}
	\item[*] Implement the static simulation of proposed method for soft manipulator actuated by cables, validate the precision of PLS Cosserat model, and compare the systems with different modes.
\end{itemize}

\vspace*{-5pt}
\subsection{Outline}
The remainder of the paper is structured as follows. In Section II, the continuum models of the soft arm via the Cosserat rod on Lie group are recalled. Section III presents the detailed mathematical derivation of the novel discrete Cosserat model approach and strain mode choice scheme via Cosserat. In Section IV, a model parameters identification method based on the discrete Cosserat model is proposed. In Section V, the comparison of accuracy and computational efficiency for the discrete Cosserat models with FEM is performed through a cantilever rod simulation. In Section VI, a conical soft manipulator is designed and used to implement the model validation. Finally, the conclusion and future work are made.

\section{Derivation of Cosserat rod PDEs}
\begin{figure}[!ht]
	\centering
	\includegraphics[width=2.5in]{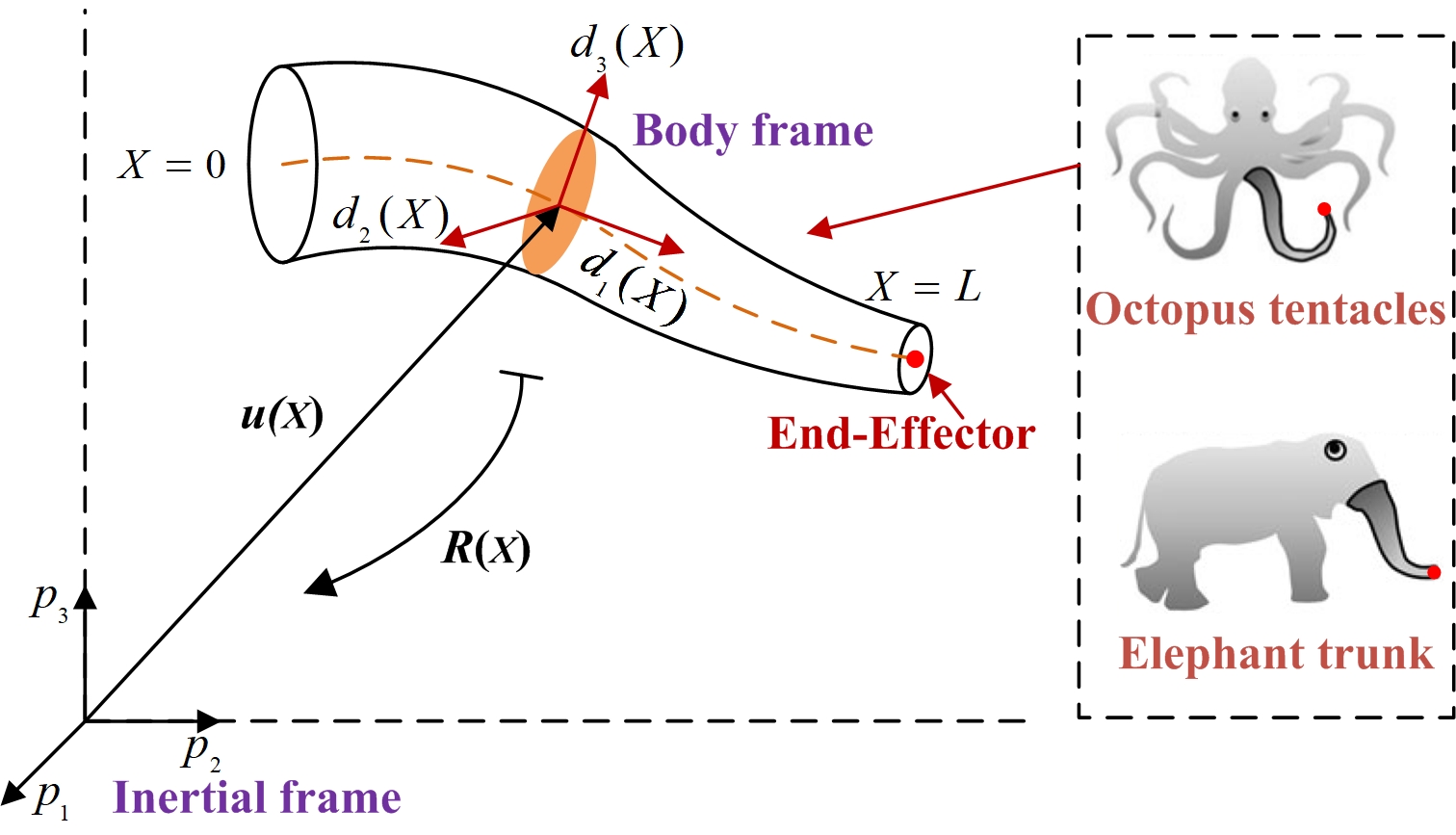}
	\caption{Geometric description of a Cosserat rod and its applications.}
	\label{Cosserat}
\end{figure}

\begin{table}
	\centering
	\caption{Nomenclature and definitions}
	\begin{tabular}{lll}
		\toprule
		\multirow{1}{*}{Symbol} & \multicolumn{1}{l}{\tabincell{l}{Unit }}&\multicolumn{1}{l}{\tabincell{c}{Definition}}\\
		\midrule
		$X$ & m  & Arc length. \\
		$t$ & s  & Time. \\
		$\boldsymbol{R}(X, t)$ & --- & Rotation matrix.\\
		$\boldsymbol{u}(X, t)$ & m & Position vector. \\
		\multirow{1}{*}{$\boldsymbol{g}(X, t)$} & \multicolumn{1}{l}{\tabincell{l}{--- }}&\multicolumn{1}{l}{\tabincell{l}{The configuration matrix $\scriptsize
		\boldsymbol{g}(X, t)=\left(\begin{matrix}
		\boldsymbol{R}&\boldsymbol{u}\\
		\boldsymbol{0}^{\rm T}&1
		\end{matrix}\right).$}}\\
		\multirow{1}{*}{$\boldsymbol{K}(X, t)$ } & \multicolumn{1}{l}{\tabincell{l}{1/m }}&\multicolumn{1}{l}{\tabincell{l}{Angular strain in the body frame.}}\\
		\multirow{1}{*}{$\boldsymbol{Q}(X, t)$ } & \multicolumn{1}{l}{\tabincell{l}{--- }}&\multicolumn{1}{l}{\tabincell{l}{Linear strain in the body frame.}}\\
		\multirow{1}{*}{$\boldsymbol{\Omega}(X, t)$ } & \multicolumn{1}{l}{\tabincell{l}{1/s }}&\multicolumn{1}{l}{\tabincell{l}{Angular velocity in the body frame.}}\\
		\multirow{1}{*}{$\boldsymbol{V}(X, t)$ } & \multicolumn{1}{l}{\tabincell{l}{m/s }}&\multicolumn{1}{l}{\tabincell{l}{Linear velocity in the body frame.}}\\
		\multirow{1}{*}{$\widetilde{(\cdot)}$} & \multicolumn{1}{l}{\tabincell{l}{--- }}&\multicolumn{1}{l}{\tabincell{l}{Mapping from $\mathbb{R}^3$ to $so(3)$, \\e.g. $\scriptsize
		\widetilde{\boldsymbol{a}}=\left[ \begin{matrix}
		0&-a_3&a_2\\
		a_3&0&-a_1\\
		-a_2&a_1&0
		\end{matrix}\right] $.}}\\
		\multirow{1}{*}{$\widehat{(\cdot)}$} & \multicolumn{1}{l}{\tabincell{l}{---}}&\multicolumn{1}{l}{\tabincell{l}{ Mapping from $\mathbb{R}^6$ to $se(3)$,\\ e.g. $\scriptsize
		\widehat{\boldsymbol{\xi}}=\left(  \begin{matrix}
		\widetilde{\boldsymbol{K}}&\boldsymbol{Q}\\
		\boldsymbol{0}&0
		\end{matrix}\right),\ \widehat{\boldsymbol{\eta}}=\left(  \begin{matrix}
		\widetilde{\boldsymbol{\Omega}}&\boldsymbol{V}\\
		\boldsymbol{0}&0
		\end{matrix}\right)\in se(3) $ }}\\
		${{\rm{d}}X}$ & m & Infinitesimal material element. \\
		$\rho$ & kg/m$^3$ & Density of material. \\
		$R(X)$ & m & Cross-sectional radius. \\
		$A(X)$ & m$^2$ & Cross-sectional area. \\
		$E$ & Pa & Young modulus. \\
		$\nu$ & --- & Poisson ratio. \\
		\multirow{1}{*}{$G$} & \multicolumn{1}{l}{\tabincell{l}{Pa }}&\multicolumn{1}{l}{\tabincell{l}{Shear modulus (For the isotropic material,\\ $G=E/(2(1+\nu))$).}}\\
		$\mu$ & Pa$\cdot$s & Viscosity modulus. \\
		%\multirow{1}{*}{$\boldsymbol{I}(X)$} & \multicolumn{1}{l}{\tabincell{l}{m$^4$ }}&\multicolumn{1}{l}{\tabincell{l}{Second moment of area tensor in the \\inertial frame.}}\\
		$\mathbf{I}_3$& --- & 3 $\times$ 3 identity matrix.\\
		\multirow{1}{*}{$\boldsymbol{\mathcal{J}}(X)$} & \multicolumn{1}{l}{\tabincell{l}{m$^4$ }}&\multicolumn{1}{l}{\tabincell{l}{Second moment of area tensor in the body \\frame, $\scriptsize
		\boldsymbol{\mathcal{J}}=\left[ \begin{matrix} J_x&0&0\\	0&J_y&0\\0&0&J_z \end{matrix}\right] $ (For a circular \\rod, $\scriptsize J_x=J_y+J_z$, and $\scriptsize J_y=J_z=\pi R^4/4$,\\ $\scriptsize J_y,\ J_z$ are separately the second moments \\of the area w.r.t. axis $Y$ and $Z$, $J_x$ is the\\ polar moment of the area around the axis $X$).}}\\ 
		\multirow{1}{*}{$\rm{ad}_{(\cdot)}$} & \multicolumn{1}{l}{\tabincell{l}{--- }}&\multicolumn{1}{l}{\tabincell{l}{The adjoint map of the Lie algebra, e.g.\\ $\scriptsize\rm{ad}_{\boldsymbol{\xi}}= \left(\begin{matrix}
		\widetilde{\boldsymbol{K}}&\boldsymbol{0}_{3\times3}\\\widetilde{\boldsymbol{Q}}&\widetilde{\boldsymbol{K}}
		\end{matrix}\right)$, $\scriptsize\ \rm{ad}_{\boldsymbol{\eta}}= \left(\begin{matrix}
		\widetilde{\boldsymbol{\Omega}}& \boldsymbol{0}_{3\times3}\\
		\widetilde{\boldsymbol{V}}&\widetilde{\boldsymbol{\Omega}}
		\end{matrix}\right).$}}\\ 
		%	\multirow{1}{*}{$\rm{ad}^{\rm T}_{(\cdot)}$ } & \multicolumn{1}{l}{\tabincell{l}{--- }}&\multicolumn{1}{l}{\tabincell{l}{The adjoint map of the Lie algebra.\\ $\scriptsize\rm{ad}^{\rm T}_{\boldsymbol{\xi}}= \left(\begin{matrix}
		%	\widetilde{\boldsymbol{K}}^{\rm T}&\widetilde{\boldsymbol{Q}}^{\rm T}\\\boldsymbol{0}_{3\times3}&\widetilde{\boldsymbol{K}}^{\rm T}
		%	\end{matrix}\right)\in \mathbb{R}^{6\times6}$,\\ $\scriptsize\ \rm{ad}^{\rm{T}}_{\boldsymbol{\eta}}= \left(\begin{matrix}
		%	\widetilde{\boldsymbol{\Omega}}^{\rm{T}}& \widetilde{\boldsymbol{V}}^{\rm{T}}\\
		%	\boldsymbol{0}_{3\times3}&\widetilde{\boldsymbol{\Omega}}^{\rm{T}}
		%	\end{matrix}\right)\in \mathbb{R}^{6\times6}$}}\\
		\multirow{1}{*}{$\boldsymbol{\mathcal{M}}(X)$} & \multicolumn{1}{l}{\tabincell{l}{--- }}&\multicolumn{1}{l}{\tabincell{l}{Cross-sectional mass matrix.}}\\
		%,\\ $\scriptsize
		%\boldsymbol{\mathcal{M}}=\rho\left[ {\rm{diag}}\left(
		%J_x, J_y, J_z, A, A, A
    	%\right)\right] \in \mathbb{R}^{6\times6}
		%$
		\multirow{1}{*}{$\boldsymbol{K}_{tb}$} & \multicolumn{1}{l}{\tabincell{l}{N$\cdot$m }}&\multicolumn{1}{l}{\tabincell{l}{Stiffness matrix for torsion and bending,\\ $\scriptsize
		\boldsymbol{K}_{tb}=\left[ \begin{matrix}
		G&0&0\\
		0&E&0\\
		0&0&E
		\end{matrix}\right]\boldsymbol{\mathcal{J}}(X) .$}}\\
		\multirow{1}{*}{$\boldsymbol{K}_{es}$}& \multicolumn{1}{l}{\tabincell{l}{N }}&\multicolumn{1}{l}{\tabincell{l}{Stiffness matrix for elongation and shearing,\\ $\scriptsize
		\boldsymbol{K}_{es}=\left[ \begin{matrix}
		E&0&0\\
		0&G&0\\
		0&0&G
		\end{matrix}\right]A(X). $}}\\
		\multirow{1}{*}{$\boldsymbol{D}_{tb}$} & \multicolumn{1}{l}{\tabincell{l}{N$\cdot$m$^2$$\cdot$s }}&\multicolumn{1}{l}{\tabincell{l}{Damping matrix for torsion and bending,\\ $\scriptsize
		\boldsymbol{D}_{tb}=\left[ \begin{matrix}
		\mu&0&0\\
		0&3\mu&0\\
		0&0&3\mu
		\end{matrix}\right]\boldsymbol{\mathcal{J}}(X) .$}}\\
		\multirow{1}{*}{$\boldsymbol{D}_{es}$} & \multicolumn{1}{l}{\tabincell{l}{N$\cdot$s}}&\multicolumn{1}{l}{\tabincell{l}{Damping matrix for elongation and \\shearing, $\scriptsize
		\boldsymbol{D}_{es}=\left[ \begin{matrix}
		3\mu&0&0\\
		0&\mu&0\\
		0&0&\mu
		\end{matrix}\right]A(X). $}}\\
	    $\boldsymbol{\mathcal{P}}$ & --- & Generalized selection matrix. \\
	    $(\cdot)^{\vee}$& --- & Mapping from a matrix to a vector.\\
		\multirow{1}{*}{${\rm{Ad}}_{\boldsymbol{g}(X)}$ } & \multicolumn{1}{l}{\tabincell{l}{--- }}&\multicolumn{1}{l}{\tabincell{l}{The matrix transforming the velocity or \\acceleration twist from body frame to \\inertial frame, \\i.e., $\scriptsize {\rm{Ad}}_{\boldsymbol{g}(X)}= \left(\begin{matrix}
		\boldsymbol{R}&\boldsymbol{0}_{3\times3}\\\widetilde{\boldsymbol{u}}\boldsymbol{R}&\boldsymbol{R}
		\end{matrix}\right)\in \mathbb{R}^{6\times6}.$}}\\
     	\multirow{1}{*}{$\boldsymbol{g}_r$} & \multicolumn{1}{l}{\tabincell{l}{--- }}&\multicolumn{1}{l}{\tabincell{l}{The transformation matrix between the \\inertial frame and the manipulator base\\ frame.}}\\
	%    $g$ & m/s$^2$ & Gravity acceleration ($z$-axis direction).\\	
		$\Vert\cdot\Vert_2$ & --- & Euclidean norm of a vector or matrix. \\
%		\multirow{1}{*}{$s$} & \multicolumn{1}{l}{\tabincell{l}{---}}&\multicolumn{1}{l}{\tabincell{l}{Number of cables attached at the end of \\the soft manipulator.}}\\
		$N$& --- & Total number of sections divided.\\
		$\overline{N}$& --- & Number of experiments.\\
		$\mathcal{L}$& --- & Lagrangian function.\\
		\multirow{1}{*}{$\boldsymbol{d}_i(X)$} & \multicolumn{1}{l}{\tabincell{l}{$\rm m$ }}&\multicolumn{1}{l}{\tabincell{l}{Local distance between the midline of soft rod\\ and the cable.}}\\	
		$\boldsymbol{{\rm t}}_{ci}(X,t)$& --- & Unit vector tangent to the cable path.\\		
		\bottomrule
	\end{tabular}
	\label{notation_definitions}
\end{table}
A Cosserat rod is a one-dimensional slender continuum deformable body, and each cross section of the rod is considered as an infinitesimal material element whose size and shape cannot change under external forces, as illustrated in orange part in Fig.~\ref{Cosserat}. Thus, all variables of the rod can be parameterized by the reference arc length $X\in [0, L]\subset \mathbb{R}$ along the undeformed rod and by the time $t\in \mathbb{R}$. For the purpose of model derivation, some variables are expressed in the inertial frame with bold small letters, but defined in the body frame attached to the cross sections as denoted by the bold capital letters. For ease of reference, the nomenclature we selected is summarized in Table~\ref{notation_definitions}. With the aim to make the discussion of the discretization more concrete, we start by presenting the PDE system describing the Cosserat rod.

As depicted in Fig. \ref{Cosserat}, the position field of any cross section can be represented by a centerline vector $\small\boldsymbol{u}(X,t)\in\mathbb{R}^3$ and an orthonormal rotation matrix $\small\boldsymbol{R}(X,t)\in SO(3)$ with respect to the inertial frame at time $t$. Hence, the homogeneous transformation matrix can be then defined as 
$\small
 \boldsymbol{g}(X, t) =\left(\begin{matrix}
\boldsymbol{R}&\boldsymbol{u}\\
\boldsymbol{0}^{\rm T}&1
\end{matrix}\right) \in SE(3)
$.
For the cross section at $X\in [0,L]$, denote its strain by $\small\boldsymbol{\xi}(X, t)=\left( \boldsymbol{K}^{\rm T}, \boldsymbol{Q}^{\rm T}\right)^{\rm T} \in\mathbb{R}^6$ where 
$\small\boldsymbol{K}(X,t)\in \mathbb{R}^3$ and $\small\boldsymbol{Q}(X,t)\in \mathbb{R}^3$ represent respectively the angular strain (bending and torsion) and linear strain (shearing and extension), 
and denote its velocity by $\small\boldsymbol{\eta}(X, t)=\left( \boldsymbol{\Omega}^{\rm T}, \boldsymbol{V}^{\rm T}\right)^{\rm T} \in\mathbb{R}^6
$
where $\small\boldsymbol{\Omega}(X,t)\in \mathbb{R}^3$ and $\small\boldsymbol{V}(X,t)\in \mathbb{R}^3$ respectively stand for the angular and linear velocity.

Based on the above notations, the corresponding kinematic model was derived in \cite{armanini2021discrete}:
\begin{equation}
\small
\boldsymbol{g}'=\boldsymbol{g}\widehat{\boldsymbol{\xi}},\ \dot{\boldsymbol{g}}=\boldsymbol{g}\widehat{\boldsymbol{\eta}}
\label{geo}
\end{equation}
\begin{equation}
\small
\boldsymbol{\eta}'=\dot{\boldsymbol{\xi}}(X)-{\rm{ad}}_{\boldsymbol{\xi}(X)}\boldsymbol{\eta}(X)
\label{kine1}
\end{equation}
\begin{equation}
\small
\dot{\boldsymbol{\eta}}'=\ddot{\boldsymbol{\xi}}(X)-{\rm{ad}}_{\dot{\boldsymbol{\xi}}(X)}\boldsymbol{\eta}(X)-{\rm{ad}}_{\boldsymbol{\xi}(X)}\dot{{\boldsymbol{\eta}}}(X)
\label{kine2}
\end{equation}
as well as the dynamical model defined below:
\begin{equation}
	\small
	\begin{split}
	\boldsymbol{\mathcal{M}}\dot{{\boldsymbol{\eta}}}-{\rm{ad}}^{\rm{T}}_{\boldsymbol{{\eta}}}\boldsymbol{\mathcal{M}}\boldsymbol{{\eta}}&=\boldsymbol{\mathcal{F}}'_i-{\rm{ad}}^{\rm{T}}_{\boldsymbol{{\xi}}}\boldsymbol{\mathcal{F}}_{i}+\overline{\boldsymbol{\mathcal{F}}}_e\\
	\boldsymbol{\mathcal{F}}_i&=\boldsymbol{\mathcal{F}}_{ie}+\boldsymbol{\mathcal{F}}_{ia}
	\label{strong_form}
	\end{split}
\end{equation}
where $\small\boldsymbol{\mathcal{M}}=\rho\left[ {\rm{diag}}\left(
J_x, J_y, J_z, A, A, A
\right)\right] \in \mathbb{R}^{6\times6}
$, 
$\small\boldsymbol{\mathcal{F}}_i$ represents the internal wrench, $\small\overline{\boldsymbol{\mathcal{F}}}_{e}$ is distributed external wrench, $\small\boldsymbol{\mathcal{F}}_{ie}$ denotes internally elastic wrench, and $\small\boldsymbol{\mathcal{F}}_{ia}$ stands for the internal wrench produced by the actuation. The above PDE (\ref{strong_form}) is defined with the boundary conditions (BCs) of the internal wrench and configurations of the tip at $X=0$ and $X=L$
\begin{equation}
	\small
	\begin{split}
		\boldsymbol{\mathcal{F}}_{i}(0)&=-\boldsymbol{\mathcal{F}}_{e0}, \ {\rm{or}}\ \boldsymbol{g}(0)=\boldsymbol{g}_0\\ \boldsymbol{\mathcal{F}}_{i}(L)&=\boldsymbol{\mathcal{F}}_{eL}, \ {\rm{or}}\ \boldsymbol{g}(L)=\boldsymbol{g}_L
		\label{BC}
	\end{split}
\end{equation}
where $\small\boldsymbol{\mathcal{F}}_{e0}$ and $\small\boldsymbol{\mathcal{F}}_{eL}$ are tip external wrenches at $X=0$ and $X=L$, respectively.

As for the internal elastic wrench, the Kelvin-Voigt model  \cite{linn2013geometrically} can be adopted both for the elastic and viscous members because of the constitutive material behavior of the soft manipulator, i.e.,
\begin{equation}
	\small
	\boldsymbol{\mathcal{F}}_{ie}(X)=\boldsymbol{\Sigma}(X)(\boldsymbol{{\xi}}(X)-\boldsymbol{{\xi}}_0)+\boldsymbol{\gamma}(X)\dot{{\boldsymbol{\xi}}}
	\label{inter}
\end{equation}
with 
$$\small\boldsymbol{\Sigma}(X)=\left[\begin{matrix}
	\boldsymbol{K}_{tb}&\\  &\boldsymbol{K}_{es}
\end{matrix} \right], \ \boldsymbol{\gamma}(X)=\left[\begin{matrix}
	\boldsymbol{D}_{tb}& \\ & \boldsymbol{D}_{es}
\end{matrix} \right]$$ 
where $\small\boldsymbol{K}_{tb}={\rm{diag}}\left( 
GJ_x(X),EJ_y(X),EJ_z(X)
\right)\in\mathbb{R}^{3\times3}$ and $\small\boldsymbol{K}_{es}={\rm{diag}}\left( 
EA(X),GA(X),GA(X)
\right) \in\mathbb{R}^{3\times3}$ are stiffness matrices determined by the material properties and cross-sectional geometry, $\small\boldsymbol{D}_{tb}={\rm{diag}}\left( 
\mu J_x(X),3\mu J_y(X),3\mu J_z(X)
\right)\in\mathbb{R}^{3\times3}$ and $\small\boldsymbol{D}_{es}={\rm{diag}}\left( 
3\mu A(X),\mu A(X),\mu A(X)
\right)\in\mathbb{R}^{3\times3}$ are separately viscosity matrices for Kelvin–Voigt-type viscous damping, $\small\boldsymbol{\xi}_0$ represents the strain related to initial configuration of the manipulator. Note that a rod cross-section does not require to be circular, and it is only needed to be slender.

The actuation wrench $\small\boldsymbol{\mathcal{F}}_{ia}$ in (\ref{strong_form}) depends on the type of actuators used, and the most common actuation manners for soft manipulators are tendon and fluidic actuators. In terms of the external wrench $\small\overline{\boldsymbol{\mathcal{F}}}_e$ in (\ref{strong_form}), we can consider distributed load produced by gravity, or point load exerted by external disturbance in accordance with the actual situation.

Generally, the solution of the strong form (\ref{strong_form}) can be approximated by using many different approaches, such as implicit finite difference method \cite{till2020dynamic}, the shooting method \cite{till2019real}, assumed mode method \cite{della2019control} and so on. In \cite{renda2018discrete}, by introducing the virtual displacement $\small\delta\boldsymbol{\phi}(X)\in\mathbb{R}^6$,  the D'Alembert's principle is used to obtain the weak form of (\ref{strong_form}), 
then the length space $[0,L]$ was discretized into $N$ sections, and the strain is assumed to be constant for each section (i.e., PCS: piecewise constant strain) in order to deduce analytic formula. Clearly, such an assumption requires finely spatial discretization which yields relatively high dimensional system. Such a method can provide enough precision and work well for numerical simulation which might take time. However, PCS will be a big issue when designing model-based controllers.

To reduce the dimension of the deduced system, \cite{boyer2020dynamics} proposed to globally approximate the strain field as $\boldsymbol{\xi} (X,t)=\boldsymbol{\Phi}(X)\boldsymbol{q}(t)$ where $\boldsymbol{\Phi}(X)=(\Phi_1,\Phi_2,\cdots,\Phi_n)$ defines $n$ basis functions to parameterize the strain space (i.e., VS: variable strain). 
	The advantage of VS is that the dimension of the resulted dynamical model is quite low with respect to PCS, but can provide comparable precision if the number of basis function is high enough. 
	 However, the choice of the number of basis function 
%	
%	Although this method provides us with a set of ODEs that can be used for numerical simulation and control, the order of basis function used 
	is highly dependent of external disturbance. For example, the external disturbance such as concentrated loads will lead to local strain mutation, and a global VS approximation method will be difficult to guarantee the local fitting precision. Moreover, all the kinematics are no more analytically integrable in contrast to PCS, but numerically reconstructed with a quaternion-based integrator. In other words, the deduced model is not anymore analytic, and will be not very friendly for control design.
%	
%	
%	, which is very easy to perform. However, the use of the numerical integration of a rotation matrix can lose orthogonality due to accumulation of numerical resolution error \cite{markley2008unit}.} %Drawbacks: numerical resolution.  the precision depends on external unknown disturbance.

In this paper, we will combine the advantages of each method, i.e.,  two ideas: small discretization to keep the local approximation precision (PCS), and interpolation to decrease the dimension (VS), so as to propose a piecewise linear strain (PLS) approach which is capable of obtaining the analytic formula of the model and facilitating the control design.
%
%
%
%In this way, the dynamics PDE (\ref{dynamic}) can be transformed from the strong form to the above weak one which simplifies the complexity of the state space and lays the foundation for discrete Cosserat dynamic modeling. Subsequently, we will introduce a novel modeling method via Cosserat in order to solve (\ref{weak form}) in the following section. 

\section{Piecewise linear strain Cosserat model}
In this section, the geometric (or kinematic), differential kinematic and dynamic model of the PLS Cosserat rod will be developed. Afterwards, the rod kinematics is reduced by neglecting any component of the six internal degrees of freedom (DoFs) via the PLS Cosserat model, which can effectively model Euler-Bernoulli (E-B) beam, extensible Kirchhoff (E-K) rod and Timoshenko beam, etc.

\subsection{Idea of PLS}

Firstly, we would like to give a global picture to illustrate the difference between PCS, VS and PLS Cosserat modeling methods, and highlight the advantage of the PLS method.
	
\begin{figure}[!h]
	\centering
	\includegraphics[width=2.5in]{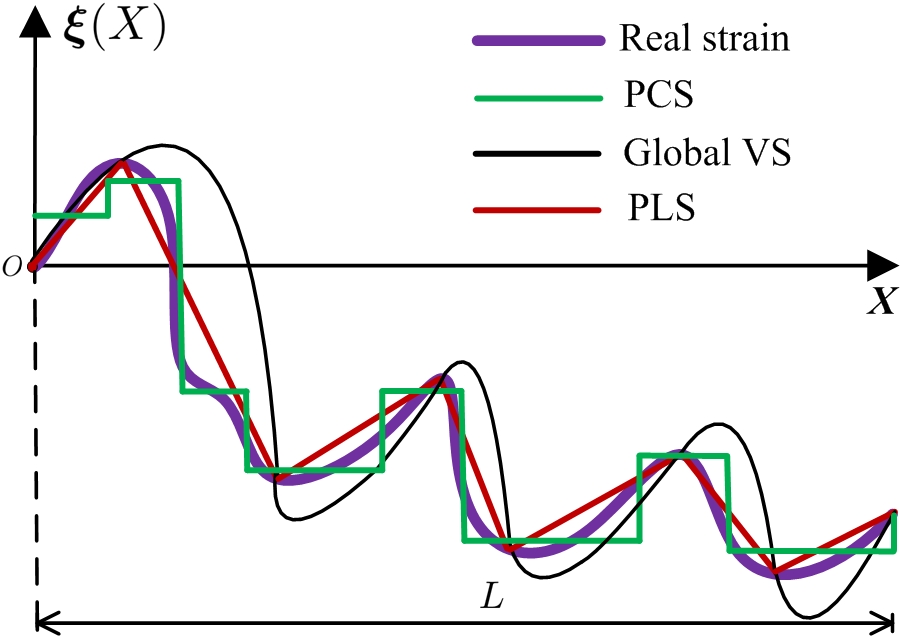}
	\caption{Sketch of comparison among VS, PCS and PLS Cosserat models with real strain.}
	\label{PLS_PCS_VS}
\end{figure}

	Given an arbitrary strain field of soft body subject to external forces, as described in Fig.~\ref{PLS_PCS_VS}, the PCS, VS and PLS share the common objective: to approximate the real strain field as precise as possible.
	PCS (proposed in \cite{renda2018discrete}) adopts a local approximation scheme with local constant strain assumption, while VS (proposed in \cite{boyer2020dynamics}) uses a global approximation manner via the chosen basis functions (polynomials for example). Clearly,
	PCS will result in high-dimensional system in order to reach a certain precision while 
	VS will suffer from lower precision near the local mutation region even the number of basis function is increased. 
	Nevertheless, PLS combines both advantages of PCS and VS, and proposes to locally use a linear strain to approximate the real one. Such a scheme is able to locally approximate to the strain field of soft body compared to VS, and get much less DoFs than PCS. The sketch described in Fig. \ref{PLS_PCS_VS} intuitively demonstrates that PLS has advantages over those proposed approaches in some engineering applications, and thus provides significance for authors to carry out this work.

\begin{figure*}[!t]
	\centering
	\subfloat[]{\includegraphics[width=2.5in]{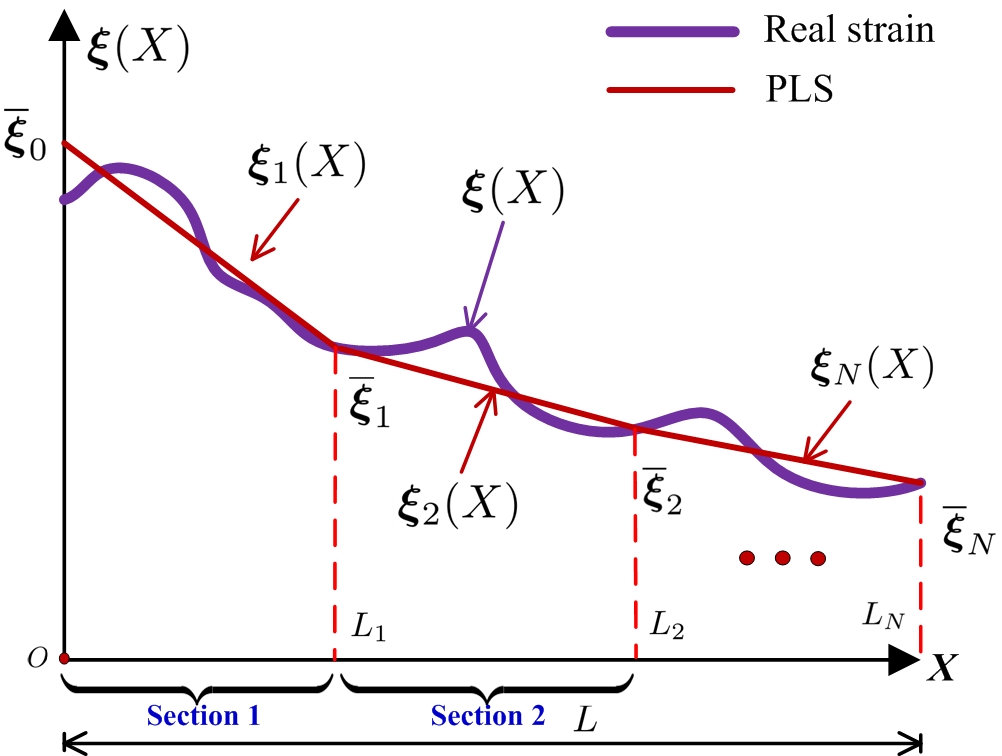}%
		\label{section}}
	\hfil
	\subfloat[]{\includegraphics[width=2.5in]{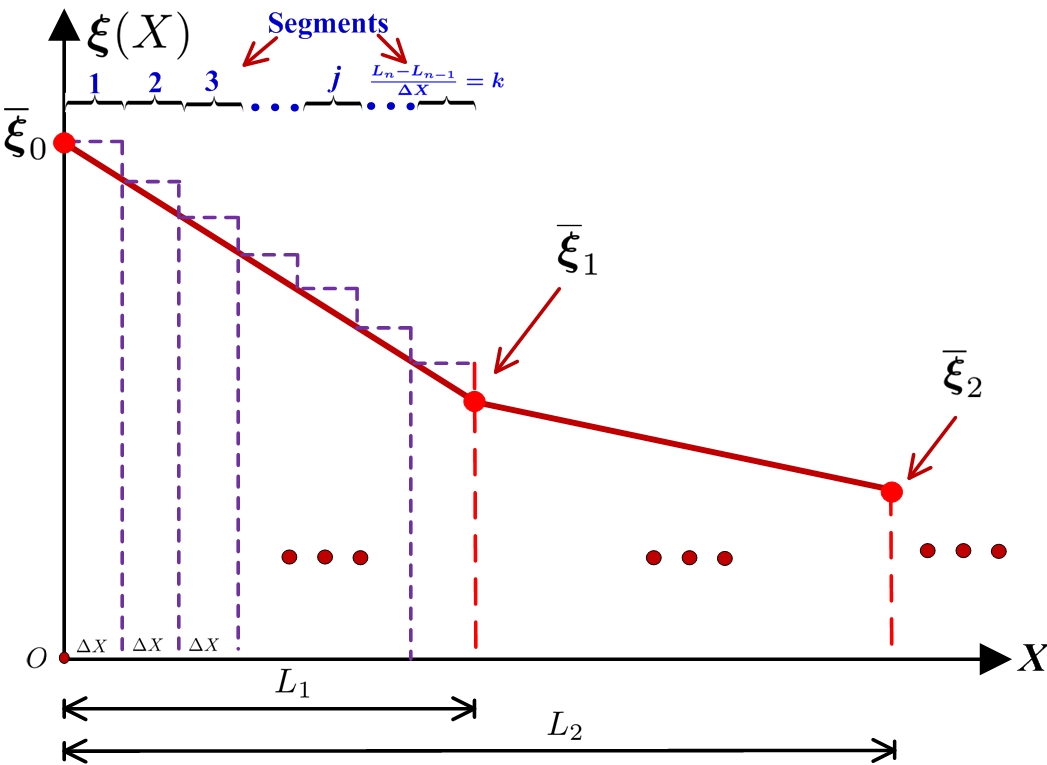}%
		\label{segment}}
	\caption{Schematic illustration of the PLS Cosserat model. (a) Soft arm divided into several sections. (b) Any section subdivided into quite a few segments.}
	\label{section_segment}
\end{figure*}

In what follows, we will focus on the PLS Cosserat modeling principle. First of all, it is noted that a ‘‘section’’ is defined as a unit block which is able to produce independent mechanical deformation while ‘‘segments’’ are a subset of one section. To put it differently, one section is made up of quite a few segments, as illustrated in Fig.~\ref{section_segment}. In view of the strain field $\boldsymbol{\xi}(X)$ along the soft manipulator varying with the arc length $X$ at any moment $t$, we divide the whole soft arm into $N$ variable length  continuum sections in the form of $\small[0, L_1]$, $[L_1,L_2]$ $\cdots$ $[L_{N-1}, L_N]$ (with $L_N=L$). Generally speaking, the continuous strain field $\boldsymbol{\xi}(X)$ is substituted for a finite set of $N$ continuous strain fields of the form  $\left\lbrace\small\boldsymbol{\xi}_1(X), \boldsymbol{\xi}_2(X),\cdots,\boldsymbol{\xi}_N(X)\right\rbrace $. 
To develop this discrete Cosserat model, for any continuum section $n$, we make the following two assumptions.
\begin{itemize}
	\item[$\bullet$] The strain twists $\small\overline{\boldsymbol{\xi}}_{n-1}$ and $\small \overline{\boldsymbol{\xi}}_n$ respectively corresponds to those at the proximal and distal ends of any section $n$, and other strain twists along the section $n$ linearly vary with $X$,  as shown in Fig.~\ref{section_segment}\subref{section}. Based on this linear assumption, the principle of the piecewise linear strain (PLS) can be then formulated as $$\small\boldsymbol{\xi}_n(X)=\overline{\boldsymbol{\xi}}_{n-1}\frac{L_n-X}{L_n-L_{n-1}}+ \overline{\boldsymbol{\xi}}_n\frac{X-L_{n-1}}{L_n-L_{n-1}} $$
for $X\in[L_{n-1}, L_n]$. 
\end{itemize}

\begin{itemize}
	\item[$\bullet$] Since the geometric and differential kinematic models are still linear time-varying systems under the PLS assumption, the section $n$ is subdivided into $k$ ($k\in\mathbb{R}$) infinitesimal segments of the form $\small[L_{n-1},L_{n-1}+\Delta X], [L_{n-1}+\Delta X, L_{n-1}+2\Delta X], \cdots, [L_{n-1}+(k-1)\Delta X,L_{n-1}+k\Delta X]$. In this way, the strain twists $\small\boldsymbol{\xi}_n(X)$ along the segment $j$ remain constant, i.e., $\small\boldsymbol{\xi}_n(X)\equiv\boldsymbol{\xi}_n(L_{n-1}+(j-1)\Delta X)$, $\small{X\in[L_{n-1}+(j-1)\Delta X,L_{n-1}+j\Delta X]}$. $\small\Delta X$ called one segment represents the infinitesimal distance between any adjacent cross sections, as displayed in Fig.~\ref{section_segment}\subref{segment}.
\end{itemize}

\subsection{PLS Cosserat: Geometric Model}

Based on the PLS assumptions, for any segment $j$ at time $t$, the system (\ref{geo}), (\ref{kine1}) and (\ref{kine2}) can be seen as liner time-invariant systems, thus they can be analytically solved. In consequence, the initial configuration, velocity or acceleration of any segment $j$ depend on the rightmost values of the previous segment ($j-1$) along the section $n$. 

To guarantee the continuity, specifying the rightmost configuration $\small\boldsymbol{g}(L_{n-1})$ of the section $(n-1)$ as the initial value of the system (\ref{geo}) for section $n$, the rightmost configuration of any segment $j$ along the section $n$ at time $t$ can be recursively derived. 
Taking the rightmost configuration $\small\boldsymbol{g}(L_{n-1}+j\Delta X)$ of the segment $j$ as the initial value of system (\ref{geo}) for segment $(j+1)$, and considering piecewise linear strain along one section, the position and orientation of any cross section at $X$ along the section $n$ at time $t$ analytically yields
\begin{equation}
\small
\begin{split}
\boldsymbol{g}(X)
&=\boldsymbol{g}(L_{n-1})\left(  \prod_{i=0}^{j-1}e^{\Delta X\Theta_{ni}}\right)  e^{(X-L_{n-1}-j\Delta X)\Theta_{nj}}\\
&\triangleq \boldsymbol{g}(L_{n-1})\boldsymbol{g}_n(X)
\label{geometric_model}
\end{split}
\end{equation}
with $\small\Theta_{ni}=\alpha_{ni}  \widehat{\overline{\boldsymbol{\xi}}}_{n-1}+\beta_{ni} \widehat{\overline{\boldsymbol{\xi}}}_n$,  $\small\Theta_{nj}=\alpha_{nj}  \widehat{\overline{\boldsymbol{\xi}}}_{n-1}+\beta_{nj} \widehat{\overline{\boldsymbol{\xi}}}_n$, 
where $\alpha_{ni}=1- \frac{i\Delta X}{L_n-L_{n-1}}$, $\beta_{ni}=\frac{i\Delta X}{L_n-L_{n-1}}$, $\alpha_{nj}=1- \frac{j\Delta X}{L_n-L_{n-1}}$, $\beta_{nj}=\frac{j\Delta X}{L_n-L_{n-1}}$, $\small\boldsymbol{g}_n(X)$ stands for the position and orientation of any cross section at $X$ along the section $n$ w.r.t. the rightmost counterpart of the section ($n-1$). % 

Intuitively, compared to PCS modeling method where $\boldsymbol{g}(X)$ is only a function of one strain field, the model deduced from PLS in (\ref{geometric_model}) depends on both $\small\overline{\boldsymbol{\xi}}_{n-1}$ and  $\small\overline{\boldsymbol{\xi}}_n$ of a certain section $n$, and this is due to linear interpolation used for PLS.

%To satisfy computational accuracy and speed, Taylor series expansion of $\small\boldsymbol{g}_n(X)$ in (\ref{geometric_model}) is performed, and it yields
%\begin{equation}
%\small
%\begin{split}
%\boldsymbol{g}_n(X)&=\boldsymbol{g}_{n(j-1)} e^{(X-L_{n-1}-j\Delta X)\Theta_{nj}}\\&=\boldsymbol{g}_{n(j-1)}\left(  \boldsymbol{{\rm I}}_4+(X-L_{n-1}-j\Delta X)\Theta_{nj}\right.\\&\quad\left.+\frac{1}{\theta^2_n}\left(1-{\rm{cos}}\left((X-L_{n-1}-j\Delta X) \theta_n\right) \right)\Theta^2_{nj}\right.\\
%&\quad\left.+\frac{1}{\theta^3_n}\left((X-L_{n-1}-j\Delta X)\theta_n\right.\right.\\&\quad\left.\left.-{\rm{sin}}\left((X-L_{n-1}-j\Delta X)\theta_n \right)  \right)\Theta^3_{nj}\right)
%\label{121212}
%\end{split}
%\end{equation}
%where $\theta_n$ in (\ref{121212}) relates to the angular strain twist $\small\boldsymbol{K}_n$ of the section $n$, and $\small\theta^2_n=\boldsymbol{K}^{\rm T}_n\boldsymbol{K}_n$. 
%%For stress-free configuration of the section $n$, the equation (\ref{121212}) is equivalent to
%\begin{comment}
%$$
%\small
%\boldsymbol{g}_n(X)=\boldsymbol{g}_{nj}\left(  \boldsymbol{{\rm I}}_4+(X-L_{n-1}-j\Delta X)\Theta_{nj}\right) 
%$$
%\end{comment}

\subsection{PLS Cosserat: Differential Kinematic Model}

The differential kinematics aims to find the mapping between the velocity twist along the manipulator and the time derivatives of the system state representing deformation twist of the soft arm. Under PLS assumption, by integrating the system (\ref{kine1}), we can then analytically obtain the velocity of each cross section at time $t$ (the detailed calculation of $\small\boldsymbol{\eta}(X)$ is given in Appendix~\ref{analytic_kinematics}).
	
According to the Proposition 2.25 in \cite{hall2003lie} that $e^{{\rm {ad}}_{\boldsymbol{\Theta^{\vee}}}}$ is equivalent to ${\rm{Ad}}_{e^{\boldsymbol{\Theta}}}$, we know that the coefficient matrix of the velocity $\small\boldsymbol{\eta}(L_{n-1})$ of tip at $X=L_{n-1}$ of the section $n$ is the exponential in the adjoint representation of the Lie group transformation $\small\boldsymbol{g}_n(X)$, i.e., 
$$\small e^{-(X-L_{n-1}-j\Delta X){\rm{ad}}_{\Theta_{nj}^{\vee}}}( \prod_{i=0}^{j-1}e^{-{\rm ad}_{\Delta X\Theta_{ni}^{\vee}}}) ={\rm{Ad}}^{-1}_{\boldsymbol{g}_n(X)}
$$ 
Defining the coefficient matrices of $\small\dot{\overline{\boldsymbol{\xi}}}_{n-1}$ and $\small\dot{\overline{\boldsymbol{\xi}}}_n$ separately as $\small{\rm T}_{\boldsymbol{g}_{n1}(X)}$ and $\small{\rm T}_{\boldsymbol{g}_{n2}(X)}$, the analytical solution of the velocity in Appendix~\ref{analytic_kinematics} is then given by
\begin{equation}
\small
\boldsymbol{\eta}(X)={\rm {Ad}}^{-1}_{\boldsymbol{g}_n(X)}\boldsymbol{\eta}(L_{n-1})+{\rm T}_{\boldsymbol{g}_{n1}(X)}\dot{\overline{\boldsymbol{\xi}}}_{n-1}+{\rm T}_{\boldsymbol{g}_{n2}(X)}\dot{\overline{\boldsymbol{\xi}}}_{n} .
\label{kine}
\end{equation}
with
\begin{equation*}
\small
\begin{split}
{\rm {T}}_{\boldsymbol{g}_{n1}(X)}&= e^{-(X-L_{n-1}-j\Delta X){\rm{ad}}_{\Theta_{nj}^{\vee}}}\sum_{i=1}^{j}\Big[\Big(\prod_{\tau=i}^{j-1}e^{-{\rm ad}_{\Delta X\Theta_{n\tau}^{\vee}}}\Big) \\&\quad\int_{L_{n-1}+(i-1)\Delta X}^{L_{n-1}+i\Delta X} e^{(s-L_{n-1}-i\Delta X){\rm ad}_{\Theta_{n(i-1)}^{\vee}}} \alpha_{n(i-1)} {\rm d}s\Big]\\
&\quad+\int_{L_{n-1}+j\Delta X}^{X}e^{-(X-s){\rm {ad}}_{\Theta_{nj}^{\vee}}} \alpha_{nj}{\rm d}s,
\end{split}
\end{equation*}
\begin{equation*}
\small
\begin{split}
{\rm {T}}_{\boldsymbol{g}_{n2}(X)}&= e^{-(X-L_{n-1}-j\Delta X){\rm{ad}}_{\Theta_{nj}^{\vee}}}\sum_{i=1}^{j}\Big[\Big( \prod_{\tau=i}^{j-1}e^{-{\rm ad}_{\Delta X\Theta_{n\tau}^{\vee}}}\Big) \\&\quad\int_{L_{n-1}+(i-1)\Delta X}^{L_{n-1}+i\Delta X} e^{(s-L_{n-1}-i\Delta X){\rm ad}_{\Theta_{n(i-1)}^{\vee}}} \beta_{n(i-1)}{\rm d}s\Big]\\&\quad+\int_{L_{n-1}+j\Delta X}^{X}e^{-(X-s){\rm {ad}}_{\Theta_{nj}^{\vee}}} \beta_{nj}{\rm d}s,
\end{split}
\end{equation*} 
where the symbol $\vee$ is an operator about mapping a matrix into a vector.

From (\ref{kine}), if we know the strain twists ($\small\overline{\boldsymbol{\xi}}_{n-1}$ and $\small\overline{\boldsymbol{\xi}}_n$) and strain twists rates ($\small\dot{\overline{\boldsymbol{\xi}}}_{n-1}$ and $\small\dot{\overline{\boldsymbol{\xi}}}_n$) of tips at $X=L_{n-1}$ and $X=L_n$,  the velocity of any cross section at $X$ and time $t$ along the section $n$ can be recursively derived. %\hh{Intuitively, there is one more item $\small{\rm {T}}_{\boldsymbol{g}_{n2}(X)}$ in (\ref{kine}) compared to PCS modeling method due to linear interpolation used for PLS.}

Due to the same reason, the acceleration $\dot{\boldsymbol{\eta}}(X)$ of any cross section at $X$ of the section $n$ along the soft manipulator at time $t$ can be analytically computed. Considering linearly variable strain twists along a certain section $n$ and using the property of Lie algebra that ${\rm {ad}}_mn=-{\rm {ad}}_nm$ holds for any $m$ and $n$, the system~(\ref{kine2}) can be analytically solved, and the detailed derivation is given in Appendix~\ref{analytic_kinematics}. For the sake of simplicity, this analytical solution is re-formulated below in the compact way as
\begin{equation}
\small
\begin{split}
\dot{\boldsymbol{\eta}}(X)&={\rm {Ad}}^{-1}_{\boldsymbol{g}_n(X)}\dot{\boldsymbol{\eta}}(L_{n-1})+{\rm AD}_{\boldsymbol{g}_{n1}(X)}\dot{\overline{\boldsymbol{\xi}}}_{n-1}\\&\quad+{\rm AD}_{\boldsymbol{g}_{n2}(X)}\dot{\overline{\boldsymbol{\xi}}}_{n}+{\rm T}_{\boldsymbol{g}_{n1}(X)}\ddot{\overline{\boldsymbol{\xi}}}_{n-1}+{\rm T}_{\boldsymbol{g}_{n2}(X)}\ddot{\overline{\boldsymbol{\xi}}}_{n}. 
\label{acccc}
\end{split}
\end{equation}
with
\begin{equation*}
\small
\begin{split}
{\rm {AD}}&_{\boldsymbol{g}_{n1}(X)}= e^{-(X-L_{n-1}-j\Delta X){\rm{ad}}_{\Theta_{nj}^{\vee}}}\sum_{i=1}^{j}\Big[\Big( \prod_{\tau=i}^{j-1}e^{-{\rm ad}_{\Delta X\Theta_{n\tau}^{\vee}}}\Big) \\&\int_{L_{n-1}+(i-1)\Delta X}^{L_{n-1}+i\Delta X}e^{(s-L_{n-1}-i\Delta X){\rm ad}_{\Theta_{n(i-1)}^{\vee}}} {\rm{ad}}_{\boldsymbol{\eta}(s)}\alpha_{n(i-1)} {\rm d}s\Big]\\&+\int_{L_{n-1}+j\Delta X}^{X}e^{-(X-s){\rm {ad}}_{\Theta_{nj}^{\vee}}} {\rm{ad}}_{\boldsymbol{\eta}(s)}\alpha_{nj}{\rm d}s,
\end{split}
\end{equation*}
and
\begin{equation*}
\small
\begin{split}
{\rm {AD}}&_{\boldsymbol{g}_{n2}(X)}= e^{-(X-L_{n-1}-j\Delta X){\rm{ad}}_{\Theta_{nj}^{\vee}}}\sum_{i=1}^{j}\Big[\Big(\prod_{\tau=i}^{j-1}e^{-{\rm ad}_{\Delta X\Theta_{n\tau}^{\vee}}}\Big) \\&\int_{L_{n-1}+(i-1)\Delta X}^{L_{n-1}+i\Delta X}e^{(s-L_{n-1}-i\Delta X){\rm ad}_{\Theta_{n(i-1)}^{\vee}}} {\rm{ad}}_{\boldsymbol{\eta}(s)}\beta_{n(i-1)} {\rm d}s\Big]\\&+\int_{L_{n-1}+j\Delta X}^{X}e^{-(X-s){\rm {ad}}_{\Theta_{nj}^{\vee}}} {\rm{ad}}_{\boldsymbol{\eta}(s)}\beta_{nj}{\rm d}s.
\end{split}
\end{equation*}

Thus, we can use (\ref{acccc}) to calculate the acceleration of all cross sections along the section $n$ if knowing strain twists ($\small\overline{\boldsymbol{\xi}}_{n-1}$ and $\small\overline{\boldsymbol{\xi}}_n$), strain twist rates ($\small\dot{\overline{\boldsymbol{\xi}}}_{n-1}$ and $\small\dot{\overline{\boldsymbol{\xi}}}_n$) and rates of strain twist rate  ($\small\ddot{\overline{\boldsymbol{\xi}}}_{n-1}$ and  $\small\ddot{\overline{\boldsymbol{\xi}}}_n$) of tips at $X=L_{n-1}$ and $X=L_n$.

The relation between the velocity twist $\small\boldsymbol{\eta}(X)$ along the robot and the strain twists ($\small\overline{\boldsymbol{\xi}}_{n-1}$ and $\small \overline{\boldsymbol{\xi}}_{n}$), and another relation between the acceleration twist $\small\dot{\boldsymbol{\eta}}(X)$ and the strain twists ($\small\overline{\boldsymbol{\xi}}_{n-1}$ and $\small \overline{\boldsymbol{\xi}}_{n}$) as well as the rate of strain twists ($\small\dot{\overline{\boldsymbol{\xi}}}_{n-1}$ and $\small \dot{\overline{\boldsymbol{\xi}}}_{n}$) of the tips for the section $n$ ought to be illustrated in order to derive the subsequent PLS Cosserat dynamic model. Applying (\ref{kine}) from base to tip for all cross sections along the soft manipulator in the chain, we can obtain the mapping as the geometric Jacobian which is an essential tool to describe the differential kinematics and dynamics of the PLS Cosserat model. Defining $\small\boldsymbol{S}_{(\cdot)}\in\mathbb{R}^{6\times 6}$ and $\small\dot{\boldsymbol{S}}_{(\cdot)}\in\mathbb{R}^{6\times 6}$ as the components of the Jacobian matrix and its partial derivative in time $t$, the structural form of Jacobian and its derivative over the length of the soft arm can be separately expressed as  $\small\boldsymbol{J}(\overline{\boldsymbol{\xi}}_0,\overline{\boldsymbol{\xi}}_1, \cdots, \overline{\boldsymbol{\xi}}_{N-1},\overline{\boldsymbol{\xi}}_N,X)=[\boldsymbol{S}_0\quad\boldsymbol{S}_1\quad\boldsymbol{S}_2\quad\boldsymbol{S}_3\cdots\boldsymbol{S}_N ]\in\mathbb{R}^{6\times6(N+1)}$, and $\small\dot{\boldsymbol{J}}(\overline{\boldsymbol{\xi}}_0,\overline{\boldsymbol{\xi}}_1, \cdots, \overline{\boldsymbol{\xi}}_N,\dot{\overline{\boldsymbol{\xi}}}_0,\dot{\overline{\boldsymbol{\xi}}}_1, \cdots,\dot{\overline{\boldsymbol{\xi}}}_{N-1}, \dot{\overline{\boldsymbol{\xi}}}_N,X)=[\dot{\boldsymbol{S}}_0\quad\dot{\boldsymbol{S}}_1\quad\dot{\boldsymbol{S}}_2\quad\dot{\boldsymbol{S}}_3\cdots\dot{\boldsymbol{S}}_N ]\in\mathbb{R}^{6\times6(N+1)}$.

The geometric Jacobian represents the relationship between the velocity twists of the soft manipulator and the time derivative of the deformations. The joint position vector $\small\boldsymbol{q}=[\begin{matrix}
	\overline{\boldsymbol{\xi}}^{\rm T}_0& \overline{\boldsymbol{\xi}}^{\rm T}_1& \overline{\boldsymbol{\xi}}^{\rm T}_2& \cdots& \overline{\boldsymbol{\xi}}^{\rm T}_{N}
	\end{matrix}]^{\rm T}\in\mathbb{R}^{6(N+1)}$ composed of strain twists of all linear interpolation nodes is introduced, and the manipulator like a cantilever rod selected as an objective is fixed to a mobile base (the velocity twist of the fixed end $\boldsymbol{\eta}(0)=\boldsymbol{0}$). As a result, the discrete model of velocity (\ref{kine}) is globally equivalent to
\begin{equation}
\small
\boldsymbol{\eta}(X)=\boldsymbol{J}(\boldsymbol{q},X)\dot{\boldsymbol{q}}.
\label{ve}
\end{equation}
where the analytical formula of $\boldsymbol{J}(\boldsymbol{q},X)$ can be found in (\ref{Jacobian}) of Appendix~\ref{Jaxobian_and_its_derivative}.
Finally, by taking the derivative of (\ref{ve}) with respect to time $t$, the acceleration twist $\dot{\boldsymbol{\eta}}(X)$ arrives at
\begin{equation}
\small
\dot{\boldsymbol{\eta}}(X)=\boldsymbol{J}(\boldsymbol{q},X)\ddot{\boldsymbol{q}}+\dot{\boldsymbol{J}}(\boldsymbol{q},\dot{\boldsymbol{q}},X)\dot{\boldsymbol{q}}.
\label{ac}
\end{equation}
and the analytical formula of $\dot{\boldsymbol{J}}(\boldsymbol{q},\dot{\boldsymbol{q}},X)$ is detailed in 
(\ref{Jacobian_deri}) of Appendix~\ref{Jaxobian_and_its_derivative}.

We would like to emphasize that, compared to PCS method, the deduced Jacobian matrix and its derivative in time are significantly different, and this is due to the linear interpolation scheme used in PLS method.
%
%
%
%\hh{The strain field is discretized by linear interpolation method, \hh{which results in the Jacobian matrix and its derivative in time significantly different from those of PCS,} as respectively represented by (\ref{Jacobian}) and (\ref{Jacobian_deri}) in Appendix~\ref{Jaxobian_and_its_derivative}}.

\subsection{PLS Cosserat: Dynamic model}
In order to deduce PLS Cosserat rod dynamics model corresponding to the PDE (\ref{strong_form}), we introduce the relation between virtual displacement and the state vector $\small\delta\boldsymbol{\phi}(X)=\boldsymbol{J}(\boldsymbol{q},X)\boldsymbol{\mathcal{P}}\delta\boldsymbol{q}_a$ based on the fact that the strain twist $\small \overline{\boldsymbol{\xi}}_N$ of the free end of the soft arm is constrained by the boundary condition. Substituting differential kinematics models (\ref{ve}) and (\ref{ac}) into the weak form of (\ref{strong_form}), a nonlinear ODE can be then obtained. Considering that the resulted ODE holds for $\small\forall \delta\boldsymbol{q}^{\rm T}_a\neq\boldsymbol{0}$, the generalized dynamics for the PLS Cosserat model yields
\setcounter{equation}{15} % equation counter
\begin{equation}
\scriptsize
\begin{split}
&\left(\boldsymbol{\mathcal{P}}^{\rm T}\int_{0}^{L_N}\boldsymbol{J}^{\rm T}\boldsymbol{\mathcal{M}}\boldsymbol{J}{\rm d}X\right) \ddot{\boldsymbol{q}}-\left[\boldsymbol{\mathcal{P}}^{\rm T} \int_{0}^{L_N}\boldsymbol{J}^{\rm T}\left({\rm ad}^{\rm{T}}_{\boldsymbol{J}\dot{\boldsymbol{q}}}\boldsymbol{\mathcal{M}}\boldsymbol{J}-\boldsymbol{\mathcal{M}}\dot{{\boldsymbol{J}}}\right) {\rm d}X\right] \dot{\boldsymbol{q}}\\
&=\boldsymbol{\mathcal{P}}^{\rm T}\int_{0}^{L_N}\boldsymbol{J}^{\rm T}\left( \boldsymbol{\mathcal{F}}'_{ie}-{\rm{ad}}^{\rm{T}}_{\boldsymbol{{\xi}}}\boldsymbol{\mathcal{F}}_{ie}\right){\rm d}X+\boldsymbol{\mathcal{P}}^{\rm T}\int_{0}^{L_N}\boldsymbol{J}^{\rm T}\overline{\boldsymbol{\mathcal{F}}}_e{\rm d}X
\\&\quad+\boldsymbol{\mathcal{P}}^{\rm T}\int_{0}^{L_N}\boldsymbol{J}^{\rm T}\left( \boldsymbol{\mathcal{F}}'_{ia}-{\rm{ad}}^{\rm{T}}_{\boldsymbol{{\xi}}}\boldsymbol{\mathcal{F}}_{ia}\right){\rm d}X
\label{nolinear_ODE}
\end{split}
\end{equation}
with $\small\boldsymbol{\mathcal{P}}=\left[ \begin{matrix}
\mathbf{I}_{6N\times6N}\\ \mathbf{0}_{6\times6N}
\end{matrix}\right]$, and $\small\delta\boldsymbol{q}_a=\left[\begin{matrix}
\delta\overline{\boldsymbol{\xi}}^{\rm T}_0&\delta\overline{\boldsymbol{\xi}}^{\rm T}_1&\cdots&\delta\overline{\boldsymbol{\xi}}^{\rm T}_{N-1}
\end{matrix} \right]^{\rm T}\in\mathbb{R}^{6N}$. %where $\small\boldsymbol{q}_a$ represents

Let us define the following generalized coefficient matrices and wrenches from (\ref{nolinear_ODE}):
\begin{itemize}
\item[$\bullet$]
$\small\boldsymbol{\mathcal{P}}^{\rm T}\int_{0}^{L_N}\boldsymbol{J}^{\rm T}\boldsymbol{\mathcal{M}}\boldsymbol{J}{\rm d}X=\boldsymbol{M}(\boldsymbol{q})$, the $6N\times6(N+1)$ mass matrix.
\end{itemize}
\begin{itemize}
\item[$\bullet$]$\small-\boldsymbol{\mathcal{P}}^{\rm T} \int_{0}^{L_N}\boldsymbol{J}^{\rm T}\left({\rm ad}^{\rm{T}}_{\boldsymbol{J}\dot{\boldsymbol{q}}}\boldsymbol{\mathcal{M}}\boldsymbol{J}-\boldsymbol{\mathcal{M}}\dot{{\boldsymbol{J}}}\right) {\rm d}X=\small\boldsymbol{C}(\boldsymbol{q}, \dot{{\boldsymbol{q}}})$, the $6N\times6(N+1)$ Coriolis matrix.
\end{itemize}
\begin{itemize}
	\item[$\bullet$]$\small\boldsymbol{\mathcal{P}}^{\rm T}\int_{0}^{L_N}\boldsymbol{J}^{\rm T}\left( \boldsymbol{\mathcal{F}}'_{ie}-{\rm{ad}}^{\rm{T}}_{\boldsymbol{{\xi}}}\boldsymbol{\mathcal{F}}_{ie}\right){\rm d}X=\boldsymbol{F}_{\rm i}(\boldsymbol{q},\dot{\boldsymbol{q}})$, the $6N\times1$ internal wrench.
\end{itemize}
\begin{itemize}
\item[$\bullet$]$\small\boldsymbol{\mathcal{P}}^{\rm T}\int_{0}^{L_N}\boldsymbol{J}^{\rm T}\overline{\boldsymbol{\mathcal{F}}}_e{\rm d}X=\boldsymbol{F}_e(\boldsymbol{q})$, the $6N\times1$ external wrench.
\end{itemize}
\begin{itemize}
\item[$\bullet$]$ \small\boldsymbol{\mathcal{P}}^{\rm T}\int_{0}^{L_N}\boldsymbol{J}^{\rm T}\left( \boldsymbol{\mathcal{F}}'_{ia}-{\rm{ad}}^{\rm{T}}_{\boldsymbol{{\xi}}}\boldsymbol{\mathcal{F}}_{ia}\right){\rm d}X=\boldsymbol{F}_a(\boldsymbol{q})$, the $6N\times1$ actuation wrench.
\end{itemize}

Thanks to the definition of the generalized internal wrench and PLS assumption, the generalized stiffness and viscosity matrices can be decoupled from $\small \boldsymbol{F}_{\rm{i}}(\boldsymbol{q}, \dot{{\boldsymbol{q}}})$ (see simplification detailed in Appendix \ref{internal_simplification}). The concise formulation can be then written as
\begin{equation}
\small
\begin{split}
\boldsymbol{F}_{\rm{i}}(\boldsymbol{q}, \dot{{\boldsymbol{q}}})=\boldsymbol{K}(\boldsymbol{q})(\boldsymbol{q}-\boldsymbol{q}_0)+\boldsymbol{D}(\boldsymbol{q})\dot{\boldsymbol{q}}
\label{dy4}
\end{split}
\end{equation}
where $\small\boldsymbol{K}(\boldsymbol{q})\in\mathbb{R}^{6N\times6(N+1)}$ is generalized stiffness matrix,  $\small\boldsymbol{D}(\boldsymbol{q})\in\mathbb{R}^{6N\times6(N+1)}$ is generalized viscosity matrix, and $\small\boldsymbol{q}_0=[\boldsymbol{\xi}^{\rm T}_{00},\boldsymbol{\xi}^{\rm T}_{10},\boldsymbol{\xi}^{\rm T}_{20},\cdots,\boldsymbol{\xi}^{\rm T}_{N0}]^{\rm T}\in\mathbb{R}^{6(N+1)}$ represents the initial configuration of the robot.

Aside from (\ref{nolinear_ODE}), the boundary condition in (\ref{BC}) must be considered. Therefore, the PLS Cosserat dynamic system can be given by
\begin{equation}
\small
\begin{split}
\boldsymbol{M}(\boldsymbol{q})\ddot{\boldsymbol{q}}+&\boldsymbol{C}(\boldsymbol{q}, \dot{{\boldsymbol{q}}})\dot{{\boldsymbol{q}}}-\boldsymbol{K}(\boldsymbol{q})(\boldsymbol{q}-\boldsymbol{q}_0)-\boldsymbol{D}(\boldsymbol{q})\dot{\boldsymbol{q}}-\boldsymbol{F}_e(\boldsymbol{q})=\boldsymbol{F}_a(\boldsymbol{q})\\
&\boldsymbol{\Gamma}\dot{\boldsymbol{q}}+\boldsymbol{\sigma}(\boldsymbol{q}-\boldsymbol{q}_0)-\boldsymbol{\mathcal{F}}_e(L_N)=-\boldsymbol{\mathcal{F}}_{ia}(L_N)
\label{dynamic_ode}
\end{split}
\end{equation}
where the second equation in (\ref{dynamic_ode}) is derived from the boundary condition (\ref{BC}) and the constitutive law (\ref{inter}), with $\small\boldsymbol{\Gamma}=[\boldsymbol{0}_{6\times6N}, \boldsymbol{\gamma}(L_N)]$, and  $\small\boldsymbol{\sigma}=[\boldsymbol{0}_{6\times6N}, \boldsymbol{\Sigma}(L_N)]$. 

Compared to PCS dynamic model where the mass matrix $\small\boldsymbol{M}(\boldsymbol{q})$ is square, the deduced PLS dynamic model has a non-square $\small\boldsymbol{M}(\boldsymbol{q})$. However, by \hh{complementing} with the boundary condition (\ref{BC}) and the constitutive law (\ref{inter}), a similar Lagrangian model can be obtained. 
If the viscosity is not considered in (\ref{inter}), i.e., $ \boldsymbol{\gamma} = 0$, then it leads to an algebraic equation in (\ref{dynamic_ode}), which in fact is a differential-algebraic system. This characteristic, distinguish with PCS and VS, is exactly due to the PLS assumption.

\subsection{Strain Mode Choice Scheme via the PLS Cosserat}
According to the Cosserat rod theory, the strain twist of any interpolation node $\small\overline{\boldsymbol{\xi}}_i$ could take any value in the six dimensional components. However, the internal rod kinematics describing the motions between the cross sections can be constrained by some restrictions in the usual application of soft robots. To tackle this restriction, by following the similar idea of \cite{boyer2020dynamics}, we decompose the strain twist of any interpolation node $i$ as
\begin{equation}
\small
\overline{\boldsymbol{\xi}}_i=\mathbf{B}_a\boldsymbol{\xi}^*_{ia}+\mathbf{B}_c\boldsymbol{\xi}^*_{ic}
\label{reduced_strain}
\end{equation}
where $\small\boldsymbol{\xi}^*_{ia}\in\boldsymbol{\mathbb{R}}^{n_i}$ determines the vector field of the free strains (i.e., the number of DoFs $n_i$) of the interpolation nodes allowed by the rod kinematics, $\small\boldsymbol{\xi}^*_{ic}\in\boldsymbol{\mathbb{R}}^{6-n_i}$ represents the vector field of constrained strains, $\small\boldsymbol{\rm B}_a$ and $\small\boldsymbol{\rm B}_c$ stand for the complementary selection matrix of $1$ and $0$ such that $\small\boldsymbol{\rm B}_a^{\rm T}{\mathbf{B}}_a={\mathbf{I}}_{(n_i)\times(n_i)}$, $\small\boldsymbol{\rm B}_c^{\rm T}{\mathbf{B}}_c={\mathbf{I}}_{(6-n_i)\times(6-n_i)}$, and $\small\boldsymbol{\rm B}_a^{\rm T}{\mathbf{B}}_c=\boldsymbol{0}$.

Substituting (\ref{reduced_strain}) into (\ref{geometric_model}), the reduced geometric model can be obtained. For the constrained soft manipulator, it is necessary to consider the strain states of all interpolation nodes. Hence, the generalized joint position vector can be expressed as
\begin{equation}
\small
\boldsymbol{q}=\overline{\mathbf{B}}_a\overline{\boldsymbol{q}}+\overline{\mathbf{B}}_c\underline{\boldsymbol{q}}
\label{q_reduced_1}
\end{equation}
with $\small\overline{\boldsymbol{\rm B}}_a=\mathbf{I}_{(N+1)\times(N+1)}\otimes\mathbf{B}_a$, $\small\overline{\boldsymbol{\rm B}}_c=\mathbf{I}_{(N+1)\times(N+1)}\otimes\mathbf{B}_c$, where $\otimes$ represents the Kronecker product. In such a way,
$\small\overline{\boldsymbol{\rm B}}_a\in\boldsymbol{\mathbb{R}}^{6(N+1)\times[n_i(N+1)]}$ is the generalized selection matrix for the allowed states, $\small\overline{\boldsymbol{q}}=\left[\begin{matrix}
\boldsymbol{\xi}^*_{0a}&\boldsymbol{\xi}^*_{1a}&\boldsymbol{\xi}^*_{2a}&\cdots&\boldsymbol{\xi}^*_{Na}
\end{matrix} \right] \in\boldsymbol{\mathbb{R}}^{n_i(N+1)}$ includes the allowed DoFs of the arm. 

For the PLS Cosserat model reduction, the internal elastic wrench should be divided into two parts: one is constrained wrench in charge of imposing the internal constraints for prohibited strains, another is the elastic wrench related to the allowed DoFs. Thus, the reduced internal wrench is given by
\begin{equation}
\scriptsize
\hh{\boldsymbol{\mathcal{F}}_{ie}=\underbrace{\boldsymbol{\Sigma}(X)\mathbf{B}_a(\boldsymbol{\xi}^*_{a}(X)-\boldsymbol{\xi}^*_{i0})+\boldsymbol{\gamma}(X)\mathbf{B}_a\dot{\boldsymbol{\xi}}^*_{a}(X)}_{\boldsymbol{\mathcal{F}}^*_{ie}}+\mathbf{B}_c\boldsymbol{\lambda}(X)
\label{reduced_internal}}
\end{equation}
\hh{where $\small\boldsymbol{\xi}^*_{a}(X)$ for $\small X\in[L_{n-1},L_n]$ can be obtained by the linear interpolation of allowed strains of the adjacent nodes, $\small\boldsymbol{\xi}^*_{i0}$ is the initial states of allowed strains, $\small\boldsymbol{\lambda}(X)\in\mathbb{R}^{(6-n_i)}$ is the constrained wrench.}

Inserting (\ref{q_reduced_1}) into (\ref{ve}) and (\ref{ac}), the reduced PLS Cosserat differential kinematics models yields
\begin{equation*}
\small
\begin{split}
\boldsymbol{\eta}(X)&=\overline{\boldsymbol{J}}(\overline{\boldsymbol{q}},X)\dot{\overline{\boldsymbol{q}}}\\
\dot{\boldsymbol{\eta}}(X)&=\overline{\boldsymbol{J}}(\overline{\boldsymbol{q}},X)\ddot{\overline{\boldsymbol{q}}}+\dot{\overline{\boldsymbol{J}}}(\overline{\boldsymbol{q}},\dot{\overline{\boldsymbol{q}}},X)\dot{\overline{\boldsymbol{q}}}
\end{split}
\end{equation*}
where $\small \overline{\boldsymbol{J}}(\overline{\boldsymbol{q}},X)= \boldsymbol{J}(\boldsymbol{q},X)\overline{\boldsymbol{\rm B}}_a\in\boldsymbol{\mathbb{R}}^{6\times[n_i(N+1)]}$ is the reduced body Jacobian matrix. Using the relation $\small\delta\boldsymbol{\phi}(X)=\overline{\boldsymbol{J}}(\overline{\boldsymbol{q}},X)\overline{\boldsymbol{\mathcal{P}}}\delta\overline{\boldsymbol{q}}_a$, and substituting the reduced kinematics relations as well as (\ref{reduced_internal}) into the weak form of (\ref{strong_form}) lead to
\begin{equation}
\scriptsize
\begin{split}
&\underbrace{\left(\overline{\boldsymbol{\mathcal{P}}}^{\rm T}\int_{0}^{L_N}\overline{\boldsymbol{J}}^{\rm T}\boldsymbol{\mathcal{M}}\overline{\boldsymbol{J}}{\rm d}X\right)}_{\overline{\boldsymbol{M}}(\overline{\boldsymbol{q}})} \ddot{\overline{\boldsymbol{q}}}-\underbrace{\left[\overline{\boldsymbol{\mathcal{P}}}^{\rm T} \int_{0}^{L_N}\overline{\boldsymbol{J}}^{\rm T}\left({\rm ad}^{\rm{T}}_{\overline{\boldsymbol{J}}\dot{\overline{\boldsymbol{q}}}}\boldsymbol{\mathcal{M}}\overline{\boldsymbol{J}}-\boldsymbol{\mathcal{M}}\dot{\overline{{\boldsymbol{J}}}}\right) {\rm d}X\right]}_{\overline{\boldsymbol{C}}(\overline{\boldsymbol{q}},\dot{\overline{\boldsymbol{q}}})} \dot{\overline{\boldsymbol{q}}}\\
&=\underbrace{\overline{\boldsymbol{\mathcal{P}}}^{\rm T}\int_{0}^{L_N}\overline{\boldsymbol{J}}^{\rm T}\left( \boldsymbol{\mathcal{F}}^{*'}_{ie}-{\rm{ad}}^{\rm{T}}_{\boldsymbol{{\xi}}}\boldsymbol{\mathcal{F}}^*_{ie}\right){\rm d}X}_{\overline{\boldsymbol{F}}_{\rm i}(\overline{\boldsymbol{q}},\dot{\overline{\boldsymbol{q}}})}+\underbrace{\overline{\boldsymbol{\mathcal{P}}}^{\rm T} \int_{0}^{L_N}\overline{\boldsymbol{J}}^{\rm T}\overline{\boldsymbol{\mathcal{F}}}_e{\rm d}X}_{\overline{\boldsymbol{F}}_e(\overline{\boldsymbol{q}})}\\
&\quad+\underbrace{\overline{\boldsymbol{\mathcal{P}}}^{\rm T}\int_{0}^{L_N}\overline{\boldsymbol{J}}^{\rm T}\left( \boldsymbol{\mathcal{F}}'_{ia}-{\rm{ad}}^{\rm{T}}_{\boldsymbol{{\xi}}}\boldsymbol{\mathcal{F}}_{ia}\right){\rm d}X}_{\overline{\boldsymbol{F}}_a(\overline{\boldsymbol{q}})}\\
&\quad+\underbrace{\overline{\boldsymbol{\mathcal{P}}}^{\rm T}\int_{0}^{L_N}\overline{\boldsymbol{J}}^{\rm T}\left(\mathbf{B}_c \boldsymbol{\lambda}'-{\rm{ad}}^{\rm{T}}_{\boldsymbol{{\xi}}}\mathbf{B}_c\boldsymbol{\lambda}\right){\rm d}X}_{\overline{\boldsymbol{F}}_{\lambda}(\overline{\boldsymbol{q}})}
\label{new_dy}
\end{split}
\end{equation}
where $\small\overline{\boldsymbol{\mathcal{P}}}\in\mathbb{R}^{n_i(N+1)\times n_iN}$ represents a selection matrix,  $\small\overline{\boldsymbol{q}}_a=\left[\begin{matrix}
\boldsymbol{\xi}^*_{0a}&\boldsymbol{\xi}^*_{1a}&\boldsymbol{\xi}^*_{2a}&\cdots&\boldsymbol{\xi}^*_{(N-1)a}
\end{matrix} \right] \in\boldsymbol{\mathbb{R}}^{n_iN}$ is composed of allowed DoFs of all strain nodes except those of the free end.

%In (\ref{new_dy}), the reduced generalized internal wrench can be simplified as
%\begin{equation*}
%\small
%\boldsymbol{F}_{\rm i}(\overline{\boldsymbol{q}},\dot{\overline{\boldsymbol{q}}})=\overline{\boldsymbol{K}}(\overline{\boldsymbol{q}})(\overline{\boldsymbol{q}}-\overline{\boldsymbol{q}}_0)+\overline{\boldsymbol{D}}(\overline{\boldsymbol{q}})\dot{\overline{\boldsymbol{q}}}
%\end{equation*}
%with $\small\overline{\boldsymbol{q}}_0=\overline{\mathbf{B}}^{\rm T}(\boldsymbol{q}_0-\underline{\boldsymbol{q}})$. where the reduced generalized stiffness matrix $\small\overline{\boldsymbol{K}}(\overline{\boldsymbol{q}})\in\boldsymbol{\mathbb{R}}^{(n_iN)\times[n_i(N+1)]}$ and viscosity matrix $\small\overline{\boldsymbol{D}}(\overline{\boldsymbol{q}})\in\boldsymbol{\mathbb{R}}^{(n_iN)\times[n_i(N+1)]}$, which can be obtained if substituting (\ref{reduced_strain}) into  $\small\overline{\boldsymbol{F}}_{\rm i}(\overline{\boldsymbol{q}},\dot{\overline{\boldsymbol{q}}})$ in (\ref{new_dy}).
In addition, it is worth noting that the reduced boundary condition should be formulated as
\begin{subequations}\label{reduced_BCs}
	\small
	\begin{equation}
	\begin{split}
	\mathbf{B}^{\rm T}_a\boldsymbol{\Sigma}(L_N)\mathbf{B}_a(\boldsymbol{\xi}^*_{Na}&-\boldsymbol{\xi}^*_{N0})+\mathbf{B}^{\rm T}_a\boldsymbol{\gamma}(L_N)\mathbf{B}_a\dot{\boldsymbol{\xi}}_{Na}^*\\
	&=\mathbf{B}^{\rm T}_a\big(-\boldsymbol{\mathcal{F}}_{ia}(L_N)+\boldsymbol{\mathcal{F}}_e(L_N)\big)
	\end{split}
	\label{allowed}
	\end{equation}
	\begin{equation}
	\boldsymbol{\lambda}(L_N)=\boldsymbol{\rm B}_c^{\rm T}\big(-\boldsymbol{\mathcal{F}}_{ia}(L_N)+\boldsymbol{\mathcal{F}}_e(L_N)\big)
	\label{constrained}
	\end{equation}
\end{subequations}
%\begin{equation}
%\boldsymbol{\mathcal{M}} \dot{\boldsymbol{\eta}}+\operatorname{ad}_{\boldsymbol{\eta}}^{*} \boldsymbol{\mathcal{M}} \boldsymbol{\eta}=\boldsymbol{\mathcal{F}}_{i}^{\prime}+\operatorname{ad}_{\boldsymbol{\xi}}^{*} \boldsymbol{\mathcal{F}}_{i}+\boldsymbol{\mathcal{F}}_{e}
%\label{internal_force_with constraint}
%\end{equation}

It must be also pointed out that all the items except the last one in (\ref{new_dy}) can be obtained by replacing the Jacobian matrix $\small\boldsymbol{J}$ in (\ref{nolinear_ODE}) with $\small \boldsymbol{J}\overline{\boldsymbol{\rm B}}_a$. The following part concerns the calculation of the item with the constrained wrench.
\begin{theorem}
For the PLS Cosserat model with full modes, if the strain field is re-formulated as $\small\boldsymbol{\xi}(X)=\boldsymbol{\Phi}(X)\boldsymbol{q}(t)$, where $\small\boldsymbol{\Phi}(X)\in\mathbb{R}^{6\times6(N+1)}$ is a matrix comprised of coefficient of strain interpolation nodes via the PLS assumption, then there exists a relationship among three quantities (i.e., $\small\overline{\mathbf{B}}_a,\ \boldsymbol{\Phi},\ {\rm {and}}\ \mathbf{B}_c$) satisfying the following equality
\begin{equation}
\scriptsize
\overline{\mathbf{B}}^{\rm T}_a\boldsymbol{\Phi}^\mathrm{T}\mathbf{B}_c=\left[\begin{matrix}
a_1(X)\mathbf{B}^{\mathbf{T}}_a\\\big(b_1(X)+a_2(X)\big)\mathbf{B}^{\mathbf{T}}_a\\\big(b_{n-1}(X)+a_n(X)\big)\mathbf{B}^{\mathbf{T}}_a\\\vdots\\b_N(X)\mathbf{B}^{\mathbf{T}}_a
\end{matrix} \right]\mathbf{B}_c=\boldsymbol{0}
\label{BaPhiBc}
\end{equation}
%\begin{equation}
%\small
%\overline{\boldsymbol{F}}_{\lambda}=\overline{\boldsymbol{\mathcal{P}}}^{\rm T}\int_{0}^{L_N}\overline{\boldsymbol{J}}^{\rm T}\left(\mathbf{B}_c \boldsymbol{\lambda}'-{\rm{ad}}^{\rm{T}}_{\boldsymbol{{\xi}}}\mathbf{B}_c\boldsymbol{\lambda}\right){\rm d}X
%\label{generalized_internal}
%\end{equation}
\end{theorem}
%with $\scriptsize\boldsymbol{\Phi}(X)=\left[ \begin{matrix}
%a_1(X)\mathbf{I}_{ni}&(b_1(X)+a_2(X))\mathbf{I}_{ni}&(b_2(X)+a_3(X))\mathbf{I}_{ni}&\cdots&b_N(X)\mathbf{I}_{ni}
%\end{matrix}\right] $,
with $\scriptsize a_n(X)=\frac{L_n-X}{L_n-L_{n-1}}$, and $\small b_n(X)=\frac{X-L_{n-1}}{L_n-L_{n-1}}$, then the generalized constrained wrench  $\small\overline{\boldsymbol{F}}_{\lambda}$ for the reduced PLS Cosserat relates to the constrained wrench of end cross section.
\begin{proof}
Note that we want to prove that $\small\overline{\boldsymbol{F}}_{\lambda}$ is only dependent of the constrained wrench at $X=L_N$. At this aim, let us insert (\ref{ve}) into (\ref{kine1}), and it yields
\begin{equation*}
	\small
	\boldsymbol{J}' \dot{\boldsymbol{q}}=-\operatorname{ad}_{\boldsymbol{\xi}} \boldsymbol{J} \dot{\boldsymbol{q}}+\dot{\boldsymbol{\xi}}(X)=-\operatorname{ad}_{\boldsymbol{\xi}} \boldsymbol{J} \dot{\boldsymbol{q}}+\boldsymbol{\Phi}(X)\dot{\boldsymbol{q}}
\end{equation*}
which holds for $\small\forall \dot{\boldsymbol{q}}\neq\boldsymbol{0}$, and thus
\begin{equation}
	\small
	\overline{\boldsymbol{J}}'=-{\rm{ad}}_{\boldsymbol{\xi}} \overline{\boldsymbol{J}} +\boldsymbol{\Phi}(X)\overline{\boldsymbol{\rm B}}_a
	\label{Jaco'}
\end{equation}

Substituting (\ref{Jaco'}) into $\small\overline{\boldsymbol{F}}_{\lambda}$ with unknown constrained wrench $\small\boldsymbol{\lambda}(X)$ in (\ref{new_dy}) then arrives at
\begin{equation}
	\scriptsize
	\begin{split}	
	&\overline{\boldsymbol{F}}_{\lambda}=\overline{\boldsymbol{\mathcal{P}}}^{\rm T}\int_{0}^{L_N}\left[ (\overline{\boldsymbol{J}}^\mathrm{T} \mathbf{B}_c\boldsymbol{\lambda})'-\overline{\boldsymbol{J}}'^{\rm{T}}\mathbf{B}_c\boldsymbol{\lambda}-\overline{\boldsymbol{J}}^\mathrm{T}{\rm{ad}}_{\boldsymbol{\xi}}^{\rm T} \mathbf{B}_c\boldsymbol{\lambda}\right] {\rm d}X\\
	&=
	\overline{\boldsymbol{\mathcal{P}}}^{\rm T}\int_{0}^{L_N}\left[  (\overline{\boldsymbol{J}}^\mathrm{T} \mathbf{B}_c\boldsymbol{\lambda})'+\overline{\boldsymbol{J}}^{\rm{T}}{\rm{ad}}_{\boldsymbol{\xi}}^\mathrm{T} \mathbf{B}_c\boldsymbol{\lambda}-\overline{\mathbf{B}}_a^{\rm T}\boldsymbol{\Phi}^\mathrm{T}\mathbf{B}_c\boldsymbol{\lambda}-\overline{\boldsymbol{J}}^\mathrm{T}{\rm{ad}}_{\boldsymbol{\xi}}^{\rm T} \mathbf{B}_c\boldsymbol{\lambda}\right] {\rm d}X\\
	&=
	\overline{\boldsymbol{\mathcal{P}}}^{\rm T}\int_{0}^{L_N}(\overline{\boldsymbol{J}}^\mathrm{T} \mathbf{B}_c\boldsymbol{\lambda})'{\rm d}X-\overline{\boldsymbol{\mathcal{P}}}^{\rm T}\int_{0}^{L_N}\overline{\mathbf{B}}^{\rm T}_a\boldsymbol{\Phi}^\mathrm{T}\mathbf{B}_c\boldsymbol{\lambda}{\rm d}X\nonumber
	\end{split}
\end{equation}

Clearly, the second item in above equation can be removed in accordance with (\ref{BaPhiBc}). Consequently, by using (\ref{constrained}), we obtain
\begin{equation*}
\scriptsize
\begin{split}
\overline{\boldsymbol{F}}_{\lambda}=
\overline{\boldsymbol{\mathcal{P}}}^{\rm T}(\overline{\boldsymbol{J}}^\mathrm{T} \mathbf{B}_c\boldsymbol{\lambda})|_0^{L_N}=\overline{\boldsymbol{\mathcal{P}}}^{\rm T}\overline{\boldsymbol{J}}^\mathrm{T}(L_N) \mathbf{B}_c\boldsymbol{\rm B}_c^{\rm T}\big(-\boldsymbol{\mathcal{F}}_{ia}(L_N)+\boldsymbol{\mathcal{F}}_e(L_N)\big).\blacksquare
\end{split}
\end{equation*}
\end{proof}
%The above result implies that $\small\overline{\boldsymbol{F}}_{\lambda}$ is only dependent of the constrained wrench at $X=L_N$.$\blacksquare$

By combining (\ref{new_dy}) and (\ref{reduced_BCs}), one can easily obtain the reduced model based on PLS Cosserat to model several simplified systems ($n_i\leq6$). An exhaustive reference of the reduced systems with the complementary selection matrices $\small\boldsymbol{\rm B}_a$ and $\small\boldsymbol{\rm B}_c$ for describing different systems is shown in Appendix \ref{selection_B}.

\section{Model Parameters Identification}
%The parameters identification aims to arrive at an appropriate input-output relationship of the model by combining information derived from the experiment with that obtained from the mechanical behavior of the model.

In practice, the exact values of the physical parameters are typically unknown or difficult to derive even for the soft robot manufacturers. 
Besides, even though there is full knowledge of the model and sufficient data available, an accurate description is most often not desirable. Therefore, it is necessary and important to identify the physical parameters for a soft robot conveniently and accurately. To reach this goal, we present an efficient algorithm framework to identify the parameters involved in the deduced PLS dynamic model.

Technologically speaking, the joint vector $\small\boldsymbol{q}$ is actually difficult to measure in the experiment. However, we can easily determine the position of the end-effector by using the position sensor. 
Therefore, the proposed identification scheme is based on the measurement of end-effector's position. To this aim, it is assumed that $\overline{N}$ sets of different experiments are effectuated and the objective is to seek optimal parameters to minimize the difference between the real measured end-effector position and that obtained from simulation, by satisfying of course the PLS Cosserat static model. Consequently, the parameter identification algorithm can be formulated by the following nonlinear programming (NLP) problem:
\begin{equation}
\small
\begin{split}
&\mathop{\arg\min}\limits_{\boldsymbol{\delta}=(\boldsymbol{\theta},\boldsymbol{q}_1,\boldsymbol{q}_2,\cdots,\boldsymbol{q}_{\overline{N}})}f(\boldsymbol{\delta})=\sum_{i=1}^{\overline{N}}\Vert\boldsymbol{u}_i-\boldsymbol{u}_{ei}\Vert_2\\
&
s.t. \quad
\begin{cases}
\mathbb{K}(\boldsymbol{\theta}, \boldsymbol{q}_1)-\mathbb{F}_a(\boldsymbol{q}_1)=\boldsymbol{0}\\
\mathbb{K}(\boldsymbol{\theta}, \boldsymbol{q}_2)-\mathbb{F}_a(\boldsymbol{q}_2)=\boldsymbol{0}\\
\qquad\qquad\qquad\vdots\\
\mathbb{K}(\boldsymbol{\theta}, \boldsymbol{q}_{\overline{N}})-\mathbb{F}_a(\boldsymbol{q}_{\overline{N}})=\boldsymbol{0}
\end{cases}
\label{NLP}
\end{split}
\end{equation}
with
$$
\small
\mathbb{K}(\boldsymbol{\theta},\boldsymbol{q}_i)=\left[ \begin{matrix}
-\boldsymbol{K}(\boldsymbol{q}_i)\\\boldsymbol{\sigma}
\end{matrix}\right](\boldsymbol{q}_i-\boldsymbol{q}_{0i})-\left[ \begin{matrix}
\boldsymbol{F}_e(\boldsymbol{q}_i)\\ \boldsymbol{\mathcal{F}}_e(L_N)
\end{matrix}\right],
$$
$$
\small
\mathbb{F}_a(\boldsymbol{q}_i)=\left[ \begin{matrix}
\boldsymbol{F}_a(\boldsymbol{q}_i)\\
-\boldsymbol{\mathcal{F}}_{ia}(L_N)
\end{matrix}\right] 
$$
where $\small\boldsymbol{u}_i=\boldsymbol{W}\boldsymbol{g}(\boldsymbol{q}_i,L)\boldsymbol{\Upsilon}$ implies the end-effector position provided by the PLS static model, with  $\small\boldsymbol{W}=\left[\begin{matrix}
\mathbf{I}_3&\boldsymbol{0}
\end{matrix}\right]$ and $\small\boldsymbol{\Upsilon}=\begin{matrix}\begin{bmatrix}
\boldsymbol{0}_3&1
\end{bmatrix}^{\rm T}
\end{matrix}$ in the $i^{\rm{th}}$ experiment, $\small\boldsymbol{g}(\boldsymbol{q}_i,L)$ stands for the position and orientation of the end-effector in the $i^{\rm{th}}$ experiment, with $\boldsymbol{q}_i$ being the strain vector, $\small\boldsymbol{\theta}=\left[\begin{matrix}
E&G&\rho
\end{matrix}\right] $ represents those parameters to be identified, including the Young's modulus $E$, shear modulus $G$ and density of material $\rho$, $\small\boldsymbol{u}_{ei}$ is the $i^{\rm{th}}$ experimental measurement of the end-effector position by using the sensor. 

To solve the above NLP problem, the Newton-type method is used by attempting to find the optimal solution $\small\boldsymbol{\delta}^*$ which can generally satisfy the Karush-Kuhn-Tucker (KKT) conditions that there exist multiplier vectors $\small\overline{\boldsymbol{\lambda}}^*\in\boldsymbol{\mathbb{R}}^{6\overline{N}(N+1)}$ such that the following equations hold:
\begin{align*}
\small
\bigtriangledown_{\boldsymbol{\delta}}\mathcal{L}(\boldsymbol{\delta}^*,\overline{\boldsymbol{\lambda}}^*)&=0\\
\overline{\boldsymbol{\lambda}}^*&\neq\boldsymbol{0}\\
\boldsymbol{\mathcal{H}}(\boldsymbol{\delta}^*)&=\boldsymbol{0}
\end{align*}
with $$\small\mathcal{L}=f(\boldsymbol{\delta})+\boldsymbol{\mathcal{H}}(\boldsymbol{\delta})^{\rm T}\overline{\boldsymbol{\lambda}},$$ 
and
$$\small\boldsymbol{\mathcal{H}}(\boldsymbol{\delta})=\left[ \begin{matrix}
\boldsymbol{\mathcal{H}}_1(\boldsymbol{\theta},\boldsymbol{q}_1)^{\rm T}&\boldsymbol{\mathcal{H}}_2(\boldsymbol{\theta},\boldsymbol{q}_2)^{\rm T}&\cdots&\boldsymbol{\mathcal{H}}_{\overline{N}}(\boldsymbol{\theta},\boldsymbol{q}_{\overline{N}})^{\rm T}
\end{matrix}\right]^{\rm T}$$  
where $\small \boldsymbol{\mathcal{H}}_i(\boldsymbol{\theta},\boldsymbol{q}_i)=\mathbb{K}(\boldsymbol{\theta}, \boldsymbol{q}_i)-\mathbb{F}_a(\boldsymbol{q}_i)$ is the static model in the $i^{\rm{th}}$ experiment, $\small\mathcal{L}$ represents the Lagrange function, and $\small\overline{\boldsymbol{\lambda}}$ is the Lagrangian multiplier vector.

%\begin{remark}
%	To guarantee convergence of the algorithm, the selection of the initial variable $\small\boldsymbol{\delta}_0$ composed of material parameters and configuration of the manipulator should be mentioned. Here we refer to the Young's modulus $E\rightarrow\infty$ and density $\rho\rightarrow0$ as the initial values of material parameters, and choose the undeformed reference straight shape of the arm as initial configuration, which conforms to the natural state of the cantilever rod with large stiffness. 
%\end{remark}

\begin{remark}
The Newton-type algorithm is sensitive to the choice of initial guess. Note that if a minimum exists, it is not necessarily unique. In other words, there may be an infinite number of feasible points that meet the KKT conditions and are thus minima. However, regardless of the number of local minima, there is always a unique optimal solution (if it exists). To obtain the material parameters of soft manipulator accurately and efficiently, the determination of the initial variable $\small\boldsymbol{\delta}_0$ composed of material parameters and configuration of the manipulator should be mentioned. In general, it is important to start the iteration with estimates that are close to the true parameter values. Conversely, when the initial point selected is far from the minimum, the scheme may be inefficient or even divergent. Here we refer to the initial guesses of material parameters (i.e., Young's modulus $E$, shear modulus $G$ and density $\rho$) of the soft manipulator provided by manufacturers, and choose the undeformed reference straight shape of the arm as initial configuration. When the material parameters are not available from robot manufacturers, we can guess an initial value in accordance with the properties of the material, and allow the algorithm to run multiple times in order to determine the optimality of the solution for this NLP problem.
\end{remark}

\section{Simulation comparison of discrete Cosserat models}\label{simulation_comparison}

This section is devoted to validating the precision of the proposed PLS Cosserat models, by comparing it with the result obtained via finite-element method (FEM). In addition, since PLS Cosserat shares the same local approximation scheme with PCS Cosserat, we will compare as well the precision of PLS model and PCS model.

\subsection{Simulation Setup}

%
%Since it is essential to compare the modeling effects of the discrete Cosserat models mainly including PLS and PCS under external forces, both of them are then applied to a cantilever rod under gravity. 

\begin{figure}[!t]
	\centering
	\includegraphics[width=3in]{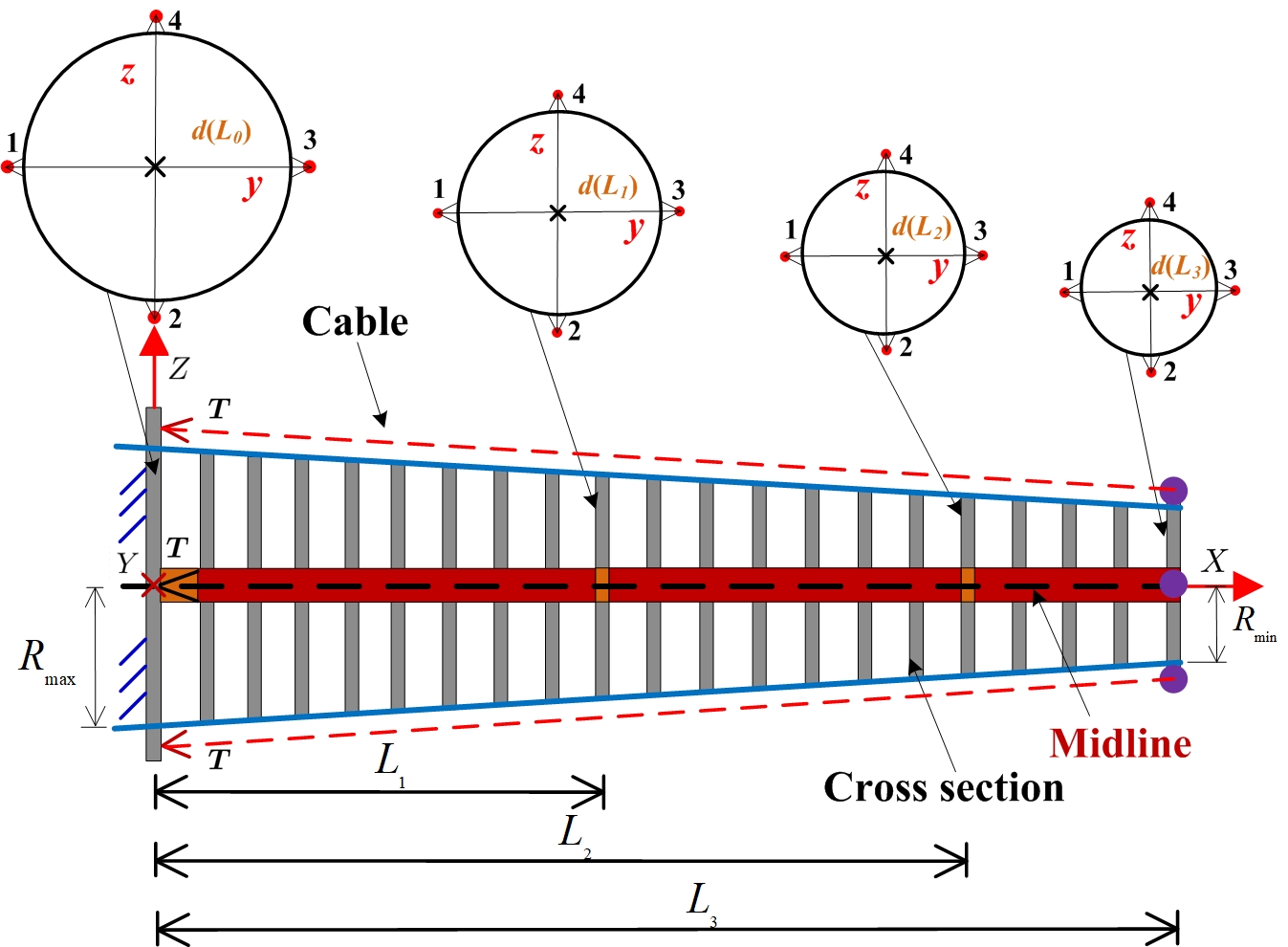}
	\caption{The side view of the PLS Cosserat model with three continuum sections actuated by four cables.}
	\label{Cables-distribution}
\end{figure}

The comparison is effectuated by simulating a cantilever rod under external forces (for example, under gravity). The simulated rod is of conical shape and actuated by cables (see Fig.~\ref{Cables-distribution}), with total length $L = 0.20\ {\rm m} $, base radius $R_{\rm {max}}=1\times 10^{-2} \ {\rm m} $, tip radius $R_{\rm {min}}=5 \times 10^{-3}\ {\rm m} $, Young modulus $E = 1.1\times 10^5 \ {\rm {Pa}}$, shear modulus $G=3.793\times 10^4 \ {\rm {Pa}}$, and density of material $\rho = 2000 $\ ${\rm kg}/\rm m^3$. $\small\boldsymbol{\xi}_0=\left[0, 0, 0, 1, 0, 0 \right]^{\rm T}$ represents the undeformed straight configuration when the rod is stress-free. In addition, the rod is divided into three sections, and the length of each section is separately $9\times 10^{-2}\ {\rm m}$, $7\times 10^{-2}\ {\rm m}$ and $4\times 10^{-2}\ {\rm m}$ from the base to tip. Besides, the rod shares the $X$-axis, $Y$-axis and $Z$-axis with the inertial frame, and thus the map $\small\boldsymbol{g}_r$ between the base frame of the rod and the inertial frame is a $4\times4$ identity matrix. 

Several cables are attached at the free end of the rod, and parallel to the surface of soft rod to produce the maximum torque and simultaneously to reduce the cables' friction, as shown in Fig.~\ref{Cables-distribution}. Thus, the friction of the cables can be neglected. Additionally, the local distance between the midline of soft rod and the cable $i\in[1,4]$ at the cross section $X$ is defined as $\small\boldsymbol{d}_i(X)\in\mathbb{R}^3$, 
we can then obtain the cable position vector  in the inertial frame $\small\boldsymbol{u}_{ci} =\boldsymbol{u}+\boldsymbol{R}\boldsymbol{d}_i$. 
By taking the derivative of cable position vector w.r.t. arc length and normalizing, the unit vector $\boldsymbol{{\rm t}}_{ci}(X,t)\in\mathbb{R}^3$ tangent to cable path yields \cite{renda2014dynamic}:
\begin{equation*}
	\small
	\begin{split}
	\boldsymbol{{\rm t}}_{ci}(X,t)&=\frac{\boldsymbol{R}^{-1}\boldsymbol{u}'_{ci}}{\Vert\boldsymbol{R}^{-1}\boldsymbol{u}'_{ci}\Vert}=\frac{\boldsymbol{R}^{-1}(\boldsymbol{u}'+\boldsymbol{R}'\boldsymbol{d}_i+\boldsymbol{R}\boldsymbol{d}'_i)}{\Vert\boldsymbol{R}^{-1}(\boldsymbol{u}'+\boldsymbol{R}'\boldsymbol{d}_i+\boldsymbol{R}\boldsymbol{d}'_i)\Vert}\\&=\frac{\boldsymbol{Q}+\boldsymbol{K}\times\boldsymbol{d}_i+\boldsymbol{d}'_i}{\Vert\boldsymbol{Q}+\boldsymbol{K}\times\boldsymbol{d}_i+\boldsymbol{d}'_i\Vert_2}.
	\end{split}
\end{equation*}
The distance $\small\boldsymbol{d}_i(X)$ is fixed when the soft manipulator is designed. Accordingly, the unit tangent vector $\boldsymbol{{\rm t}}_{ci}$ hinges on the angular strain $\small\boldsymbol{K}(X)$ and linear strain $\small\boldsymbol{Q}(X)$ of the soft manipulator. The actuation wrench $\small\boldsymbol{\mathcal{F}}_{ia}(X)$  for unit of $X$ can be obtained by calculating the torque and force exerted by the cables, and it is given by
\begin{equation}
	\small
	\begin{split}
		\boldsymbol{\mathcal{F}}_{i{\tiny }a}(X)&=\left[\begin{matrix}
			\boldsymbol{d}_1\times\boldsymbol{{\rm t}}_{c1}&\boldsymbol{d}_2\times\boldsymbol{{\rm t}}_{c2}&\boldsymbol{d}_3\times\boldsymbol{{\rm t}}_{c3}&\boldsymbol{d}_4\times\boldsymbol{{\rm t}}_{c4}\\
			\boldsymbol{{\rm t}}_{c1}&\boldsymbol{{\rm t}}_{c2}&\boldsymbol{{\rm t}}_{c3}&\boldsymbol{{\rm t}}_{cs}
		\end{matrix}\right] \boldsymbol{T}\\&=\boldsymbol{\Lambda}(X)\boldsymbol{T}
		\label{act}
	\end{split}
\end{equation}
where %the minus in (\ref{act}) represents the direction of cable's tension opposite to the $X$-axis,
$\small\boldsymbol{\Lambda}(X)\in\mathbb{R}^{6\times 4}$ is a matrix function whose columns are composed of vector functions, and with $\small\boldsymbol{T}\in\mathbb{R}^4$ is the system input vector consisted of magnitude of all cables' tension. %$\mathbf{\Lambda}(X)\in\mathbb{R}^{6\times s}$ and $\boldsymbol{T}\in\mathbb{R}^s$.

With regards to the external wrenches, under gravity is considered in our test for the sake of simplicity, which can be then expressed as 
\begin{equation}
	\small
	\overline{\boldsymbol{\mathcal{F}}}_{e}(X)=\boldsymbol{\mathcal{M}}{\rm {Ad}}^{-1}_{\boldsymbol{g}{(X)}}{\rm {Ad}}^{-1}_{\boldsymbol{g}_r}\boldsymbol{\mathcal{G}}.
	\label{gravity}
\end{equation}
where the inverse of adjoint representation of the Lie group is used to transform twists from the inertial to body frame, $\small\boldsymbol{\mathcal{G}}=\left[0, 0, 0, 0, 0, 9.81 \right]^{\rm T} $ is the gravity acceleration twist w.r.t. the inertial frame.

%The PLS Cosserat dynamic model discretizes the weak form of the dynamical equilibrium equation by using the geometric Jacobian. 
By the selection of cable actuation manner, %a new PLS Cosserat dynamic system without considering the boundary conditions yields
%\begin{equation}
%\small
%\begin{split}
%\boldsymbol{M}(\boldsymbol{q})\ddot{\boldsymbol{q}}+\boldsymbol{C}(\boldsymbol{q},\dot{\boldsymbol{q}})\dot{\boldsymbol{q}}-\hh{\boldsymbol{F}_e(\boldsymbol{q})}&+\cdots\\\cdots-\boldsymbol{K}(\boldsymbol{q})(\boldsymbol{q}-\boldsymbol{q}_0)-\boldsymbol{D}(\boldsymbol{q})\dot{\boldsymbol{q}}&=\boldsymbol{H}(\boldsymbol{q})\boldsymbol{T}.
%\label{dy_final}
%\end{split}
%\end{equation}
%Aside from the above motion equation, the boundary condition in (\ref{BC}) and the free end  of the soft manipulator at the cross section $X=L_N$ not subjected to the external wrench must be also considered. Thus, 
the strain twist at $X=L_N$ satisfies the following differential equation
\begin{equation}
\small
\boldsymbol{\Sigma}(L_N)(\overline{\boldsymbol{\xi}}_N-\boldsymbol{\xi}_0)+\boldsymbol{\gamma}(L_N)\dot{\overline{\boldsymbol{\xi}}}_N=-\boldsymbol{\Lambda}(L_N)\boldsymbol{T}.
\label{boundary_condition}
\end{equation}
%If the actuation wrench $\small\boldsymbol{\mathcal{F}}_{a}(L_N)=\boldsymbol{0}$, we have $\small\overline{\boldsymbol{\xi}}_{N}=\boldsymbol{\xi}_0$. While there are cables attached at $X=L_N$ with $\small\boldsymbol{\mathcal{F}}_{a}(L_N)$, the rightmost strain twist $\small\overline{\boldsymbol{\xi}}_N$ of the section $N$ can be computed by solving the ODE (\ref{last}). 

Substituting (\ref{gravity}) together with (\ref{boundary_condition}) into (\ref{dynamic_ode}), the PLS Cosserat dynamic system driven by cables can be given by
\begin{equation}
\small
\begin{split}
\left[\begin{matrix}
\boldsymbol{M}(\boldsymbol{q})\\
\boldsymbol{\Gamma}
\end{matrix}\right]
\ddot{\boldsymbol{q}}+\left[ \begin{matrix}
\boldsymbol{C}(\boldsymbol{q},\dot{\boldsymbol{q}})\\\boldsymbol{\sigma}	\end{matrix}\right]
\dot{\boldsymbol{q}}&-\left[ \begin{matrix}	\boldsymbol{G}(\boldsymbol{q}){\rm Ad}^{-1}_{\boldsymbol{g}_r}\boldsymbol{\mathcal{G}}+\boldsymbol{F}_{\rm i}(\boldsymbol{q},\dot{\boldsymbol{q}})\\
\boldsymbol{0}_{6\times1}
\end{matrix}\right]\\=\left[ \begin{matrix}	\boldsymbol{H}(\boldsymbol{q}) \\
\boldsymbol{0}_{6\times 4}
\end{matrix}\right]\boldsymbol{T}&+\left[ \begin{matrix}
\boldsymbol{0}_{6N\times 4}\\
-\boldsymbol{\Lambda}(L_N)
\end{matrix}\right]\dot{\boldsymbol{T}}
\end{split}
\label{dynamic_ode_final}
\end{equation}
with
$$
\small\boldsymbol{G}(\boldsymbol{q})=\boldsymbol{\mathcal{P}}^{\rm T}\int_{0}^{L_N}\boldsymbol{J}^{\rm T}\boldsymbol{\mathcal{M}}{\rm Ad}^{-1}_{\boldsymbol{g}(X)}{\rm d}X,
$$
and
$$
\small\boldsymbol{H}(\boldsymbol{q})=\boldsymbol{\mathcal{P}}^{\rm T}\int_{0}^{L_N}\boldsymbol{J}^{\rm T}\left( \boldsymbol{\Lambda}'-{\rm ad}^{\rm{T}}_{\boldsymbol{\xi}(X)}\boldsymbol{\Lambda}\right){\rm d}X.
$$

Redefining each term and naming the coefficient matrices of (\ref{dynamic_ode_final}), 
the PLS Cosserat dynamic model for numerical simulation can be formulated as the following general Lagrangian structural form
\begin{equation}
\small
\underline{\boldsymbol{M}}(\boldsymbol{q})\ddot{\boldsymbol{q}}+\underline{\boldsymbol{C}}(\boldsymbol{q},\dot{\boldsymbol{q}})\dot{\boldsymbol{q}}-\underline{\boldsymbol{K}}(\boldsymbol{q},\dot{\boldsymbol{q}})=\underline{\boldsymbol{H}}(\boldsymbol{q})\boldsymbol{T}+\overline{\boldsymbol{H}}\dot{\boldsymbol{T}}
\label{ddd}
\end{equation}
where $\small\underline{\boldsymbol{M}}(\boldsymbol{q})$ is a symmetric positive-definite mass matrix, $\small\underline{\boldsymbol{H}}(\boldsymbol{q})$ represents the actuation matrix.
It is worth noting that continuous Cosserat rod dynamic formulation of soft robots leads to a PDE in the form of a boundary value problem (BVP), however, the PLS dynamic Cosserat model actuated by cables takes the form of a nonlinear ODE, which paves the way to design its model-based controller.

\subsection{Comparison of Accuracy for Static Models}
The model deduced by using the geometrically exact FEM approach is chosen as an alternative of the real soft arm. In short, the main idea of FEM is to spatially discretize the geometric shape of the rod by using finite number of finer elements. The geometric model of the rod is established in the SolidWorks, we then use the FEM to obtain the equilibrium position of the cantilever rod. Specifically, in terms of spatial discretization of the studied cantilever rod, quadrilateral mesh elements are used, and the rod is spatially discretized into $650$ elements along the $X$-axis. The static information that the mesh average element quality is $0.8033$ and minimum element quality is $0.6181$ indicates the high discretization accuracy of the model.
 %\hh{Finally, the Newton method is used to compute the discretized model.}

\begin{figure*}[!t]
	\centering
	\subfloat[]{\includegraphics[width=2.57in]{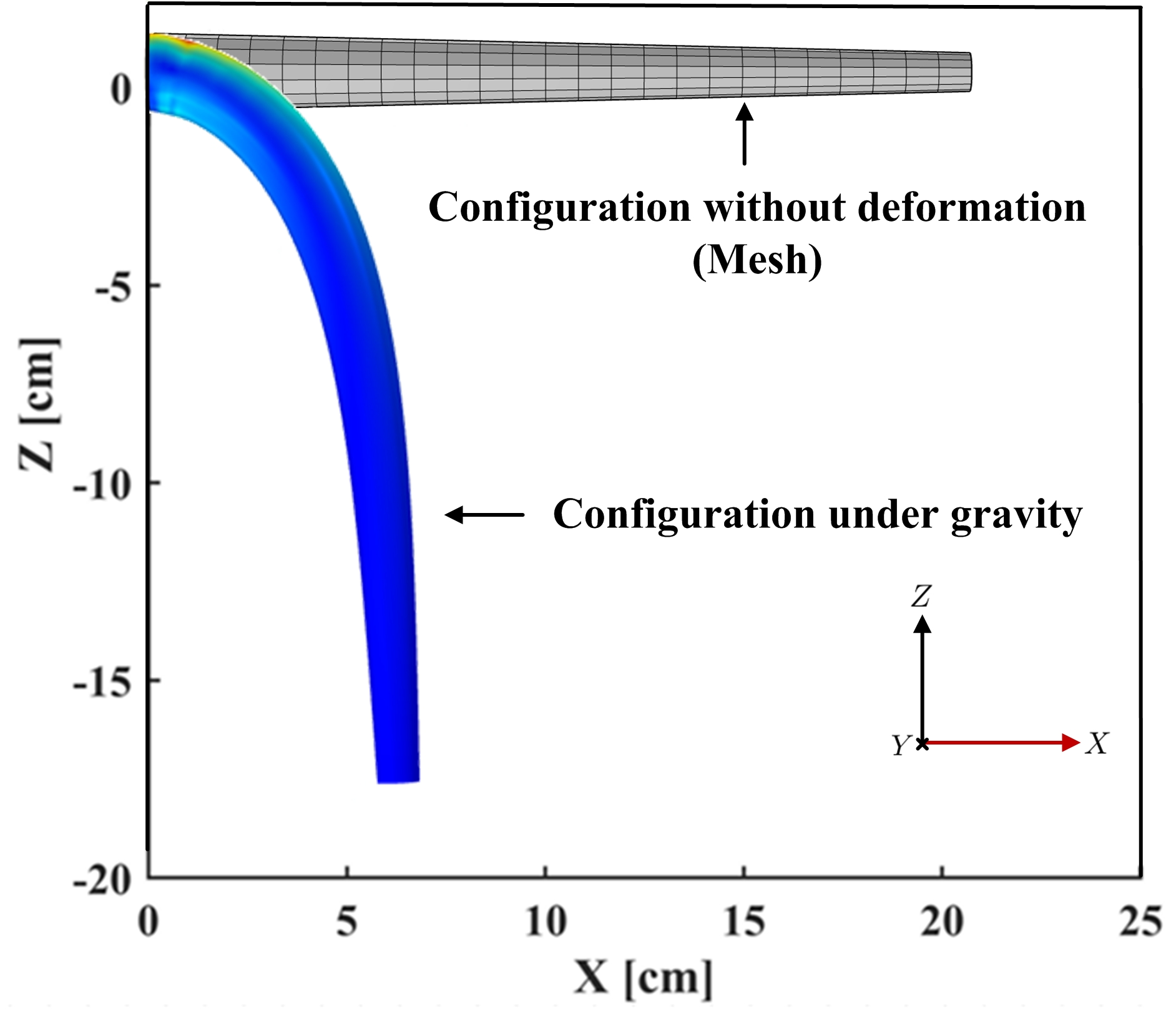}%
		\label{FEM-Comsol}}
	\hfil
	\subfloat[]{\includegraphics[width=2.5in]{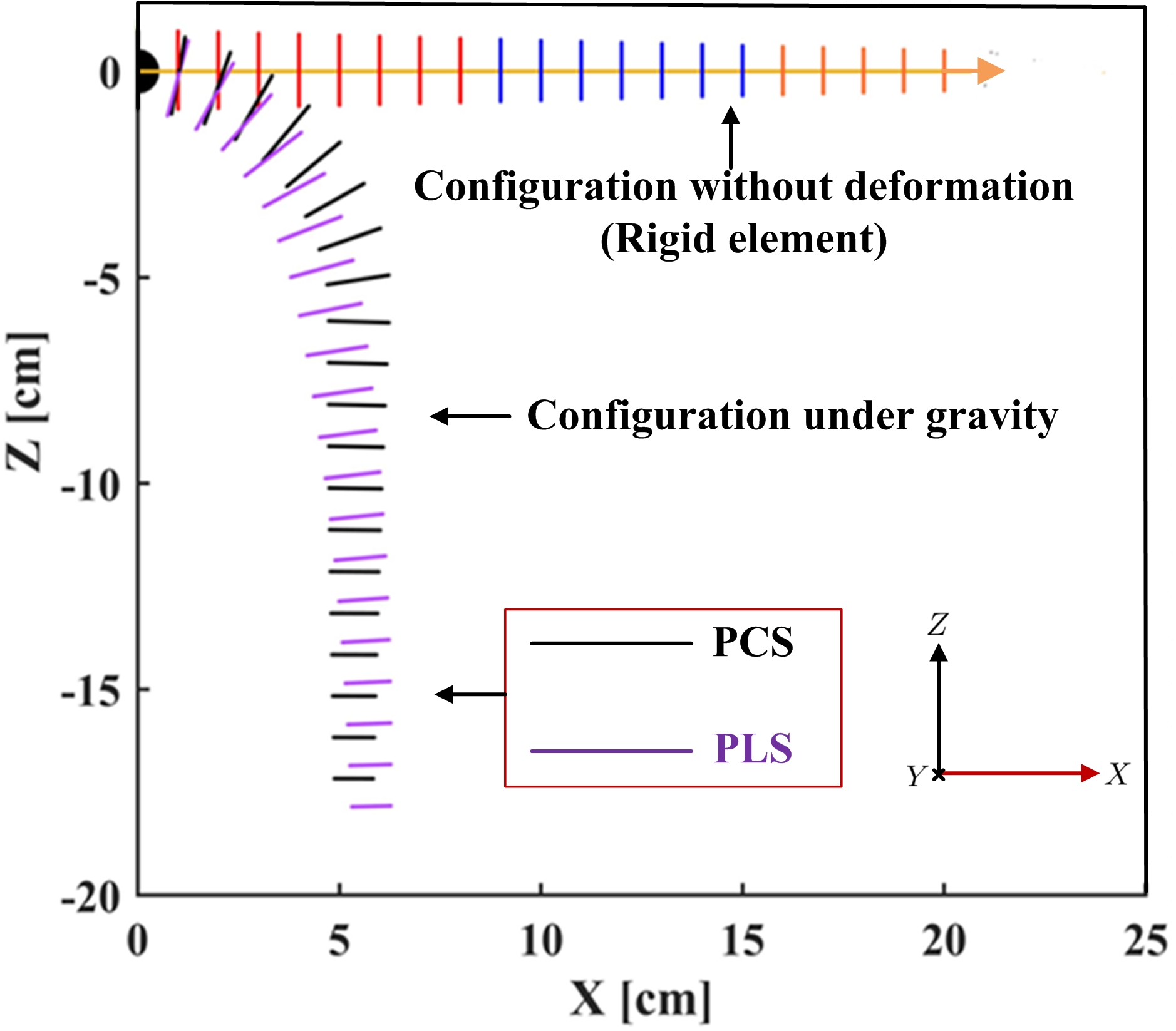}%
		\label{PCSandPLS_st}}
	\caption{Simulation comparison of three different modeling methods for the cantilever rod before and after deformation. (a) FEM. (b) PCS and PLS Cosserat models in MATLAB.}
	\label{fig:numerical}
\end{figure*}

For a more specific comparison and evaluation, the material and geometric parameters for the discrete Cosserat models with the same sections in MATLAB are in accordance with those of the FEM. From the simulation results, the equilibrium position of the end-effector of the cantilever rod via FEM under gravity is $\small\boldsymbol{u}_e=[5.8479, \ 0, \ -17.8395]^{\rm T}$, as shown in Fig.~\ref{fig:numerical}\subref{FEM-Comsol}. It took around $14$ seconds to complete one simulation because the FEM for large deformation always require unnecessary computation. Similarly, we can derive the positions of the end-effector of the discrete Cosserat static models by using the Newton method. From the perspective of computation time of discrete Cosserat static models, we observe that the systems can converge in less than $2$ seconds mainly due to the use of the basic idea of order-model reduction. Comprehensively considering the comparison results among them as illustrated in Table~\ref{PLS-PCS} and Fig.~\ref{fig:numerical}, we come to a conclusion that the model via the PLS Cosserat modeling approach fits much better with the FEM compared to the PCS, with the relative position coordinate error less than $5\%$. In other words, the PLS Cosserat static model is essentially comparable to the FEM in terms of accuracy, which can be further verified by the purple PLS configuration tendency plotted in Fig.~\ref{fig:numerical}\subref{PCSandPLS_st}, almost same to the FEM in Fig.~\ref{fig:numerical}\subref{FEM-Comsol}. Logically, this can be explained by the fact that the piecewise linear interpolation technique applied to all cross sections of the proposed discrete Cosserat model makes the system locally approximate to the deformation behavior of the soft manipulator in the real scenario.
\begin{small}
\begin{table}
	\centering
	\setlength{\abovecaptionskip}{0pt}
	\setlength{\belowcaptionskip}{-5pt}
	\caption{Comparison results of PCS and PLS Cosserat static models w.r.t. FEM under gravity in terms of end-effector position and its relative error}
	\begin{tabular}{ccccccc} 
	\toprule
	\multirow{2}{*}{\tabincell{c}{Modeling\\ method}} & \multicolumn{3}{c}{\tabincell{c}{Position of end-effector\\ (Unit:$\rm{cm}$)}}&\multicolumn{3}{c}{\tabincell{c}{Relative error of end-\\effector position w.r.t.\\ FEM (Unit: $\%$)}}\\ 
	\cline{2-7}& $u_x$ & $u_y$ & $u_z$&$e_x$ & $e_y$ & $e_z$\\
	\midrule
	FEM 	&5.8479&0.0000&-17.8395&$\times$&$\times$&$\times$\\ 
	PCS \cite{renda2018discrete} &5.3450&0.0000&-17.1693&-8.60&0.00&-16.88\\ 
	PLS&5.7787&0.0000&-17.8394&-1.18&0.00&-0.01\\
	\bottomrule
    \end{tabular}
	\label{PLS-PCS}
\end{table}
\end{small}

\subsection{Accuracy Comparison of PLS Model with Different Modes}
\begin{figure*}[!t]
	\centering
	\subfloat[]{\includegraphics[width=2.3in]{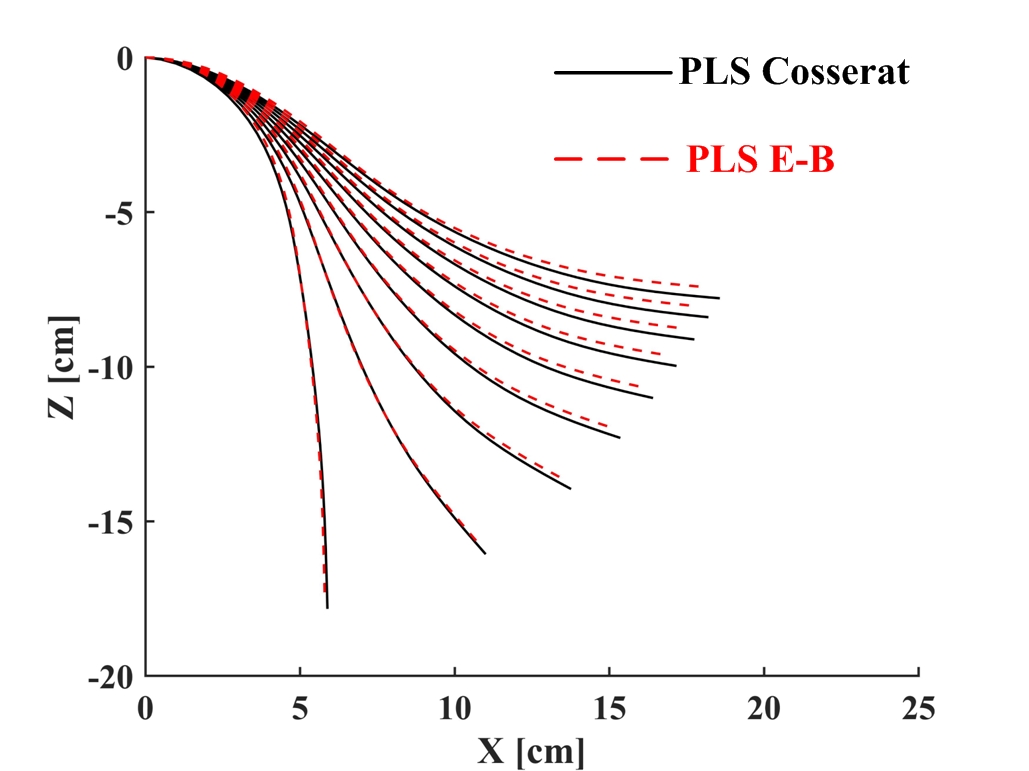}%
		\label{E-C}}
	\hfil
	\subfloat[]{\includegraphics[width=2.3in]{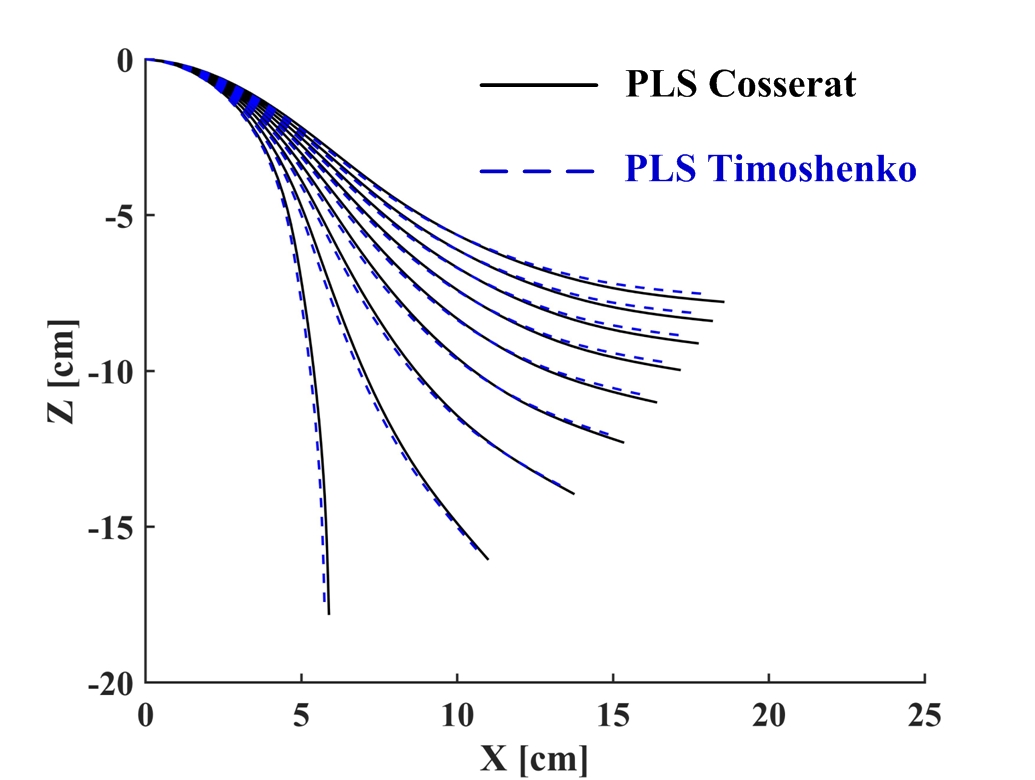}%
		\label{C-T}}
		\hfil
	\subfloat[]{\includegraphics[width=2.3in]{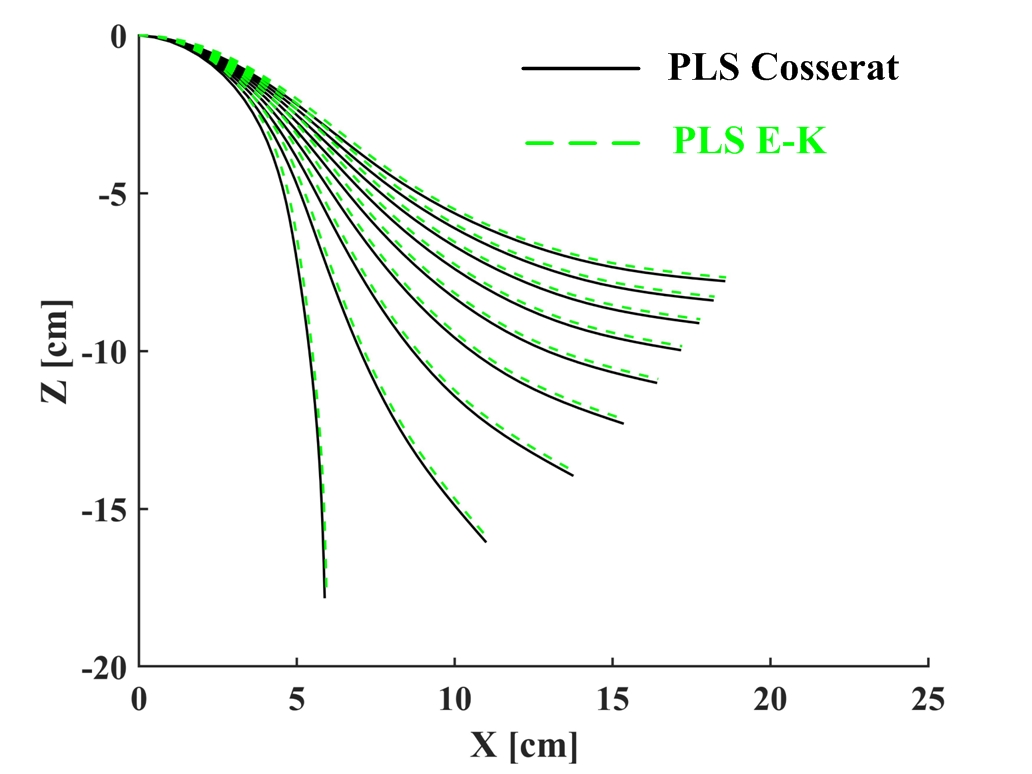}%
		\label{C_E-K}}
	\caption{Deformation comparison among beam models via PLS Cosserat under external loads. (a) Cosserat versus Euler-Bernoulli. (b) Cosserat versus Timoshenko. (c) Cosserat versus extensible Kirchhoff.}
	\label{E-C-T}
\end{figure*}

Based on the aforementioned manipulator parameters setting, different models via the PLS Cosserat involving Euler-Bernoulli (E-B), extensible Kirchhoff (E-K) and Timoshenko beams are established by strain mode selection. These beams are fixed at $X=0$, and subject to gravity as well as an external imposed concentrated load with $\small\boldsymbol{\mathcal{F}}_{\rm{tip}}=[0,0,0,F,0,0]^{\rm T}$ (i.e., tension along $X$-axis) at $X=L$. The three-dimensional static simulation is implemented by increasing load with increment of $0.05\ {\rm N}$ at a time. Fig.~\ref{E-C-T}(\subref{E-C}-\subref{C_E-K}) displays a contrast of the evolution of the equilibrium configurations among these beams for several sets of the tip load, and Table~\ref{E-B_Timoshenko_Kirchhoff} shows different beams' end-effector positions versus that of FEM. The results indicate that it is feasible to remove negligible modes in some particular case with low-precision requirement, which contributes to the real-time simulation and control.

\begin{scriptsize}
	\begin{table}
		\centering
		\setlength{\abovecaptionskip}{0pt}
		\setlength{\belowcaptionskip}{-5pt}
		\caption{Comparison results of different beam models via PLS Cosserat w.r.t. FEM under gravity in terms of end-effector position and its relative error}
		\begin{tabular}{ccccccc} 
			\toprule
			\multirow{2}{*}{\tabincell{c}{Modeling\\ method}} & \multicolumn{3}{c}{\tabincell{c}{Position of end-effector\\ (Unit:$\rm{cm}$)}}&\multicolumn{3}{c}{\tabincell{c}{Relative error of end-\\effector position w.r.t.\\ FEM (Unit: $\%$) }}\\ 
			\cline{2-7}& $u_x$ & $u_y$ & $u_z$&$e_x$ & $e_y$ & $e_z$\\
			\midrule
			FEM 	&5.8479&0.0000&-17.8395&$-$&$-$&$-$\\ 
			Cosserat &5.7787&0.0000&-17.8394&-1.18&0.00&-0.01\\ 
			E-B&5.4940&0.0000&-17.7527&-6.05&0.00&-0.49\\
			E-K &5.6925&0.0000&-17.8523&-2.66&0.00&
			0.07\\
			Timoshenko &5.3596&0.0000&-17.9002&-8.35&0.00&
			0.34\\
			\bottomrule
		\end{tabular}
		\label{E-B_Timoshenko_Kirchhoff}
	\end{table}
\end{scriptsize}
\vspace*{10pt}

\section{PLS Cosserat model validation}

\subsection{Studied Soft Manipulator}
\begin{figure}[!t]
	\centering
	\includegraphics[width=3.2in]{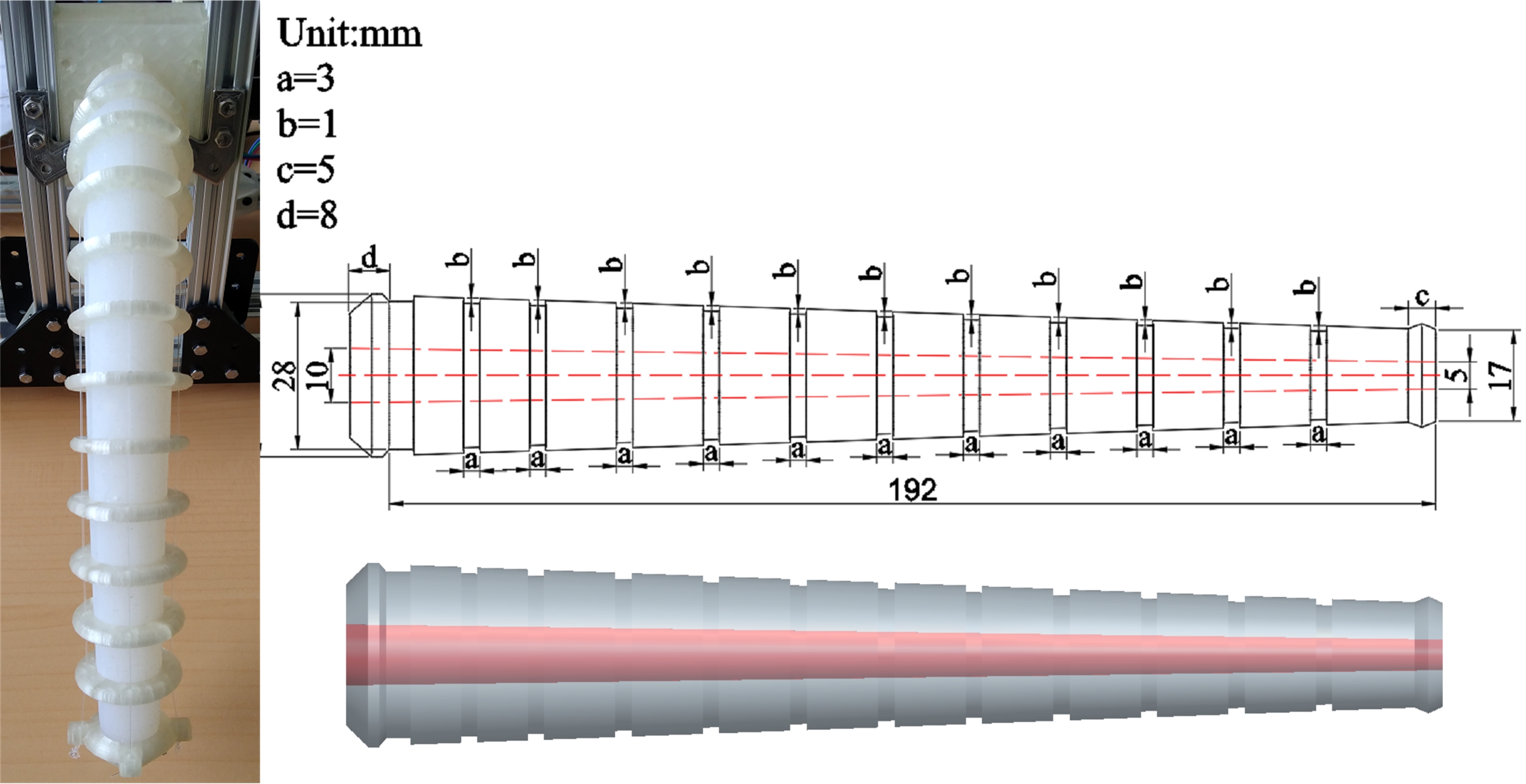}
	\caption{Manipulator prototype design. }
	\label{prototype}
\end{figure}

\begin{figure}[!t]
	\centering
	\includegraphics[width=3in]{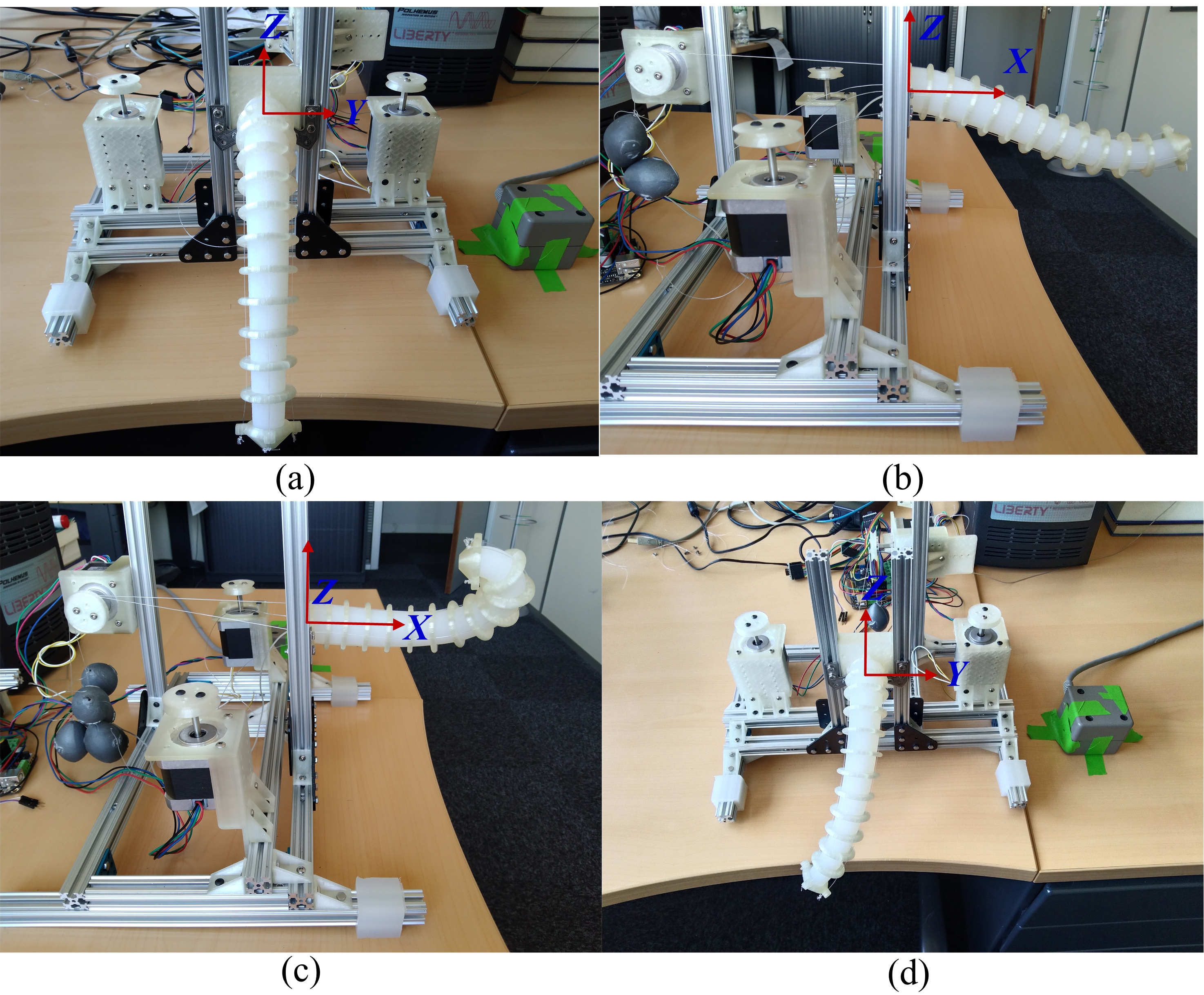}
	\caption{Several snapshots of the different experiments for implementing parameters identification.}
	\label{test_sample}
\end{figure}

A prototype of soft manipulator, similar to that used in the simulation, was designed to carry out the material parameters identification of the PLS Cosserat model by the real input-output relationship obtained from the experimental setup. The exact geometric parameters of the manipulator are illustrated in Fig.~\ref{prototype}. The investigated soft arm is controlled by 4 cables mounted through it from the base to the tip, and the cables are respectively driven by different weights, as shown in Fig.~\ref{test_sample}. Aiming at obtaining the position of the end-effector, a magnetic sensor is positioned on the tip of the arm, and for this purpose, a long and conical hole was made along the whole length of the arm. In addition, several 3D-printed rigid rings are mounted along the soft manipulator to minimize friction between the cables and a single conical piece of silicone. Finally, the casting material of the manipulator is an isotropic silicone rubber.

\vspace*{-3pt}

\begin{small}
\begin{table}
	\begin{center}
		\caption{Experimental samples for identification algorithm}
		\label{specific_input_output}
		\begin{tabular}{| c | c | c |}
			\hline
			Order of & Cables' tension $\boldsymbol{T}_i$& Position of the\\
			experiment & (Unit: N) &   end-effector $\boldsymbol{u}_{ei}$ (Unit: $\rm{cm}$)\\
			\hline
			1& $[\begin{matrix}
			0&0&0&0
			\end{matrix}]$ & $[\begin{matrix}
			15.77&0.03&-10.10
			\end{matrix}]$\\
			\hline
			2&$[\begin{matrix}
			0&0&0&0.98
			\end{matrix}]$ &$[\begin{matrix}
			17.73&0.02&-6.44
			\end{matrix}]$\\ 
			\hline
			3&$[\begin{matrix}
			0.98&0&0&4.90
			\end{matrix}]$ &$[\begin{matrix}
			6.85&-4.77&7.20
			\end{matrix}]$\\ 
			\hline
			4&$[\begin{matrix}
			0&0&0&1.96
			\end{matrix}]$ &$[\begin{matrix}
			18.31&0.01&-1.90
			\end{matrix}]$\\
			\hline
			5&$[\begin{matrix}
			1.96&0&0&1.96
			\end{matrix}]$ &$[\begin{matrix}
			13.09&-9.24&-2.46
			\end{matrix}]$\\
			\hline
			6&$[\begin{matrix}
			0.98&0&0&0
			\end{matrix}]$ &$[\begin{matrix}
			14.72&-4.20&-9.78
			\end{matrix}]$\\
			\hline
		\end{tabular}
	\end{center}
\end{table}
\end{small}

\subsection{Validation and Discussion}
 The model validation which determines whether the model is proper enough for its intended use is implemented. Six sets of experiments are performed to acquire the position information of end-effector under the effect of different loads exerted by weights equivalent to the cable's tension. Table~\ref{specific_input_output} provides specific input and output values of the experiments, and the position as well as orientation of the manipulator in several cases are displayed in Fig.~\ref{test_sample}. Subsequently, the experimental outputs obtained by the position sensor are utilized to realize the proposed parameters identification algorithm, and thus the material parameters $\small\boldsymbol{\theta}$ can be calculated.

The material parameters obtained by solving the NLP problem are as follows: Young's modulus $E=2.563\times10^5\ {\rm{Pa}}$, shear modulus $G=8.543\times10^4\ {\rm{Pa}}$, and density of material $\rho=1.41\times10^3\ {\rm {kg/m^3}}$.

\begin{figure}[!t]
	\centering
	\includegraphics[width=3in]{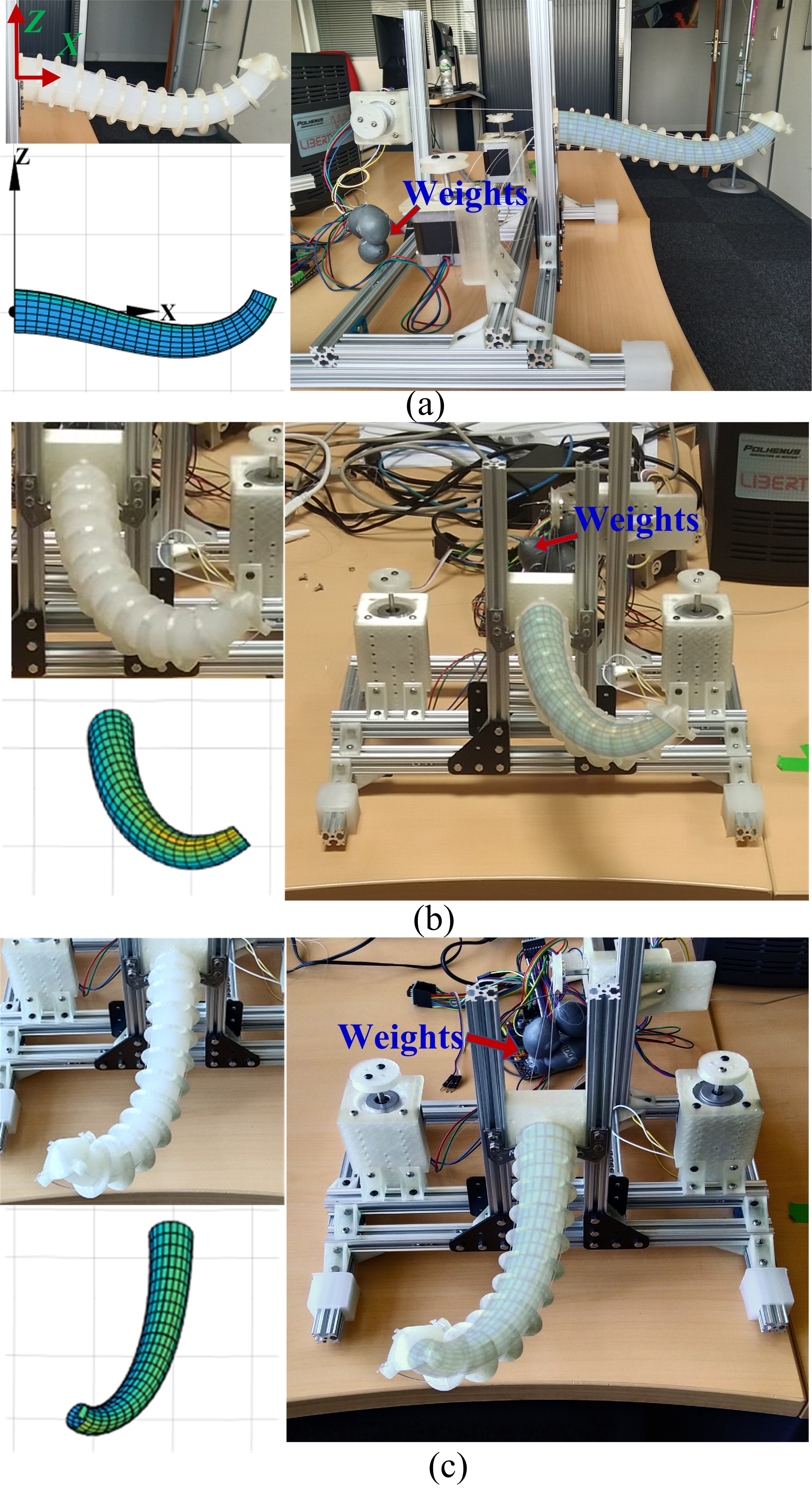}
	\caption{Configuration comparison between simulation and experiment under gravity and different cables' tension. (a) $\boldsymbol{T}=\left[ \begin{matrix}
		0&0&0&2.94
		\end{matrix}\right]^{\rm T} $. (b) $\boldsymbol{T}=\left[ \begin{matrix}
		0&0&1.96&1.96
		\end{matrix}\right]^{\rm T} $. (c) $\boldsymbol{T}=\left[ \begin{matrix}
		0.98&0&0&2.94
		\end{matrix}\right]^{\rm T}$.}
	\label{model_validation}
\end{figure}

After that, model validation is performed to verify the accuracy of the PLS Cosserat model with the identified parameters. Considering uncertain factors of the experiment, we repeat 10 times for each input and then obtain the average of end-effector position to compare with simulation, five different groups of control inputs and average of outputs provided in Table~\ref{error_input_output} are selected to compare the position and orientation of the soft arm between the experiments and simulations. The comparison results indicate the outputs (position of the end-effector) from the PLS Cosserat model are almost identical to those of the experiments in three cases illustrated in Fig.~\ref{model_validation}. As for the remaining two sets of experiments, there are larger absolute errors of end-effector position between the model and manipulator than those from the other three cases, which may be due to tiny pleats on the silicone surface caused by relatively larger cables' tension.
\begin{small}
	\begin{table}
		\begin{center}
			\caption{Different experiments for model validation}
			\label{error_input_output}
			\begin{tabular}{| c | c |c |}
				\hline
				Order of & Cables' tension $\boldsymbol{T}_i$ & Average of position of end- \\
				 control input&(Unit: N)&effector $\small\boldsymbol{u}_{ei}$ (Unit: ${\rm{cm}}$)\\
				\hline
				1 (10 times)& $[\begin{matrix}
				0&0&0&2.94
				\end{matrix}]$& $[\begin{matrix}
				16.81&0.03&2.87
				\end{matrix}]$\\
				\hline
				2 (10 times)&$[\begin{matrix}
				0&0&1.96&1.96
				\end{matrix}]$& $[\begin{matrix}
				13.22&9.42&-2.37
				\end{matrix}]$\\ 
				\hline
				3 (10 times)&$[\begin{matrix}
				0.98&0&0&2.94
				\end{matrix}]$ & $[\begin{matrix}
				14.97&-6.03&2.53
				\end{matrix}]$\\
				\hline
				4 (10 times)&$[\begin{matrix}
				0&0&2.94&6.86
				\end{matrix}]$ & $[\begin{matrix}
				1.50&3.69&2.26
				\end{matrix}]$\\
				\hline
				5 (10 times)&$[\begin{matrix}
				0&0&0&5.88
				\end{matrix}]$& $[\begin{matrix}
				4.28&0.02&7.55
				\end{matrix}]$\\
				\hline
			\end{tabular}
		\end{center}
	\end{table}
\end{small}
%\vspace*{-3pt}
\begin{figure}[!t]
	\centering
	\includegraphics[width=3.5in]{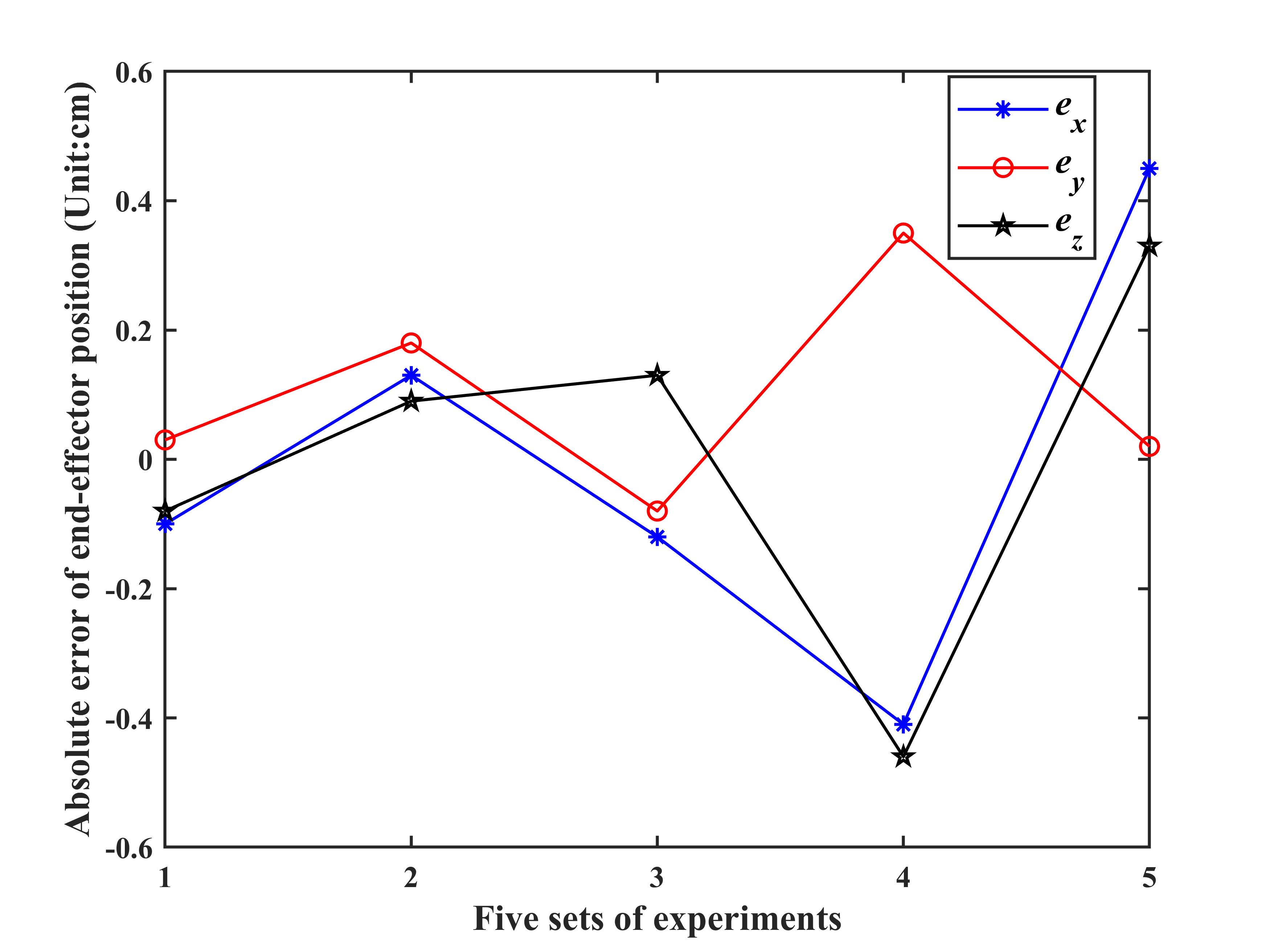}
	\caption{The illustration of absolute errors of the end-effector position for the PLS Cosserat model and soft manipulator under different control inputs.}
	\label{error_identification}
\end{figure}

The absolute errors of the end-effector position of the model with respect to those of the experiments are all less than $5\ {\rm{mm}}$, as shown in Fig.~\ref{error_identification}, further showing the effectiveness of this parameters identification method and the accuracy of the PLS Cosserat model.

\section{Conclusion and future work}
A piecewise linear strain Cosserat dynamics model for the soft manipulator has been developed for the first time, which combines the advantages of the PCS and VS Cosserat models. This method depends on a rigorous mathematical framework via the Lie group theory which facilitates a natural coupling of the position and the orientation variables, and exhibits an important advantage over avoiding the difficulty in the parameterization of rotation matrix. The PLS Cosserat model has been compared with the discrete models reported in the published literature, showing comparable to the FEM and even better results than the PCS Cosserat model in terms of accuracy. As proved in the section~\ref{simulation_comparison}, the PLS Cosserat model shows great potential to be universally applied to the modeling of slender rod-like soft arms in a real scenario. 

%\hh{In the practical application of soft robotics, the dimension of the system can be reduced based on the proposed Cosserat model, a generalization of strain modes...}

The parameters identification based on the PLS Cosserat static model can be considered as a NLP problem with several nonlinear equality constraints. A general framework has been proposed for the material parameters estimation, which is applicable to identify the parameters of the soft manipulators with arbitrary actuation manner. To carry out the model validation, we designed a manipulator prototype made of silicone and established the experimental platform. Both simulation and experiment results indicate the proposed scheme is able to identify the material parameters of soft arm with higher precision\hh{, and provides the possibility for developing the model-based controller.}.

The standard Lagrangian formulation in (\ref{ddd}) for PLS Cosserat dynamic model will be used to design static and dynamic model-based controllers due to the relative parameterization of the soft arm actuated by cables in the near future.

%Sample MATLAB code that implements the PLS Cosserat modeling approach for a cantilever rod is provided in the attachment, and this code has been tested for the MATLAB 2020a release version.

\section*{Acknowledgments}
This work is partially supported by project COSSEROOTS. The author Haihong Li gratefully acknowledge the support from the China Scholarship Council (Grant no. 202008440356).
%This should be a simple paragraph before the References to thank those individuals and institutions who have supported your work on this article.


\appendix[]
\subsection{Analytic solution of differential kinematics model}\label{analytic_kinematics}
Based on the aforementioned two assumptions, the rightmost velocity of any segment $j$ along the section $n$ at time $t$ can be calculated by means of the integral of DE (\ref{kine1}).
\begin{align*}
\scriptsize	
\begin{split}
&\boldsymbol{\eta}(L_{n-1}+j\Delta X)
=\left[ \prod_{i=0}^{j-1}e^{-\Delta X{\rm ad}_{\boldsymbol{\xi}_n(L_{n-1}+i\Delta X)}}\right] \boldsymbol{\eta}(L_{n-1})+\\&\sum_{i=1}^{j}\Biggl\{\Bigg[ \prod_{\tau=i}^{j-1}e^{-\Delta X{\rm ad}_{\boldsymbol{\xi}_n(L_{n-1}+\tau\Delta X)}}\Bigg]\int_{L_{n-1}+(i-1)\Delta X}^{L_{n-1}+i\Delta X}\\ &e^{(s-L_{n-1}-i\Delta X){\rm ad}_{\boldsymbol{\xi}_n(L_{n-1}+(i-1)\Delta X)}}\dot{\boldsymbol{\xi}}_n\Big(L_{n-1}+(i-1)\Delta X\Big){\rm d}s\Biggr\}
\end{split}
\end{align*}
After that, the velocity of each cross section along the section $n$ at time $t$ can be analytically obtained
\begin{equation*}
\scriptsize
\begin{split}
&\boldsymbol{\eta}(X)
=e^{-(X-L_{n-1}-j\Delta X){\rm{ad}}_{\boldsymbol{\xi}_n(L_{n-1}+j\Delta X)}}\\&\Biggl\{\Big[ \prod_{i=0}^{j-1}e^{-\Delta X{\rm ad}_{\boldsymbol{\xi}_n(L_{n-1}+i\Delta X)}}\Big]   \boldsymbol{\eta}(L_{n-1})\\&\quad+ \sum_{i=1}^{j}\Big[\Big( \prod_{\tau=i}^{j-1}e^{-\Delta X{\rm ad}_{\boldsymbol{\xi}_n(L_{n-1}+\tau\Delta X)}}\Big)\int_{L_{n-1}+(i-1)\Delta X}^{L_{n-1}+i\Delta X}\\&\quad e^{(s-L_{n-1}-i\Delta X){\rm ad}_{\boldsymbol{\xi}_n(L_{n-1}+(i-1)\Delta X)}}\dot{\boldsymbol{\xi}}_n\Big(L_{n-1}+(i-1)\Delta X\Big){\rm d}s\Big]\Biggr\}\\
&\quad+\int_{L_{n-1}+j\Delta X}^{X}e^{-(X-s){\rm {ad}}_{\boldsymbol{\xi}_n(L_{n-1}+j\Delta X)}}\dot{{\boldsymbol{\xi}}}_n\Big(L_{n-1}+j\Delta X\Big){\rm d}s
\end{split}
\end{equation*}
Under the PLS assumption, the velocity of any cross section at $X$ and along the section $n$ at time $t$ can be re-formulated as follows
\setcounter{equation}{31}
\begin{equation}
\scriptsize	
\begin{split}
\boldsymbol{\eta}(X)&= e^{-(X-L_{n-1}-j\Delta X){\rm{ad}}_{\Theta_{nj}^{\vee}}}\Biggl\{ \Big( \prod_{i=0}^{j-1}e^{-{\rm ad}_{\Delta X\Theta_{ni}^{\vee}}}\Big)  \boldsymbol{\eta}(L_{n-1})\\
&\quad+\sum_{i=1}^{j}\Big[\Big(  \prod_{\tau=i}^{j-1}e^{-{\rm ad}_{\Delta X\Theta_{n\tau}^{\vee}}}\Big) \int_{L_{n-1}+(i-1)\Delta X}^{L_{n-1}+i\Delta X}\\&\qquad e^{(s-L_{n-1}-i\Delta X){\rm ad}_{\Theta_{n(i-1)}^{\vee}}} \alpha_{n(i-1)} {\rm d}s\Big] \dot{\overline{\boldsymbol{\xi}}}_{n-1}\\
&\quad+\sum_{i=1}^{j}\Big[\Big(  \prod_{\tau=i}^{j-1}e^{-{\rm ad}_{\Delta X\Theta_{n\tau}^{\vee}}}\Big) \int_{L_{n-1}+(i-1)\Delta X}^{L_{n-1}+i\Delta X}\\&\qquad e^{(s-L_{n-1}-i\Delta X){\rm ad}_{\Theta_{n(i-1)}^{\vee}}} \beta_{n(i-1)}{\rm d}s\Big] \dot{\overline{\boldsymbol{\xi}}}_n\Biggr\} \\
&\quad+\int_{L_{n-1}+j\Delta X}^{X}e^{-(X-s){\rm {ad}}_{\Theta_{nj}^{\vee}}} \alpha_{nj}{\rm d}s  \dot{\overline{\boldsymbol{\xi}}}_{n-1}\\&\quad+\int_{L_{n-1}+j\Delta X}^{X}e^{-(X-s){\rm {ad}}_{\Theta_{nj}^{\vee}}} \beta_{nj}{\rm d}s\dot{\overline{\boldsymbol{\xi}}}_n
\label{velo2}
\end{split}
\end{equation}
with $\scriptsize\Theta_{ni}=\alpha_{ni}  \widehat{\overline{\boldsymbol{\xi}}}_{n-1}+\beta_{ni} \widehat{\overline{\boldsymbol{\xi}}}_n$,  $\scriptsize\Theta_{nj}=\alpha_{nj}  \widehat{\overline{\boldsymbol{\xi}}}_{n-1}+\beta_{nj} \widehat{\overline{\boldsymbol{\xi}}}_n$, 
where $\alpha_{ni}=1- \frac{i\Delta X}{L_n-L_{n-1}}$, $\beta_{ni}=\frac{i\Delta X}{L_n-L_{n-1}}$, $\alpha_{nj}=1- \frac{j\Delta X}{L_n-L_{n-1}}$, $\beta_{nj}=\frac{j\Delta X}{L_n-L_{n-1}}$.

By the same reasoning, the rightmost acceleration of any segment $j$ of the section $n$ along the soft manipulator can be computed at time $t$ by the use of the integral of (\ref{kine2}).
\begin{equation*}
\scriptsize	
\begin{split}
&\dot{\boldsymbol{\eta}}(L_{n-1}+j\Delta X)=\left[ \prod_{i=0}^{j-1}e^{-\Delta X{\rm ad}_{\boldsymbol{\xi}_n(L_{n-1}+i\Delta X)}}\right] \dot{\boldsymbol{\eta}}(L_{n-1})\\&+\sum_{i=1}^{j}\Biggl\{\left[\prod_{\tau=i}^{j-1}e^{-\Delta X{\rm ad}_{\boldsymbol{\xi}_n(L_{n-1}+\tau\Delta X)}}\right]\int_{L_{n-1}+(i-1)\Delta X}^{L_{n-1}+i\Delta X}\\
&e^{(s-L_{n-1}-i\Delta X){\rm ad}_{\boldsymbol{\xi}_n(L_{n-1}+(i-1)\Delta X)}}{\rm d}s\ddot{\boldsymbol{\xi}}_n\Big(L_{n-1}+(i-1)\Delta X\Big)\Biggr\}\\
&+\sum_{i=1}^{j}\Biggl\{\left[\prod_{\tau=i}^{j-1}e^{-\Delta X{\rm ad}_{\boldsymbol{\xi}_n(L_{n-1}+\tau\Delta X)}}\right] \int_{L_{n-1}+(i-1)\Delta X}^{L_{n-1}+i\Delta X}\\&e^{(s-L_{n-1}-i\Delta X){\rm ad}_{\boldsymbol{\xi}_n(L_{n-1}+(i-1)\Delta X)}}{\rm {ad}_{\boldsymbol{\eta}(s)}}{\rm d}s\dot{\boldsymbol{\xi}}_n\Big(L_{n-1}+(i-1)\Delta X\Big)\Biggr\}
\end{split}
\end{equation*}
Considering the PLS hypothesis, the acceleration of any cross section at $X$ and along the section $n$ at time $t$ can be re-written as follows
\begin{equation}
\scriptsize	
\begin{split}
\dot{\boldsymbol{\eta}}(X)
&=e^{-(X-L_{n-1}-j\Delta X){\rm{ad}}_{\Theta_{nj}^{\vee}}}\Biggl\{ \Big( \prod_{i=0}^{j-1}e^{-{\rm ad}_{\Delta X\Theta_{ni}^{\vee}}}\Big)  \dot{\boldsymbol{\eta}}(L_{n-1})\\
&+\sum_{i=1}^{j}\Big[\Big(  \prod_{\tau=i}^{j-1}e^{-{\rm ad}_{\Delta X\Theta_{n\tau}^{\vee}}}\Big) \int_{L_{n-1}+(i-1)\Delta X}^{L_{n-1}+i\Delta X}\\&\quad e^{(s-L_{n-1}-i\Delta X){\rm ad}_{\Theta_{n(i-1)}^{\vee}}} \alpha_{n(i-1)} {\rm d}s\Big]\ddot{\overline{\boldsymbol{\xi}}}_{n-1}\\
&+\sum_{i=1}^{j}\Big[\Big(  \prod_{\tau=i}^{j-1}e^{-{\rm ad}_{\Delta X\Theta_{n\tau}^{\vee}}}\Big) \int_{L_{n-1}+(i-1)\Delta X}^{L_{n-1}+i\Delta X}\\&\quad e^{(s-L_{n-1}-i\Delta X){\rm ad}_{\Theta_{n(i-1)}^{\vee}}} \beta_{n(i-1)}{\rm d}s\Big]\ddot{\overline{\boldsymbol{\xi}}}_n\\
&+\sum_{i=1}^{j}\Big[\Big(  \prod_{\tau=i}^{j-1}e^{-{\rm ad}_{\Delta X\Theta_{n\tau}^{\vee}}}\Big) \int_{L_{n-1}+(i-1)\Delta X}^{L_{n-1}+i\Delta X}\\&\quad e^{(s-L_{n-1}-i\Delta X){\rm ad}_{\Theta_{n(i-1)}^{\vee}}}{\rm {ad}}_{\boldsymbol{\eta}(s)}\alpha_{n(i-1)}{\rm d}s\Big]\dot{\overline{\boldsymbol{\xi}}}_{n-1}\\
&+ \sum_{i=1}^{j}\Big[\Big( \prod_{\tau=i}^{j-1}e^{-{\rm ad}_{\Delta X\Theta_{n\tau}^{\vee}}}\Big) \int_{L_{n-1}+(i-1)\Delta X}^{L_{n-1}+i\Delta X}\\&\quad e^{(s-L_{n-1}-i\Delta X){\rm ad}_{\Theta_{n(i-1)}^{\vee}}}{\rm {ad}_{\boldsymbol{\eta}(s)}}\beta_{n(i-1)}{\rm d}s\Big]\dot{\overline{\boldsymbol{\xi}}}_n\Biggr\}\\
&+\int_{L_{n-1}+j\Delta X}^{X}e^{-(X-s){\rm {ad}}_{\Theta_{nj}^{\vee}}} \alpha_{nj}{\rm d}s\ddot{\overline{{\boldsymbol{\xi}}}}_{n-1}\\&+\int_{L_{n-1}+j\Delta X}^{X}e^{-(X-s){\rm {ad}}_{\Theta_{nj}^{\vee}}} \beta_{nj}{\rm d}s\ddot{\overline{{\boldsymbol{\xi}}}}_n\\
&+\int_{L_{n-1}+j\Delta X}^{X}e^{-(X-s){\rm {ad}}_{\Theta_{nj}^{\vee}}}{\rm{ad}}_{\boldsymbol{\eta}(s)}\alpha_{nj}{\rm d}s\dot{{\overline{\boldsymbol{\xi}}}}_{n-1}\\&+\int_{L_{n-1}+j\Delta X}^{X}e^{-(X-s){\rm {ad}}_{\Theta_{nj}^{\vee}}}{\rm{ad}}_{\boldsymbol{\eta}(s)}\beta_{nj}{\rm d}s\dot{{\overline{\boldsymbol{\xi}}}}_n
\label{acce2}
\end{split}
\end{equation}
\subsection{Jacobian Matrix and Its Derivative in Time}\label{Jaxobian_and_its_derivative}
From recursive use of (\ref{velo2}) and (\ref{acce2}) for the whole manipulator, the thorough calculations of the Jacobian matrix $\small\boldsymbol{J}(\boldsymbol{q},X)$ and its partial derivative in time $\small\dot{\boldsymbol{J}}(\boldsymbol{q},\dot{\boldsymbol{q}},X)$ are respectively given by
%Extracting $\small\overline{\boldsymbol{\xi}}_{n-1}$ and $\small\overline{\boldsymbol{\xi}}_n$ of section $n$ from $(\ref{velo2})$ and gluing the coefficient matrices together, then repeating this for the whole manipulator yields 
\begin{figure*}[!t]	
	\normalsize
	% Store the current equation number.
	\setcounter{MYtempeqncnt}{\value{equation}}
	
	\setcounter{equation}{33}
	\begin{scriptsize}
		\begin{align}	
		\boldsymbol{J}(\boldsymbol{q},X)&=\begin{cases}0 \xlongrightarrow{\qquad\qquad\qquad \ n\qquad\qquad\qquad}N\\\underbrace{{\rm T}_{\boldsymbol{g}_{11}(X)}\quad{\rm T}_{\boldsymbol{g}_{12}(X)}}_{\left[ \begin{matrix}
		\boldsymbol{S}_0& \boldsymbol{S}_1
		\end{matrix}\right] }\qquad \boldsymbol{0}_{6}\quad\cdots\quad \boldsymbol{0}_{6}\quad\qquad \ X\in(0, L_1] \vspace{1ex}\vspace{1ex}\\
		\underbrace{{\rm Ad}^{-1}_{\boldsymbol{g}_2(X)}{\rm T}_{\boldsymbol{g}_{11}(L_1)}\quad {\rm Ad}^{-1}_{\boldsymbol{g}_{2}(X)}{\rm T}_{\boldsymbol{g}_{12}(L_1)}+{\rm T}_{\boldsymbol{g}_{21}(X)}\quad{\rm T}_{\boldsymbol{g}_{22}(X)}}_{\left[ \begin{matrix}
		\boldsymbol{S}_0& \boldsymbol{S}_1& \boldsymbol{S}_2
		\end{matrix}\right] }\quad\boldsymbol{0}_6\ \cdots\ \boldsymbol{0}_{6} \quad X\in(L_1, L_2]\vspace{1ex}\vspace{1ex}\\
		\underbrace{{\rm Ad}^{-1}_{\boldsymbol{g}_3(X)}{\rm Ad}^{-1}_{\boldsymbol{g}_2(L_2)}{\rm T}_{\boldsymbol{g}_{11}(L_1)}\quad {\rm Ad}^{-1}_{\boldsymbol{g}_3(X)}\left[ {\rm Ad}^{-1}_{\boldsymbol{g}_2(L_2)}{\rm T}_{\boldsymbol{g}_{12}(L_1)}+{\rm T}_{\boldsymbol{g}_{21}(L_2)}\right] \ \cdots\ {\rm T}_{\boldsymbol{g}_{32}(X)}}_{\left[ \begin{matrix}
		\boldsymbol{S}_0& \boldsymbol{S}_1& \boldsymbol{S}_2& \boldsymbol{S}_3 \end{matrix}\right] } \ \boldsymbol{0}_6 \ \cdots\ \boldsymbol{0}_{6}\ X\in(L_2, L_3]\\
		\qquad\qquad\vdots \qquad\qquad\qquad\qquad\quad\qquad\vdots\qquad\qquad\qquad\qquad\qquad\qquad\qquad\qquad\qquad\qquad\qquad\qquad\qquad\qquad\vdots\\
		\underbrace{\prod_{i=2}^{N}{\rm Ad}^{-1}_{\boldsymbol{g}_i{\rm min}(L_i,X)}{\rm T}_{\boldsymbol{g}_{11}(L_1)}\qquad\quad\cdots\qquad\quad{\rm Ad}^{-1}_{\boldsymbol{g}_{N}(X)}{\rm T}_{\boldsymbol{g}_{(N-1)2}(X)}+{\rm T}_{\boldsymbol{g}_{N1}(X)}\quad{\rm T}_{\boldsymbol{g}_{N2}(X)}}_{\left[ \begin{matrix}
		\boldsymbol{S}_0& \boldsymbol{S}_1& \boldsymbol{S}_2& \boldsymbol{S}_3& \cdots& \boldsymbol{S}_N
		\end{matrix}\right] }\quad X\in(L_{N-1}, L_N]
		\label{Jacobian}
		\end{cases}
		\end{align}
	\end{scriptsize}
	% Restore the current equation number.
	\setcounter{equation}{\value{MYtempeqncnt}}
	% The IEEE uses as a separator
		\hrulefill
	% The spacer can be tweaked to stop underfull vboxes.
	\vspace*{-7pt}
\end{figure*}
\begin{figure*}[!t]	
	\normalsize
	% Store the current equation number.
	\setcounter{MYtempeqncnt}{\value{equation}}
	
	\setcounter{equation}{34}
	\begin{scriptsize}
		\begin{align}	
		\dot{\boldsymbol{J}}(\boldsymbol{q},\dot{\boldsymbol{q}},X)&=\begin{cases}{\rm AD}_{\boldsymbol{g}_{11}(X)}\quad{\rm AD}_{\boldsymbol{g}_{12}(X)}\qquad \boldsymbol{0}_{6}\quad\cdots\quad \boldsymbol{0}_{6}\vspace{1ex}\vspace{1ex}\\
		{\rm Ad}^{-1}_{\boldsymbol{g}_2(X)}{\rm AD}_{\boldsymbol{g}_{11}(L_1)}\quad {\rm Ad}^{-1}_{\boldsymbol{g}_{2}(X)}{\rm AD}_{\boldsymbol{g}_{12}(L_1)}+{\rm AD}_{\boldsymbol{g}_{21}(X)}\quad{\rm AD}_{\boldsymbol{g}_{22}(X)}\quad\boldsymbol{0}_6\quad \cdots\quad\boldsymbol{0}_{6} \vspace{1ex}\vspace{1ex}\\
		{\rm Ad}^{-1}_{\boldsymbol{g}_3(X)}{\rm Ad}^{-1}_{\boldsymbol{g}_2(L_2)}{\rm AD}_{\boldsymbol{g}_{11}(L_1)}\quad {\rm Ad}^{-1}_{\boldsymbol{g}_3(X)}\left[ {\rm Ad}^{-1}_{\boldsymbol{g}_2(L_2)}{\rm AD}_{\boldsymbol{g}_{12}(L_1)}+{\rm AD}_{\boldsymbol{g}_{21}(L_2)}\right] \ \cdots \ {\rm AD}_{\boldsymbol{g}_{32}(X)} \quad \boldsymbol{0}_6 \ \cdots\ \boldsymbol{0}_{6}\vspace{1ex}\vspace{1ex}\\
		\qquad\qquad\vdots \qquad\qquad\qquad\qquad\quad\qquad\vdots\qquad\qquad\qquad\qquad\qquad\vdots\vspace{1ex}\\
		\prod_{i=2}^{N}{\rm Ad}^{-1}_{\boldsymbol{g}_i{\rm min}(L_i,X)}{\rm AD}_{\boldsymbol{g}_{11}(L_1)}\qquad\cdots\qquad\cdots\qquad\cdots\quad{\rm Ad}^{-1}_{\boldsymbol{g}_{N}(X)}{\rm AD}_{\boldsymbol{g}_{(N-1)2}(X)}+{\rm AD}_{\boldsymbol{g}_{N1}(X)}\quad {\rm AD}_{\boldsymbol{g}_{N2}(X)}
		\label{Jacobian_deri}
		\end{cases}
		\end{align}
	\end{scriptsize}
	% Restore the current equation number.
	\setcounter{equation}{\value{MYtempeqncnt}}
	% The IEEE uses as a separator
	\hrulefill
	% The spacer can be tweaked to stop underfull vboxes.
	\vspace*{-7pt}
\end{figure*}
\subsection{Simplification of Generalized Internal Wrench}\label{internal_simplification}
Based on linear constitutive relationship chosen for both the elastic and viscous members, the generalized internal wrench $\small\boldsymbol{F}_{\rm i}(\boldsymbol{q},\dot{\boldsymbol{q}})$ which is a nonlinear function of the joint positions and velocities can be simplified as follows
\begin{equation*}
		\scriptsize
		\begin{split}
		&\boldsymbol{F}_{{\rm i}}(\boldsymbol{q},\dot{\boldsymbol{q}})=\boldsymbol{\mathcal{P}}^{\rm T}\int_{0}^{L_N}\boldsymbol{J}^{\rm T}\left( \boldsymbol{\mathcal{F}}'_{ie}-{\rm{ad}}^{\rm T}_{\boldsymbol{{\xi}}}\boldsymbol{\mathcal{F}}_{ie}\right){\rm d}X\\
		&=\boldsymbol{\mathcal{P}}^{\rm T}\int_{0}^{L_N}\boldsymbol{J}^{\rm T}\Big[\Big(\boldsymbol{\Sigma}'-{\rm ad}^{\rm T}_{\boldsymbol{\xi}}\boldsymbol{\Sigma}\Big)\Big( \boldsymbol{{\xi}}(X)-\boldsymbol{{\xi}}_0\Big) +\boldsymbol{\Sigma}\Big(\boldsymbol{{\xi}}(X)-\boldsymbol{\xi}_0\Big)'\\
		&\quad+\left(\boldsymbol{\gamma}'-{\rm ad}^{\rm T}_{\boldsymbol{\xi}}\boldsymbol{\gamma}\right) \dot{{\boldsymbol{\xi}}}(X)+\boldsymbol{\gamma}\dot{{\boldsymbol{\xi}}}'(X)\Big]{\rm d}X\\
		&=\boldsymbol{\mathcal{P}}^{\rm T}\sum_{n=1}^{N}\int_{L_{n-1}}^{L_n}\boldsymbol{J}^{\rm T}\Big[\Big(\boldsymbol{\Sigma}'-{\rm ad}^{\rm T}_{\boldsymbol{\xi}_n}\boldsymbol{\Sigma}\Big)\Big( (\overline{\boldsymbol{\xi}}_{n-1}-\boldsymbol{{\xi}}_{(n-1)0})a_n(X)+\\
		&\quad (\overline{\boldsymbol{\xi}}_n-\boldsymbol{\xi}_{n0})b_n(X)\Big)\\ &\quad+\boldsymbol{\Sigma}(\overline{\boldsymbol{\xi}}_{n-1}-\boldsymbol{{\xi}}_{(n-1)0})a_n(X)'+ (\overline{\boldsymbol{\xi}}_n-\boldsymbol{\xi}_{n0})b_n(X)'\Big]{\rm d}X\\
		&\quad+\boldsymbol{\mathcal{P}}^{\rm T}\sum_{n=1}^{N}\int_{L_{n-1}}^{L_n}\boldsymbol{J}^{\rm T}\Big[ \left(\boldsymbol{\gamma}'-{\rm ad}^{\rm T}_{\boldsymbol{\xi}_n}\boldsymbol{\gamma}\right) \left(\dot{\overline{\boldsymbol{\xi}}}_{n-1}a_n(X)+\dot{\overline{\boldsymbol{\xi}}}_nb_n(X)\right)\\
		&\quad +\boldsymbol{\gamma}\left(\dot{\overline{\boldsymbol{\xi}}}_{n-1}a_n(X)'+\dot{\overline{\boldsymbol{\xi}}}_nb_n(X)'\right)\Big] {\rm d}X\\	&=\boldsymbol{\mathcal{P}}^{\rm T}\sum_{n=1}^{N}\int_{L_{n-1}}^{L_n}\boldsymbol{J}^{\rm T}\Biggl\{ \Big(\boldsymbol{\Sigma}'-{\rm ad}^{\rm T}_{\boldsymbol{\xi}_n}\boldsymbol{\Sigma}\Big)\Big[\begin{matrix}
		a_n(X)\boldsymbol{\mathbf{I}}_6&b_n(X)\boldsymbol{\mathbf{I}}_6\end{matrix}\Big]  {\rm d}X\boldsymbol{\mathcal{Y}}_n\\
		&\quad+\boldsymbol{\Sigma}\Big[ \begin{matrix}	a_n(X)'\boldsymbol{\mathbf{I}}_6&b_n(X)'\boldsymbol{\mathbf{I}}_6
		\end{matrix}\Big]{\rm d}X\boldsymbol{\mathcal{Y}}_n\Biggr\} \\	&\quad+\boldsymbol{\mathcal{P}}^{\rm T}\sum_{n=1}^{N}\int_{L_{n-1}}^{L_n}\boldsymbol{J}^{\rm T}\Biggl\{ \Big(\boldsymbol{\gamma}'-{\rm ad}^{\rm T}_{\boldsymbol{\xi}_n}\boldsymbol{\gamma}\Big)\Big[\begin{matrix}a_n(X)\boldsymbol{\mathbf{I}}_6&b_n(X)\boldsymbol{\mathbf{I}}_6
		\end{matrix}\Big]{\rm d}X\dot{\boldsymbol{\mathcal{Y}}}_n\\
		&\quad+\boldsymbol{\gamma}\Big[ \begin{matrix}
		a_n(X)'\boldsymbol{\mathbf{I}}_6&b_n(X)'\boldsymbol{\mathbf{I}}_6
		\end{matrix}\Big] {\rm d}X\dot{\boldsymbol{\mathcal{Y}}}_n\Biggr\} \\
		&=\boldsymbol{\mathcal{P}}^{\rm T}\sum_{n=1}^{N}\int_{L_{n-1}}^{L_n}\Big( \boldsymbol{A}_n+\boldsymbol{B}_n\Big){\rm d}X \boldsymbol{\mathcal{Y}}_n\\
		&\quad+\boldsymbol{\mathcal{P}}^{\rm T}\sum_{n=1}^{N}\int_{L_{n-1}}^{L_n}\Big( \boldsymbol{C}_n+\boldsymbol{D}_n\Big){\rm d}X \dot{\boldsymbol{\mathcal{Y}}}_n\\
		&=\boldsymbol{\mathcal{Q}}_1 \overline{\boldsymbol{\mathcal{Y}}}+\boldsymbol{\mathcal{Q}}_2 \dot{\overline{\boldsymbol{\mathcal{Y}}}}\\
		&=\boldsymbol{\mathcal{Q}}_1\boldsymbol{\mathcal{I}}(\boldsymbol{q}-\boldsymbol{q}_0)+\boldsymbol{\mathcal{Q}}_2\boldsymbol{\mathcal{I}}\dot{\boldsymbol{q}}\\
		&=\boldsymbol{K}(\boldsymbol{q})\Big(\boldsymbol{q}-\boldsymbol{q}_0\Big)+\boldsymbol{D}(\boldsymbol{q})\dot{\boldsymbol{q}}
		\label{generalize-internal}
		\end{split}
\end{equation*}
with
$$\scriptsize\boldsymbol{\mathcal{Y}}_n=\left[\begin{matrix}
\overline{\boldsymbol{\xi}}_{n-1}-\boldsymbol{\xi}_{(n-1)0}\\\overline{\boldsymbol{\xi}}_n-\boldsymbol{\xi}_{n0}\end{matrix}\right],\
\overline{\boldsymbol{\mathcal{Y}}}=\left[\begin{matrix}
\overline{\boldsymbol{\xi}}_{0}-\boldsymbol{\xi}_{00}\\\overline{\boldsymbol{\xi}}_1-\boldsymbol{\xi}_{10}\\
\overline{\boldsymbol{\xi}}_1-\boldsymbol{\xi}_{10}\\
\overline{\boldsymbol{\xi}}_2-\boldsymbol{\xi}_{20}\\
\vdots\\
\overline{\boldsymbol{\xi}}_{N-1}-\boldsymbol{\xi}_{(N-1)0}\\
\overline{\boldsymbol{\xi}}_N-\boldsymbol{\xi}_{N0}\\
\end{matrix}\right],$$
\begin{equation*}
		\scriptsize
		\begin{split}
		\boldsymbol{A}_n&=\boldsymbol{J}^{\rm T} (\boldsymbol{\Sigma}'-{\rm ad}^{\rm T}_{\boldsymbol{\xi}_n}\boldsymbol{\Sigma})\left[\begin{matrix}
		a_n(X)\boldsymbol{\mathbf{I}}_6&b_n(X)\boldsymbol{\mathbf{I}}_6
		\end{matrix}\right] 
		,\\ 
		\boldsymbol{B}_n&=\boldsymbol{J}^{\rm T}\boldsymbol{\Sigma}\left[ \begin{matrix}
		a_n(X)'\boldsymbol{\mathbf{I}}_6&b_n(X)'\boldsymbol{\mathbf{I}}_6
		\end{matrix}\right], \\ 
		\boldsymbol{C}_n&=\boldsymbol{J}^{\rm T} (\boldsymbol{\gamma}'-{\rm ad}^{\rm T}_{\boldsymbol{\xi}_n}\boldsymbol{\gamma})\left[\begin{matrix}
		a_n(X)\boldsymbol{\mathbf{I}}_6&b_n(X)\boldsymbol{\mathbf{I}}_6
		\end{matrix}\right]  
		,\\
		\boldsymbol{D}_n&=\boldsymbol{J}^{\rm T}\boldsymbol{\gamma}\left[ \begin{matrix}
		a_n(X)'\boldsymbol{\mathbf{I}}_6&b_n(X)'\boldsymbol{\mathbf{I}}_6
		\end{matrix}\right],
		\end{split}
\end{equation*} 
		\begin{scriptsize}
		\begin{equation*}
		\begin{split}
		\small&\boldsymbol{\mathcal{Q}}_1=\boldsymbol{\mathcal{P}}^{\rm T}\Big[
		\int_{L_{0}}^{L_1}\Big( \boldsymbol{A}_1+\boldsymbol{B}_1\Big){\rm d}X\quad \int_{L_{1}}^{L_2}\Big( \boldsymbol{A}_2+\boldsymbol{B}_2\Big){\rm d}X \\
		&\quad\cdots\quad \int_{L_{N-1}}^{L_N}\Big( \boldsymbol{A}_N+\boldsymbol{B}_N\Big){\rm d}X
		\Big],
		\end{split}
		\end{equation*}
	    \end{scriptsize}
		\begin{equation*}
		\scriptsize
		\begin{split}
        \boldsymbol{\mathcal{Q}}_2&=\boldsymbol{\mathcal{P}}^{\rm T}\Big[
		\int_{L_{0}}^{L_1}\Big( \boldsymbol{C}_1+\boldsymbol{D}_1\Big){\rm d}X\quad \int_{L_{1}}^{L_2}\Big( \boldsymbol{C}_2+\boldsymbol{D}_2\Big){\rm d}X\\
		&\quad\cdots\quad\int_{L_{N-1}}^{L_N}\Big( \boldsymbol{C}_N+\boldsymbol{D}_N\Big){\rm d}X
	    \Big],
		\end{split}
		\end{equation*}	
	    $$\scriptsize \boldsymbol{\mathcal{I}}=\left[\begin{matrix}
	    \mathbf{I}_6&\mathbf{0}&\mathbf{0}&\cdots&\mathbf{0}&\mathbf{0}\\
	    \mathbf{0}&\mathbf{I}_6&\mathbf{0}&\cdots&\mathbf{0}&\mathbf{0}\\
	    \mathbf{0}&\mathbf{I}_6&\mathbf{0}&\cdots&\mathbf{0}&\mathbf{0}\\
	    \mathbf{0}&\mathbf{0}_6&\mathbf{I}_6&\cdots&\mathbf{0}&\mathbf{0}\\
	    \vdots&\vdots&\vdots&\vdots&\vdots&\vdots\\
	    \mathbf{0}&\mathbf{0}&\cdots&\mathbf{0}&\mathbf{I}_6&\mathbf{0}\\
	    \mathbf{0}&\mathbf{0}&\cdots&\mathbf{0}&\mathbf{I}_6&\mathbf{0}\\
	    \mathbf{0}&\mathbf{0}&\mathbf{0}&\cdots&\mathbf{0}&\mathbf{I}_6\\
	    \end{matrix} \right]\in\mathbb{R}^{12N\times6(N+1)} $$
		where $\small\boldsymbol{\mathcal{I}}$ represents the matrix for reducing the dimension of the strain twists, $\small a_n(X)=\frac{L_n-X}{L_n-L_{n-1}}$, and $\small b_n(X)=\frac{X-L_{n-1}}{L_n-L_{n-1}}$.

\subsection{Selection schemes of $\small\boldsymbol{\rm B}_a$ and $\small \boldsymbol{\rm B}_c$ }\label{selection_B}
\begin{itemize}
	\item[$\blacksquare$]  All strain modes but two curvatures on the $Y-$axis and $Z-$axis are neglected for Euler-Bernoulli beam in 3-D space.
	$$\scriptsize\boldsymbol{\rm B}_a=\left[ \begin{matrix}
	0&0\\
	1&0\\
	0&1\\
	0&0\\
	0&0\\
	0&0
	\end{matrix}\right], \  \boldsymbol{\rm B}_c=\left[ \begin{matrix}
	1&0&0&0\\
	0&0&0&0\\
	0&0&0&0\\
	0&1&0&0\\
	0&0&1&0\\
	0&0&0&1
	\end{matrix}\right].$$
\end{itemize}
\begin{itemize}
	\item[$\blacksquare$] For extensible Kirchhoff rod, bending twist, and extension modes are considered. 
	$$\scriptsize\boldsymbol{\rm B}_a=\left[ \begin{matrix}
    1&0&0&0\\
    0&1&0&0\\
	0&0&1&0\\
	0&0&0&1\\
	0&0&0&0\\
	0&0&0&0
	\end{matrix}\right], \  \boldsymbol{\rm B}_c=\left[ \begin{matrix}
	0&0\\
	0&0\\
	0&0\\
	0&0\\
	1&0\\
	0&1
\end{matrix}\right].$$
\end{itemize}
\begin{itemize}
	\item[$\blacksquare$] As for Timoshenko beam, all modes except twist and extension about $X$ axis are included. 
	$$\scriptsize\boldsymbol{\rm B}_a=\left[ \begin{matrix}
	0&0&0&0\\
	0&1&0&0\\
	0&0&1&0\\
	0&0&0&0\\
	0&0&0&1\\
	0&0&0&1
	\end{matrix}\right], \  \boldsymbol{\rm B}_c=\left[ \begin{matrix}
	1&0\\
	0&0\\
	0&0\\
	0&1\\
	0&0\\
	0&0
\end{matrix}\right].$$
\end{itemize}

%\bibliographystyle{IEEEtran}

%\bibliography{references}


%\newpage
%If you have an EPS/PDF photo (graphicx package needed), extra braces are needed around the contents of the optional argument to biography to prevent the LaTeX parser from getting confused when it sees the complicated $\backslash${\tt{includegraphics}} command within an optional argument. (You can create your own custom macro containing the $\backslash${\tt{includegraphics}} command to make things simpler here.)
 
%\vspace{-20pt}

%\bf{If you include a photo:}\vspace{-33pt}

\begin{IEEEbiography}[{\includegraphics[width=1in,height=1.25in,clip,keepaspectratio]{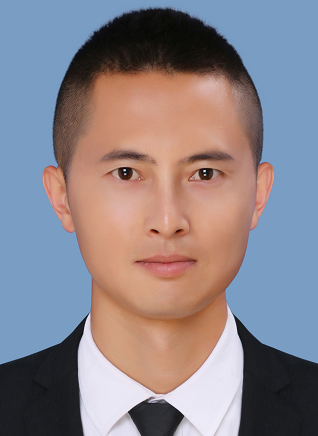}}]{Haihong Li}
	received the B.E degree in mechanical engineering from Chongqing University of Arts and Sciences, Chongqing, China, in 2016 and a M.E degree in Traffic and transportation engineering from Central South University, China, in 2019. He is now a Ph.D. candidate in soft robotics at INRIA, Lille, France. His research interests include the modeling of soft robotics and its applications in design and control.
\end{IEEEbiography}

\begin{IEEEbiography}[{\includegraphics[width=1in,height=1.25in,clip,keepaspectratio]{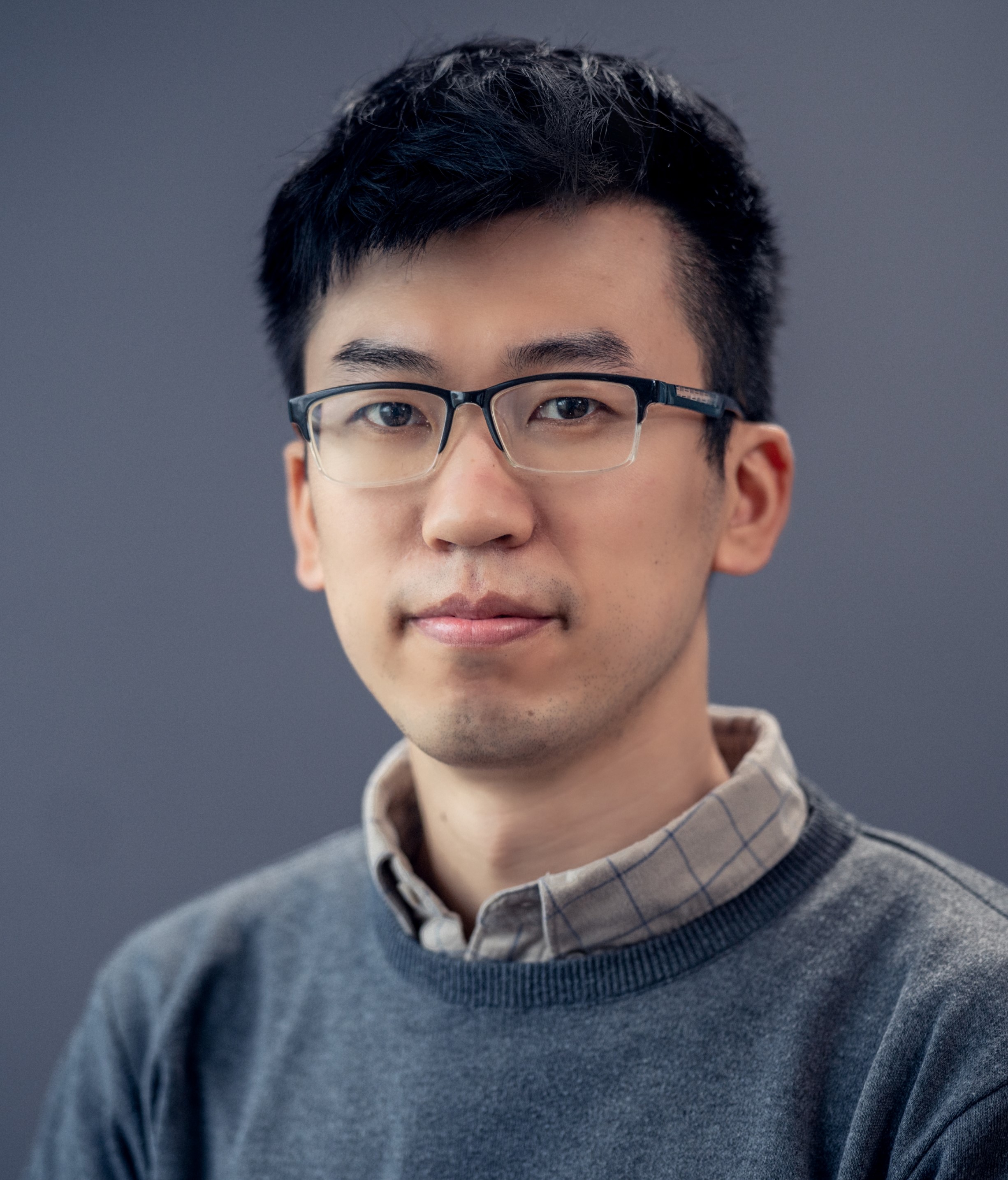}}]{Lingxiao Xun}
	received the B.E degree in mechanical engineering  from Nanjing University of Aeronautics and Astronautics, Nanjing, China, in 2017 and a M.E degree in Mechatronics system from ENSAM, France, in 2019. He is now a Ph.D. candidate in robotics at INRIA, Lille, France. His research interests include mechatronics, automation and robotics.
\end{IEEEbiography}

\begin{IEEEbiography}[{\includegraphics[width=1in,height=1.25in,clip,keepaspectratio]{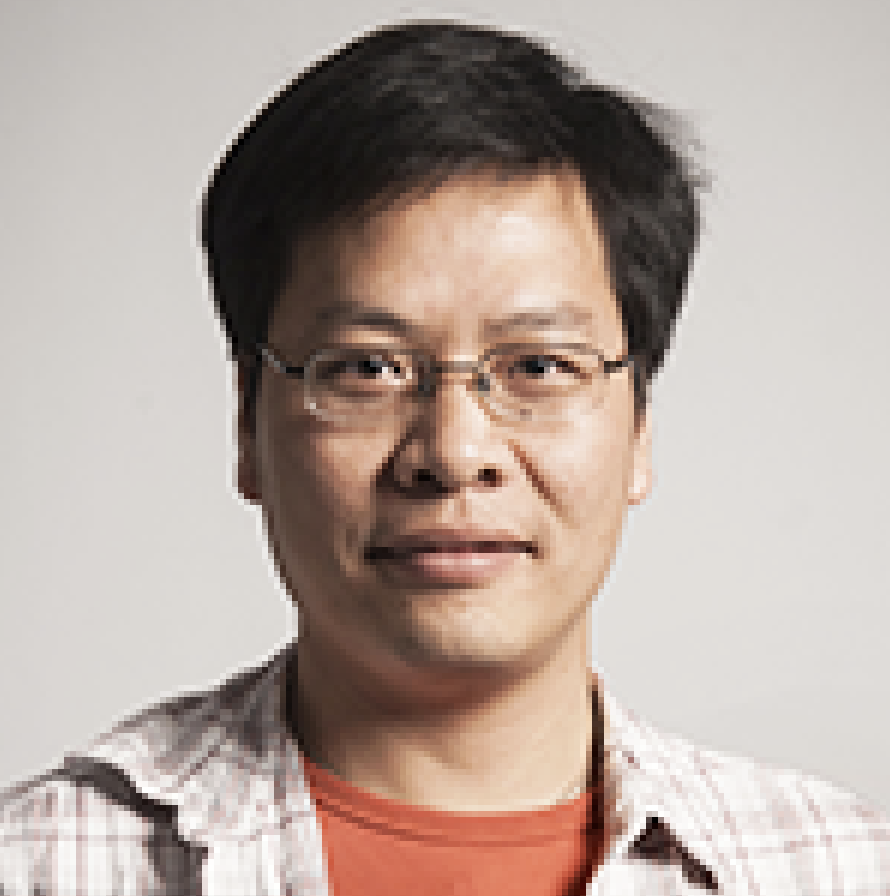}}]{Gang Zheng}
received the B.E. and M.E. degrees in Communication and systems from Wuhan University, China, in 2001 and 2004, respectively, and the Ph.D. degree in automatic control from ENSEA, Cergy-Pontoise, France, in 2006. Since 2007, he has held postdoctoral positions at INRIA Grenoble, at the Laboratoire Jean Kuntzmann, and at ENSEA. He joined INRIA Lille as a permanent researcher from September 2009. His research interests include
	control and observation of nonlinear systems, and its applications to rigid and soft robotics. Gang Zheng is a senior member of IEEE.
\end{IEEEbiography}

\vfill

\end{document}